\RequirePackage{fix-cm}
\documentclass[twocolumn]{svjour3}          
\smartqed  
\usepackage{graphicx}
\usepackage{mathptmx}      
\usepackage{amsmath,amssymb} 
\usepackage{color}
\usepackage{epsfig}
\usepackage{comment}

\usepackage{tabularx,enumitem,mathrsfs,bm}
\usepackage{algorithm,algpseudocode}
\usepackage{textcomp}

\usepackage{natbib}

\newcommand{\mat}[1]{\vec{#1}}

\newcommand{\parder}[2]{\frac{\partial #1}{\partial #2}}

\newcommand{\etal}{\textit{et al}. }
\newcommand{\ie}{\textit{i}.\textit{e}., }
\newcommand{\eg}{\textit{e}.\textit{g}., }

\begin{document}

\title{Differential Scene Flow from Light Field Gradients
}


\author{Sizhuo Ma \and Brandon M. Smith \and Mohit Gupta}

\authorrunning{S. Ma, B. M. Smith and M. Gupta} 


\institute{Sizhuo Ma \at
           \email{sizhuoma@cs.wisc.edu}    
           \and
           Brandon M. Smith \at
           \email{bmsmith@cs.wisc.edu}    
           \and
           Mohit Gupta \at
           \email{mohitg@cs.wisc.edu}  
           \and
           Department of Computer Sciences,
           University of Wisconsin-Madison, 
           Madison, WI 53706, USA
}

\maketitle

\begin{abstract}
This paper presents novel techniques for recovering 3D dense scene flow, based on differential analysis of 4D light fields. The key enabling result is a per-ray linear equation, called the ray flow equation, that relates 3D scene flow to 4D light field gradients. The ray flow equation is invariant to 3D scene structure and applicable to a general class of scenes, but is under-constrained (3 unknowns per equation). Thus, additional constraints must be imposed to recover motion. We develop two families of scene flow algorithms by leveraging the structural similarity between ray flow and optical flow equations: local `Lucas-Kanade' ray flow and global `Horn-Schunck' ray flow, inspired by corresponding optical flow methods. We also develop a combined local-global method by utilizing the correspondence structure in the light fields. We demonstrate high precision 3D scene flow recovery for a wide range of scenarios, including rotation and non-rigid motion. We analyze the theoretical and practical performance limits of the proposed techniques via the light field structure tensor, a $3 \times 3$ matrix that encodes the local structure of light fields. We envision that the proposed analysis and algorithms will lead to design of future light-field cameras that are optimized for motion sensing, in addition to depth sensing. 
\end{abstract}

\section{Introduction} \label{sec:introduction}
The ability to measure dense 3D scene motion has numerous applications, including robot navigation, human-computer interfaces and augmented reality. Imagine a head-mounted camera tracking the 3D motion of hands for manipulation of objects in a virtual environment, or a social robot trying to determine a person's level of engagement from subtle body movements. These applications require precise measurement of per-pixel 3D scene motion, also known as scene flow~\citep{vedula1999three}.

\begin{figure*}[t!]
\centerline{\hfill 
\includegraphics[width=0.45\linewidth]{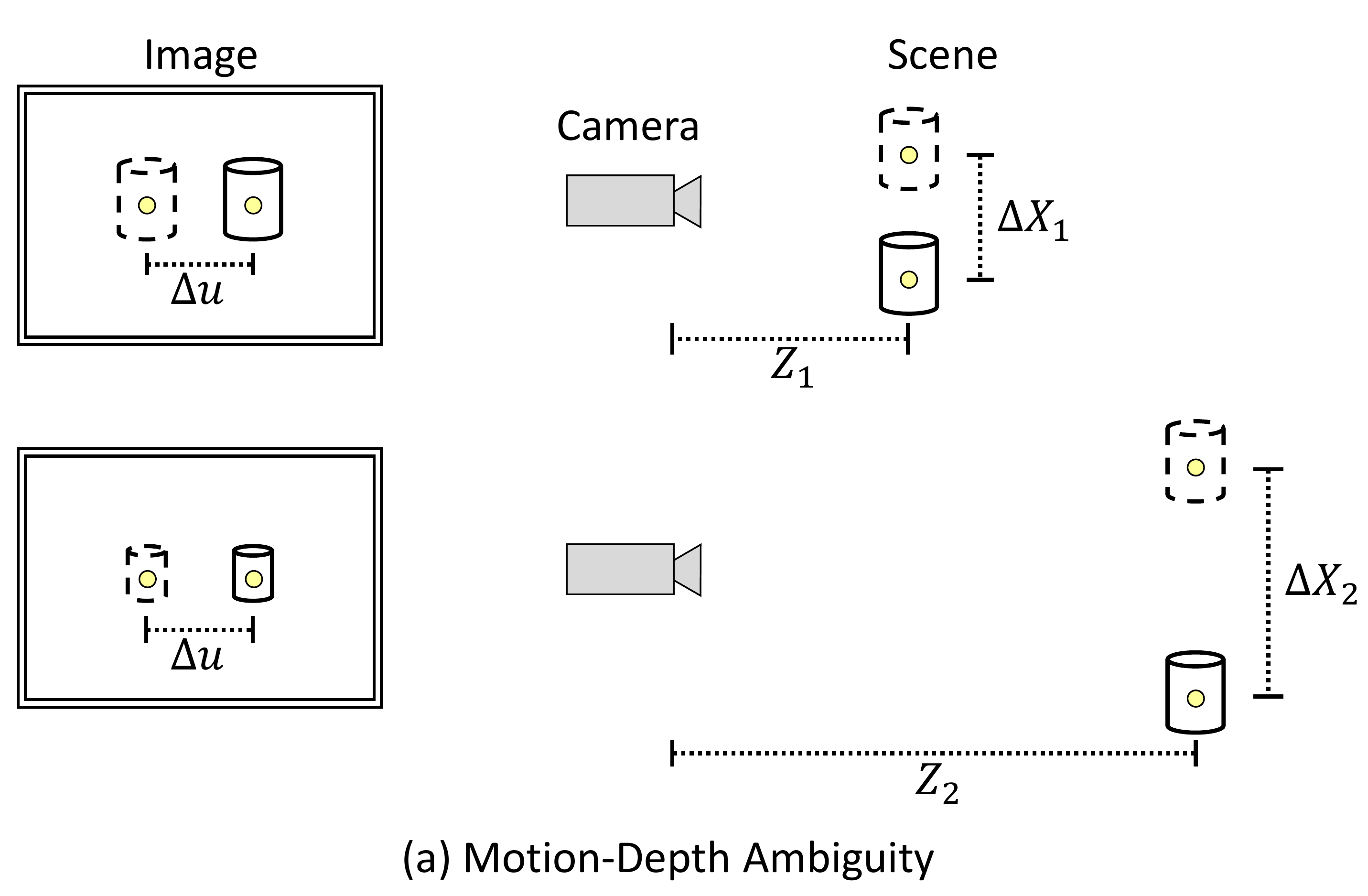} \hfill
\includegraphics[width=0.45\linewidth]{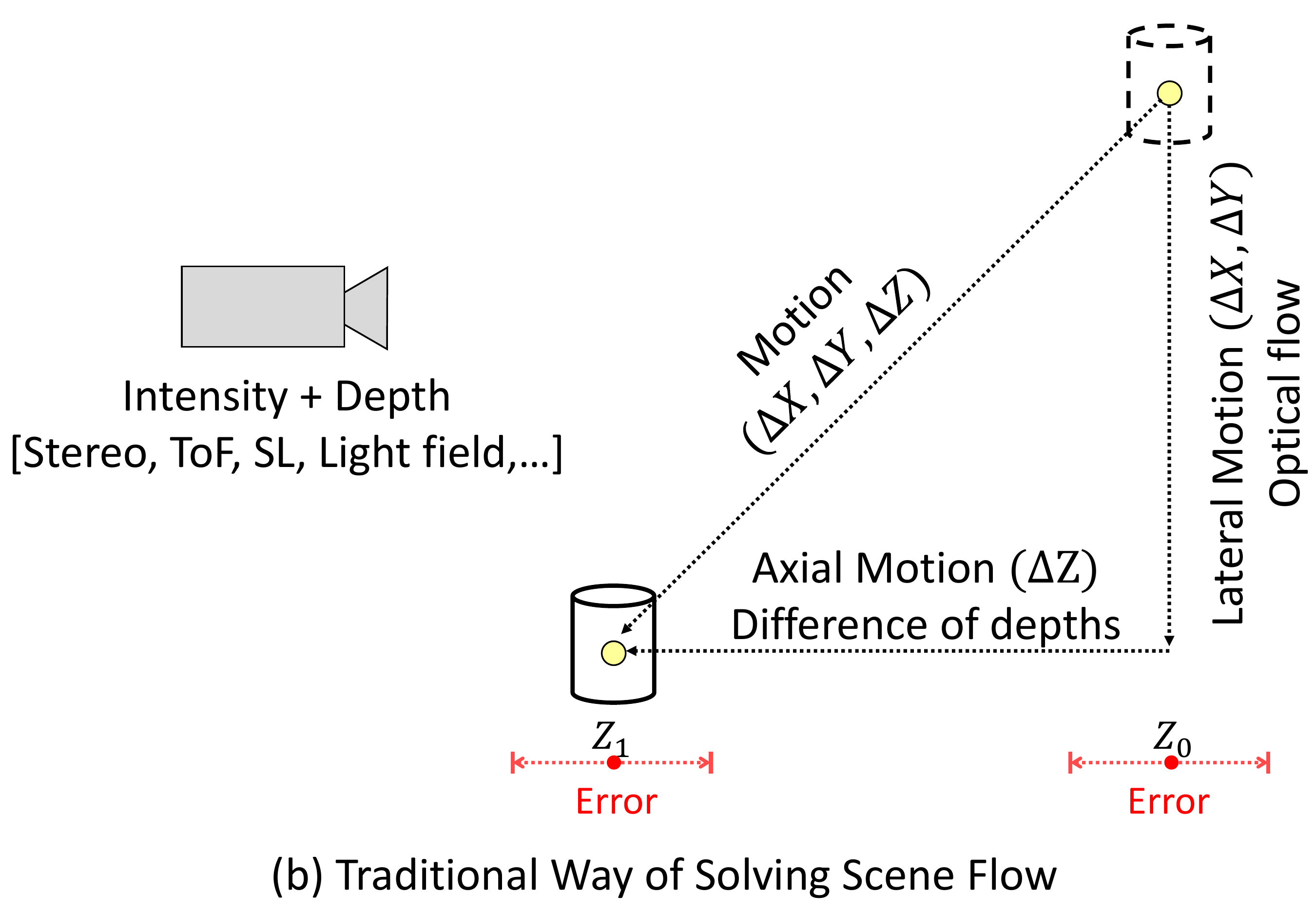} \hfill}
\caption{\textbf{Motion-depth ambiguity in conventional scene flow.} \textbf{(a)} The same amount of image-space motion $\Delta u$ may result from a combination of small scene depth $Z_1$ and small scene motion $\Delta X_1$ (top), or large depth $Z_2$ and large motion $\Delta X_2$ (bottom). Because of this motion-depth ambiguity, scene flow cannot be recovered from a single camera. \textbf{(b)} Traditional scene flow methods use an intensity+depth camera, which estimates the lateral motion via optical flow, and axial motion via difference of depths between subsequent frames ($Z_0$ and $Z_1$). Errors in the depth estimation (indicated by the red bars) get enhanced, causing large errors in axial motion estimates.} \label{fig:Tradition} 
\end{figure*}


In this paper, we present a novel approach for measuring 3D scene flow  with light field sensors \citep{adelson_single_1992,ng2005light}. This approach is based on the derivation of a new constraint, the \emph{ray flow equation}, which relates dense 3D motion field of a scene to gradients of the measured light field, as follows:
\begin{equation*}
\hspace{0.8in} \boxed{L_X \, V_X + L_Y \, V_Y + L_Z \, V_Z + L_t = 0} \,,
\end{equation*}

\noindent where $V_X, V_Y, V_Z$ are per-pixel 3D scene flow components, $L_X, L_Y, L_Z$ are spatio-angular gradients of the 4D light field, and $L_t$ is the temporal light field derivative. This simple, linear equation results from deriving the \emph{ray flow}, defined as local changes in the 4D light field, due to small, differential, 3D scene motion. 

The ray flow equation is an under-constrained linear equation with three unknowns ($V_X, V_Y, V_Z$) per equation. Therefore, it is impossible to recover the full 3D scene flow without imposing further constraints. Our key observation is that, due to the structural similarity between ray flow and the classical optical flow equations \citep{horn1981determining}, the regularization techniques developed over three decades of optical flow research can be easily adapted to constrain ray flow. The analogy between ray flow and optical flow provides a general recipe for designing ray flow based algorithms for recovering 3D dense scene flow directly from measured light field gradients. 

We develop two basic families of scene flow recovery algorithms: local \emph{Lucas-Kanade} methods, and global \emph{Horn-Schunck} methods, based on local and global optical flow \citep{lucas1981iterative,horn1981determining}. We also design a high-performance structure-aware global (SAG) method by utilizing the correspondence structure in the light fields. We adopt best practices and design choices from modern, state-of-the-art optical flow algorithms (\eg techniques for preserving motion discontinuities, recovering large motions). Using these techniques, we demonstrate 3D flow computation with \emph{sub-millimeter} precision along all three axes, for a wide range of scenarios, including complex non-rigid motion. \smallskip

\noindent {\bf Theoretical and practical performance analysis:} What is the space of motions that are recoverable by the proposed techniques? What factors influence their ability to recover 3D motion? To address these fundamental questions, we define the \emph{light field structure tensor}, a $3 \times 3$ matrix that encodes local light field structure. We show that the space of recoverable motions is determined by the properties (rank and eigenvalues) of the light field structure tensor, which depends on the scene texture. We also analyze the performance dependence of ray flow techniques on the imaging parameters of the light field camera (\eg angular resolution, aperture size and field of view \citep{dansereau2017wide}). This analysis determines theoretical and practical performance limits of the proposed algorithms, and can inform design of future light field cameras optimized for motion sensing. \smallskip

\noindent {\bf Scope and implications:} The main goal of the paper is to establish theoretical foundations of 3D scene flow computation from light field gradients. In doing so, this paper takes the first steps towards positioning light field cameras as effective 3D motion sensors, in addition to their depth estimation capabilities. Although we have implemented several proof-of-concept ray flow methods, it is possible to leverage the vast body of optical flow research and design novel, practical ray flow algorithms in the future.
These algorithms, along with novel light field camera designs optimized for motion sensing, can potentially provide high-precision 3D motion sensing capabilities in a wide range of applications, including robotic manipulation, user interfaces, and augmented reality.

\section{Related Work}

\noindent \textbf{Conventional scene flow and motion-depth ambiguity:} A single perspective camera is not sufficient for recovering 3D scene flow due to motion-depth ambiguity (Figure~\ref{fig:Tradition}(a)). Therefore, conventional scene flow approaches typically use the combination of  intensity and depth cameras (RGB-D cameras) to capture both an intensity image and a depth image of the scene \citep{jaimez2015primal,sun2015layered}. Typically, the depth information is derived from stereo \citep{wedel2008efficient,hung2013consistent}, structured light \citep{gottfried2011computing}, or time-of-flight cameras \citep{letouzey2011scene}. The lateral component of the motion is estimated via optical flow between the intensity images, and the axial component is estimated by taking the difference of the depths of corresponding points, as illustrated in Figure~\ref{fig:Tradition}(b).

This traditional approach for estimating scene flow has two major disadvantages. First, motion and depth must be solved simultaneously, which usually involves sophisticated, computationally expensive nonlinear optimization. Second, because axial motion is estimated indirectly via depth differences, errors in the individual depth estimates can amplify errors in the axial motion estimates. In contrast, the simplicity of the ray-flow equation enables estimating scene flow in a computationally tractable way. Furthermore, the ray flow equation decouples motion and depths and is independent of scene depth, which enables high-precision axial motion recovery. \smallskip



\noindent \textbf{Light field scene flow:} Scene flow methods using light fields cameras have also been proposed before \citep{heber2014scene,srinivasan2015oriented,navarro2016variational}, where light fields are used for recovering depths. Our goal is different: we use light fields for recovering 3D scene motion directly. Thus, the proposed approaches are not adversely affected by errors in measured depths, resulting in precise motion estimation, especially for subtle motions.

\smallskip\noindent \textbf{Light field odometry:} Light fields have been used for recovering a camera's ego-motion \citep{neumann2003polydioptric,dansereau2011plenoptic}, and to compute high-quality 3D scene reconstructions via structure-from-motion techniques \citep{johannsen2015linear,zhang2017light}. These methods are based on a constraint relating camera motion and light fields. This constraint has the same structural form as the equation derived in this paper, although they are derived in different contexts (camera motion vs. non-rigid scene motion) with different assumptions. These works aim to recover 6-degrees-of-freedom (6DOF) camera motion, which is an over-constrained problem. Our focus is on recovering 3D non-rigid scene motion at every pixel, which is under-constrained due to considerably higher number of degrees of freedom.

\smallskip \noindent \textbf{Micro-motion estimation using speckle imaging:} Recently, speckle-based imaging systems have been developed for measuring micron scale motion of a small number of rigidly moving objects~\citep{jo_spedo_2015,smith_colux:_2017,smith_cvpr2018}. In contrast, our goal is to recover dense, pixelwise 3D scene flow.\smallskip

\noindent \textbf{Shape recovery from differential motion:} Chandraker \etal developed a comprehensive theory for recovering shape and reflectance from differential motion of the light source, object or camera \citep{chandraker2014shape,chandraker2014camera,chandraker2016information,wang2016svbrdf,li2017robust}. While our approach is also based on a differential analysis of light fields, our goal is different -- to recover scene motion itself.

\section{The Ray Flow Equation} \label{sec:motion_tracking}

\begin{figure}[t!]
\def\dotwidB{0.18in} 
\def\dotwidC{1.5in} 
\def\dotwidD{1.5in} 
\def\dotwidE{6.7in} 
\centerline{\hfill 
\includegraphics[width=\dotwidC,angle=0]{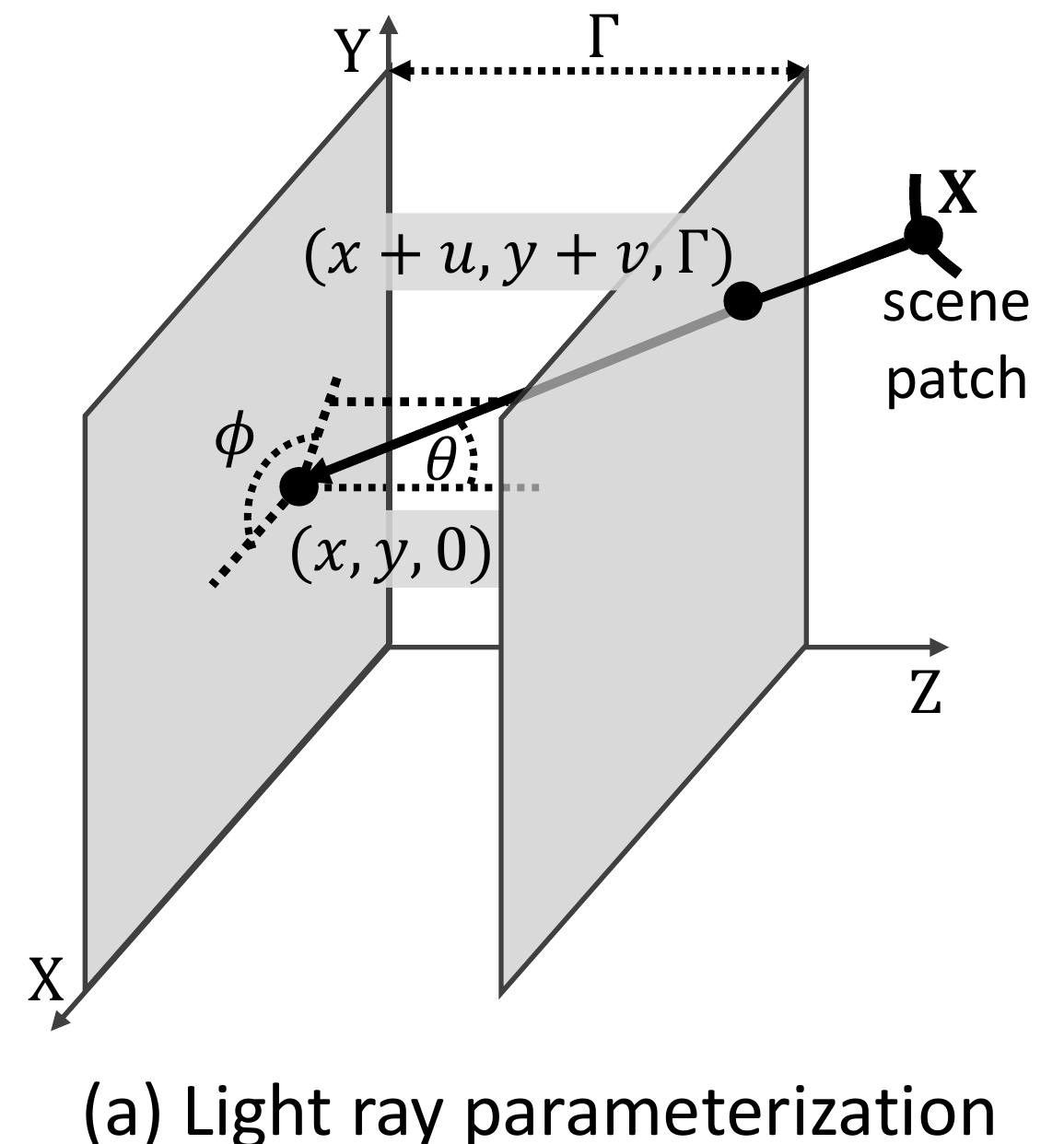} \hfill
\includegraphics[width=\dotwidC,angle=0]{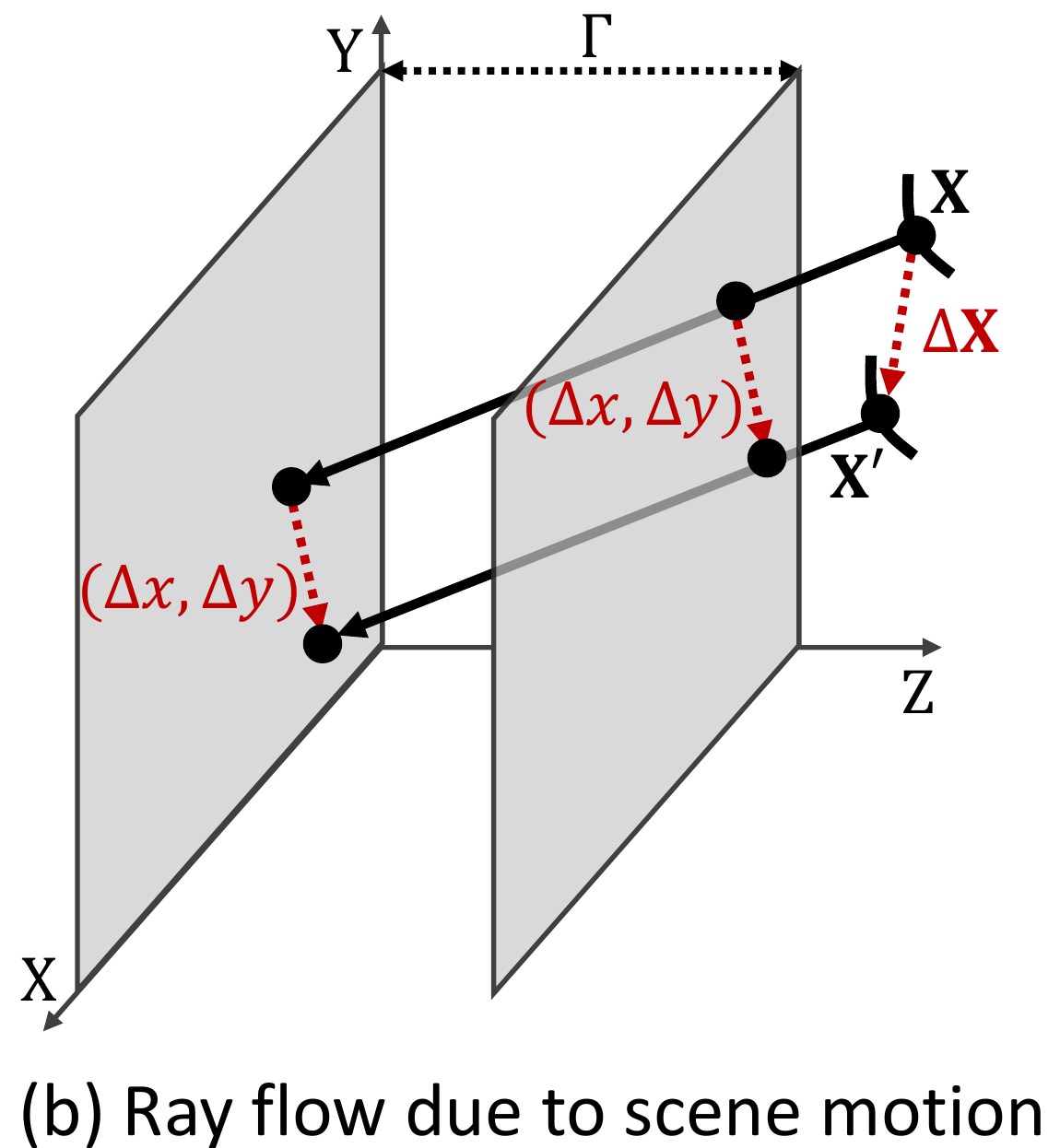} \hfill}
\caption{{\bf } \textbf{(a)} A light ray is parameterized by 4D coordinates $(x, y, u, v)$, which are determined by the ray's intersection points $(x, y, 0)$ and $(x + u, y + v, \Gamma)$ with planes $Z=0$ and $Z=\Gamma$, where $\Gamma$ is a fixed constant. \textbf{(b)} Motion (translation) of the scene point that emits or reflects the ray results in a change in the $(x, y)$ coordinates of the ray, but the $(u, v)$ coordinates remain constant. } \label{fig:Geom} 
\end{figure}

\begin{figure*}[t!]
\def\dotwidB{0.18in} 
\def\dotwidC{6.7in} 
\def\dotwidD{1.5in} 
\def\dotwidE{6.7in} 
\centerline{\includegraphics[width=\linewidth,angle=0]{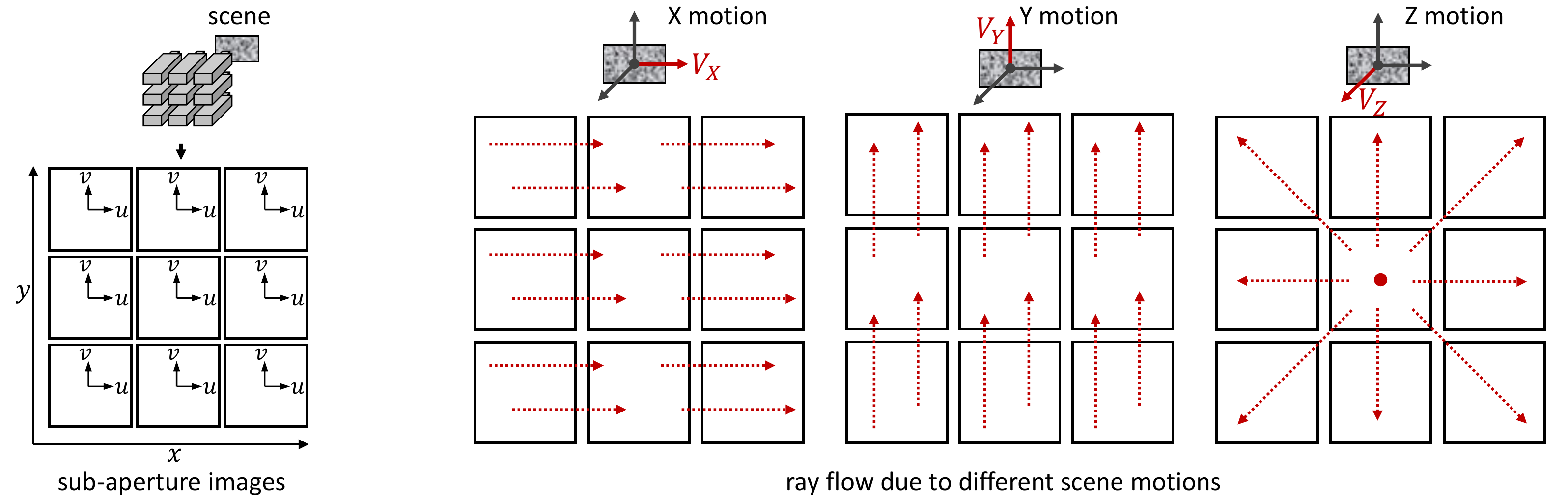}}
\caption{{\bf Ray flow due to different scene motions.} {\bf (Left)} We represent a light field sensor as a 2D array of pinhole cameras, each of which captures a 2D image (sub-aperture images). $(u, v)$ denotes the pixel indices within each sub-aperture image. $(x, y)$ denotes the locations of the cameras. {\bf (Right)} For $X/Y$ scene motion, rays move horizontally/vertically, across sub-aperture images. The amount of change $(\Delta x, \Delta y)$ in the sub-aperture index is independent of the rays' coordinates. For $Z$-motion, rays shift radially across sub-aperture images. The shift depends on each ray's $(u,v)$ coordinates. Rays at the center of each sub-aperture image $(u=0, v=0)$ do not shift. In all cases, rays retain the same pixel index $(u, v)$, but move to a different sub-aperture image.} \label{fig:RayFlow} 
\end{figure*}

Consider a scene point $P$ at 3D location $\mathbf{X} = (X, Y, Z)$. Let $L (\mathbf{X}, \theta, \phi)$ be the radiance of $P$ along direction $(\theta, \phi)$, where $\theta,\phi$ are the polar angle and azimuth angle as defined in spherical coordinates.
The function $L (\mathbf{X}, \theta, \phi)$ is called the \emph{plenoptic function}: it defines the radiance at all positions, along all possible ray directions. Assuming the radiance does not change along a ray, the 5D function $L (\mathbf{X}, \theta, \phi)$ can be simplified to the 4D \emph{light field} $L(x,y,u,v)$, with each ray parameterized by its intersections with two parallel planes $Z=0$ and $Z=\Gamma$, where $\Gamma$ is a fixed constant. This is shown in Figure~\ref{fig:Geom}(a). Let the ray intersect the planes at points $(x, y, 0)$ and $(x + u, y + v, \Gamma)$, respectively. Then, the ray is represented by the coordinates $(x, y, u, v)$. Note that $(u, v)$ are \emph{relative coordinates}; they represent the differences in the $X$ and $Y$ coordinates of the two intersection points. This is called the \emph{two-plane parameterization} of the light field \citep{levoy1996light,ng2005light}, and is widely used to represent light fields captured by cameras.

By basic trigonometry, the relationship between the \emph{scene-centric} coordinates $(X, Y, Z, \theta, \phi)$ of a light ray, and its \emph{camera-centric} coordinates $(x, y, u, v)$ is given by:
\begin{align}
&x = X - Z\tan\theta\cos\phi , &u &= \Gamma \tan\theta\cos\phi , \nonumber \\ 
&y = Y - Z\tan\theta\sin\phi , &v &= \Gamma \tan\theta\sin\phi .\label{eq:Representations}
\end{align} 

\noindent {\bf Effect of scene motion on light fields:} Let the 3D locations of a scene point $P$ at time $t$ and $t + \Delta t$ be $\mathbf{X}$ and $\mathbf{X'} = \mathbf{X} + \Delta \mathbf{X}$, where $\Delta \mathbf{X} = \left( \Delta X, \Delta Y, \Delta Z \right)$ is the small (differential) 3D motion (shown in Figure~\ref{fig:Geom}(b)). Consider a ray reflected (emitted) by $P$. We assume that the scene patch containing $P$ only translates during motion\footnote{For a rotating object, in general, the motion of small scene patches can be modeled as translation, albeit with a change in the surface normal. For small rotations (small changes in surface normal), the brightness of a patch can be assumed to be approximately constant \citep{vedula1999three}. See Section~\ref{sec:results_rotation} for simulated results.}, so that the ray only moves parallel to itself, \ie $(u, v)$ coordinates of the ray remain constant. Let the coordinates of the ray before and after motion be $(x, y, u, v)$ and $(x + \Delta x, y + \Delta y, u, v)$. Then, assuming that the ray brightness remains constant during motion\footnote{This is true under the assumption that the light sources are distant such that $\vec{N}\cdot\vec{L}$, the dot-product of surface normal and lighting direction, does not change \citep{vedula1999three}.}:
\begin{equation}\label{eq:rayBrightness}
L(x, y, u, v, t) = L(x + \Delta x, y + \Delta y, u, v, t + \Delta t) \,.
\end{equation}

This \emph{ray brightness constancy assumption} is similar to the \emph{scene brightness constancy assumption} made in optical flow. First-order Taylor expansion of Eq.~\ref{eq:rayBrightness} gives:
\begin{equation}\label{eq:rayBrightnessTaylor}
\frac{\partial L}{\partial x} \Delta x + \frac{\partial L}{\partial y} \Delta y + \frac{\partial L}{\partial t} \Delta t = 0 \,.
\end{equation}

\noindent We define \emph{ray flow} as the change $(\Delta x,\Delta y)$ in a light ray's coordinates due to scene motion. Eq.~\ref{eq:rayBrightnessTaylor} relates ray flow and light field gradients ($\frac{\partial L}{\partial x}, \frac{\partial L}{\partial y}, \frac{\partial L}{\partial t}$). From Eq.~\ref{eq:Representations}, we can also find a relationship between ray flow and scene motion:
\begin{align}
\Delta x &= \frac{\partial x}{\partial X} \Delta X + \frac{\partial x}{\partial Z} \Delta Z = \Delta X - \frac{u}{\Gamma} \Delta Z \,, \nonumber \\
\Delta y &= \frac{\partial y}{\partial Y} \Delta Y + \frac{\partial y}{\partial Z} \Delta Z = \Delta Y - \frac{v}{\Gamma} \Delta Z \,.\label{eq:DerivRelat}
\end{align}

\noindent By substituting Eq.~\ref{eq:DerivRelat} in Eq.~\ref{eq:rayBrightnessTaylor} and using symbols $L_*$ for light field gradients,  we get: 
\begin{equation}
\hspace{0.8in} \underbrace{\boxed{L_X \, V_X + L_Y \, V_Y + L_Z \, V_Z + L_t = 0}}_{\textbf{\small{Ray Flow Equation}}} \,, 
\label{eq:RayFlow}
\end{equation}
where  $L_X = \frac{\partial L}{\partial x}$, $L_Y = \frac{\partial L}{\partial y}$, $L_Z = -\frac{u}{\Gamma} \frac{\partial L}{\partial x} -\frac{v}{\Gamma} \frac{\partial L}{\partial y}$, $L_t = \frac{\partial L}{\partial t}$, $\vec{V}=(V_X,V_Y,V_Z)=(\frac{\Delta X}{\Delta t},\frac{\Delta Y}{\Delta t},\frac{\Delta Z}{\Delta t})$. We call this the \emph{ray flow equation}; it relates the 3D scene motion and the measured light field gradients. This simple, yet powerful equation enables recovery of dense scene flow from measured light field gradients, as we describe in Sections~\ref{sec:local} to~\ref{sec:combined}. In the rest of this section, we discuss salient properties of the ray flow equation in order to gain intuitions and insights into its implications. 


\subsection{Ray Flow Due to Different Scene Motions} 
Ray flows due to different scene motions have interesting qualitative differences. To visualize the difference, we represent a 4D light field sensor as a 2D array of pinhole cameras, each with a 2D image plane. In this representation $(u, v)$ coordinates of the light field $L(x,y,u,v)$ denote the pixel indices within individual images (sub-aperture images). $(x, y)$ coordinates denote the locations of the cameras, as shown in Figure~\ref{fig:RayFlow}. 

For $X/Y$ scene motion, a light ray \emph{shifts horizontally\slash vertically} across sub-aperture images. The amount of shift $(\Delta x, \Delta y)$ is \emph{independent} of the ray's original coordinates, as evident from Eq.~\ref{eq:DerivRelat}. For $Z$-motion, the ray \emph{shifts radially} across sub-aperture images. The amount of shift \emph{depends} on the ray's $(u,v)$ coordinates (c.f. Eq.~\ref{eq:DerivRelat}). For example, rays at the center of each sub-aperture image $(u=0, v=0)$ do not shift. In all cases, rays retain the \emph{same pixel index} $(u, v)$ after the motion, but in a \emph{different sub-aperture image} $(x, y)$, since scene motion results in rays translating parallel to themselves.

\subsection{Invariance of Ray Flow to Scene Depth}\label{sec:invar_depth}
An important observation is that the ray flow equation does not involve the depth or 3D position of the scene point. 
In conventional motion estimation techniques, depth and motion estimation are coupled together, and thus need to be performed simultaneously \citep{alexander2016focal}. In contrast, the ray flow equation decouples depth and motion estimation. This has important practical implications: 
3D scene motion can then be directly recovered from the light field gradients, without explicitly recovering scene depths, thereby avoiding the errors due to the intermediate depth estimation step. 

Notice that although motion estimation via ray flow does not need depth estimation, the accuracy of the estimated motion depends on scene depth. For distant scenes, the captured light field is convolved with a 4D low-pass point spread function, which makes gradient computation unreliable. As a result, scene motion
cannot be estimated reliably.

\begin{table*}[t!]
\caption{{\bf Comparisons between optical flow and ray flow.}}
\begin{center}
\begin{tabularx}{0.99\linewidth}{>{\centering} X|>{\centering\arraybackslash} X}
\hline
{\bf Optical Flow} & {\bf Ray Flow}\\
\hline \hline
Linear equation: $I_x u_x + I_y u_y + I_t = 0$ & Linear equation: $L_X V_X + L_Y V_Y + L_Z V_Z + V_t = 0$ \\ \hline
Coefficients: Image gradients $\left(I_x, I_y, I_t \right)$ &Coefficients: Light field gradients $(L_X, L_Y, L_Z, L_t )$ \\ \hline
$2$ unknowns per pixel: Pixel motion $(u_x, u_y)$ & $3$ unknowns per pixel: Scene motion $(V_X, V_Y, V_Z)$ \\ \hline
Motion $(u_x, u_y)$ computed in 2D image space (pixels) & Motion $(V_X, V_Y, V_Z)$ computed in 3D scene space (millimeters) \\ \hline
Gradients $(I_x, I_y)$ defined on 2D image grid & Gradients $(L_X, L_Y, L_Z)$ defined on 4D light field grid \\ \hline
$u_x$ and $u_y$ flow computations are symmetric & $X/Y$ and $Z$ motion computations are asymmetric \\ \hline
Size of structure tensor: $2 \times 2$ & Size of structure tensor: $3 \times 3$ \\ \hline
Possible ranks of structure tensor: $[0, 1, 2]$ & Possible ranks of structure tensor: $[0, 2, 3]$ \\ \hline
\hline
\end{tabularx}
\end{center}
\label{tab:comparison}
\end{table*}

\begin{figure}[t!]
\def\dotwidB{0.18in} 
\def\dotwidC{2.55in} 
\def\dotwidD{1.5in} 
\def\dotwidE{6.7in} 
\centerline{\includegraphics[width=\dotwidC,angle=0]{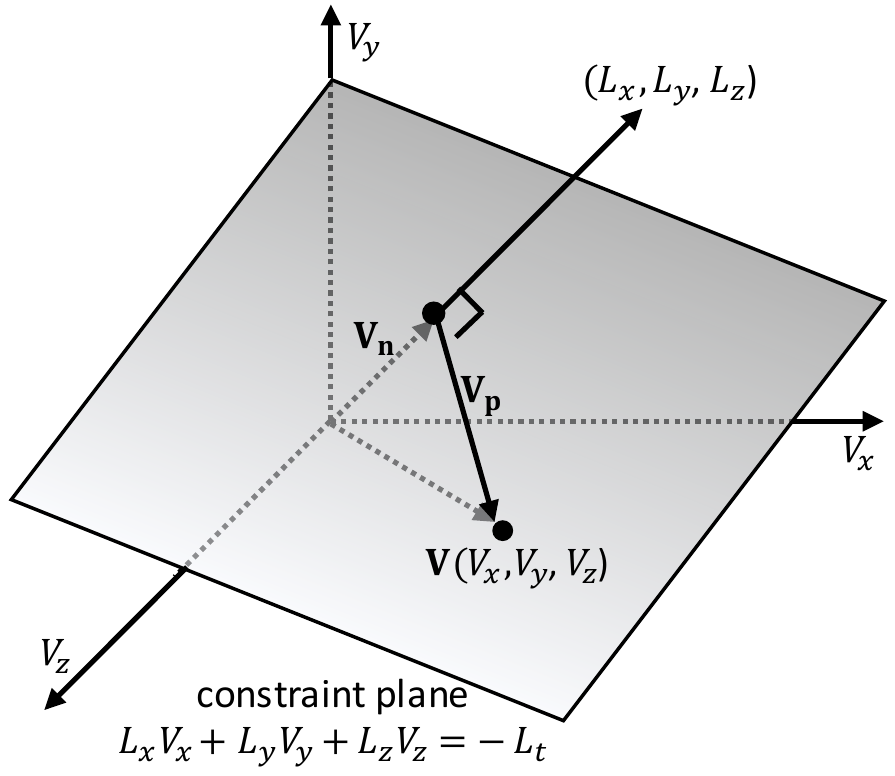}}
\vspace{0.05in}
\caption{{\bf Geometric interpretation.} The ray flow equation defines a plane in 3D velocity space. Velocity vector $\mathbf{V}$ for a given ray can be decomposed into two components: $\mathbf{V_{n}}$ (perpendicular to the plane), and $\mathbf{V_p}$ (parallel to the plane). Since $\mathbf{V_p}$ is parallel to the constraint plane, it cannot be uniquely recovered from a single equation. We call this ambiguity the ray flow aperture problem.} \label{fig:normal_flow} \vspace{-0.1in}
\end{figure}

\subsection {Geometric Interpretation of Ray Flow Equation}
Consider a three-dimensional space with axes $V_X, V_Y$ and $V_Z$, which we call \emph{motion space}. The ray flow equation defines a plane in 
motion space with the normal vector $\vec{N_L} = ( L_X, L_Y, L_Z)$, as shown in Figure~\ref{fig:normal_flow}. 

\smallskip \noindent{\bf Result 1 (Aperture problem in ray flow)} The motion vector $\mathbf{V} = \left( V_X, V_Y, V_Z\right)$ corresponding to each ray can be decomposed into two components: the recoverable flow along the plane normal $\mathbf{V_{n}}$, 
and the unrecoverable flow parallel to the plane $\mathbf{V_p}$. 

\noindent {\bf Proof:} The ray flow equation can be written as: $\mathbf{N_L} \cdot \left(\mathbf{V_{n}} + \mathbf{V_{p}} \right) = -L_t$, where $\left( \cdot \right)$ is the dot-product operator. Since $\mathbf{N_L} \cdot \mathbf{V_{p}} = 0$, this can be re-written as: $\mathbf{N_L} \cdot \mathbf{V_{n}} = -L_t$. Therefore, only the normal motion component can be recovered from a single ray flow equation. We call this ambiguity the \emph{ray flow aperture problem}. \smallskip

\subsection{Similarities between Ray Flow and Optical Flow}
Since the parallel motion component cannot be recovered, additional assumptions need to be made to further constrain the problem.
This is similar to the well-known \emph{aperture problem} in 2D optical flow, where the optical flow equation $I_x u_x + I_y u_y + I_t = 0$ is also under-constrained ($1$ equation, $2$ unknowns $\left( u_x, u_y \right)$). There are some interesting differences between ray flow and optical flow (see Table~\ref{tab:comparison}), but the key similarity is that both ray flow and optical flow are \emph{under-constrained linear equations.} 

Fortunately, optical flow is one of the most researched problems in computer vision. Broadly, there are two families of differential optical flow techniques, based on the additional constraints imposed for regularizing the problem. The first is local methods (\eg Lucas-Kanade \citep{lucas1981iterative}), which assume that the optical flow is constant within small image neighborhoods. Second is global methods (\eg Horn-Schunck \citep{horn1981determining}), which assume that the optical flow varies smoothly across the image. By exploiting the structural similarity between the optical flow and ray flow equations, we develop two families of ray flow techniques accordingly: local ray flow (Section~\ref{sec:local}) and global ray flow (Section~\ref{sec:global}).

\section{Local `Lucas-Kanade' Ray Flow} \label{sec:local}

In this section, we develop the local ray flow based scene flow recovery methods, inspired by Lucas-Kanade optical flow \citep{lucas1981iterative}. This class of ray flow methods assume that the motion vector $\mathbf{V}$ is constant in local 4D light field windows. Consider a ray with coordinates $\mathbf{x}_c = (x, y, u, v)$. We stack all the equations of form Eq.~\ref{eq:RayFlow} from rays in a local neighborhood of $\mathbf{x}_c$, $\mathbf{x}_i\in \mathscr{N}(\mathbf{x}_c)$ into a linear system $\mathbf{A V} = \mathbf{b}$, where:
\begin{equation}\label{eq:linear_system}
\mathbf{A} = \begin{bmatrix}
L_X(\mathbf{x}_1) & L_Y(\mathbf{x}_1) & L_Z(\mathbf{x}_1) \\
\vdots & \vdots & \vdots \\
L_X(\mathbf{x}_n) & L_Y(\mathbf{x}_n) & L_Z(\mathbf{x}_n) \\
\end{bmatrix}, \,\,
\mathbf{b} = \begin{bmatrix}
-L_t(\mathbf{x}_1)\\
\vdots\\
-L_t(\mathbf{x}_n)\\
\end{bmatrix}.
\end{equation}
Then, the motion vector $\mathbf{V}$ can be estimated by the normal equation: 
\begin{equation} 
\mathbf{V} = (\mathbf{A}^T\mathbf{A})^{-1}\mathbf{A}^T\mathbf{b} .
\label{eq:normal_equation}\end{equation} 

\begin{figure*}[t!]
\def\dotwidB{0.18in} 
\def\dotwidC{2.0in} 
\def\dotwidD{0.28in} 
\centerline{\hfill \hspace{\dotwidD} \hfill
\includegraphics[width=\dotwidC,angle=0]{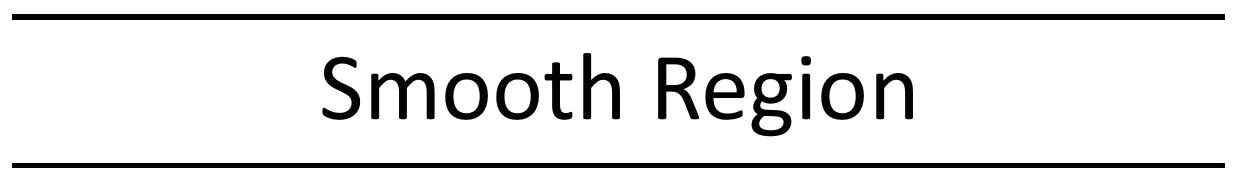} \hfill
\includegraphics[width=\dotwidC,angle=0]{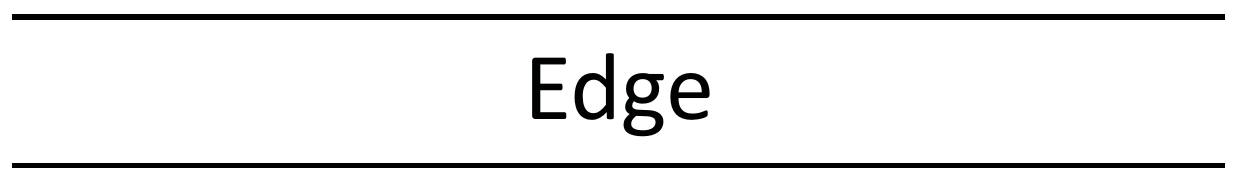} \hfill
\includegraphics[width=\dotwidC,angle=0]{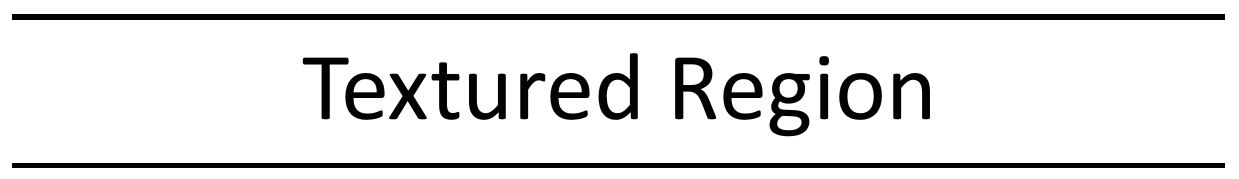} \hfill}
\centerline{\hfill 
\includegraphics[width=\dotwidD,angle=0]{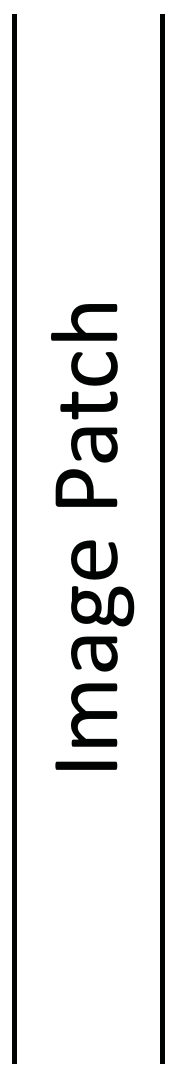} \hfill
\includegraphics[width=\dotwidC,angle=0]{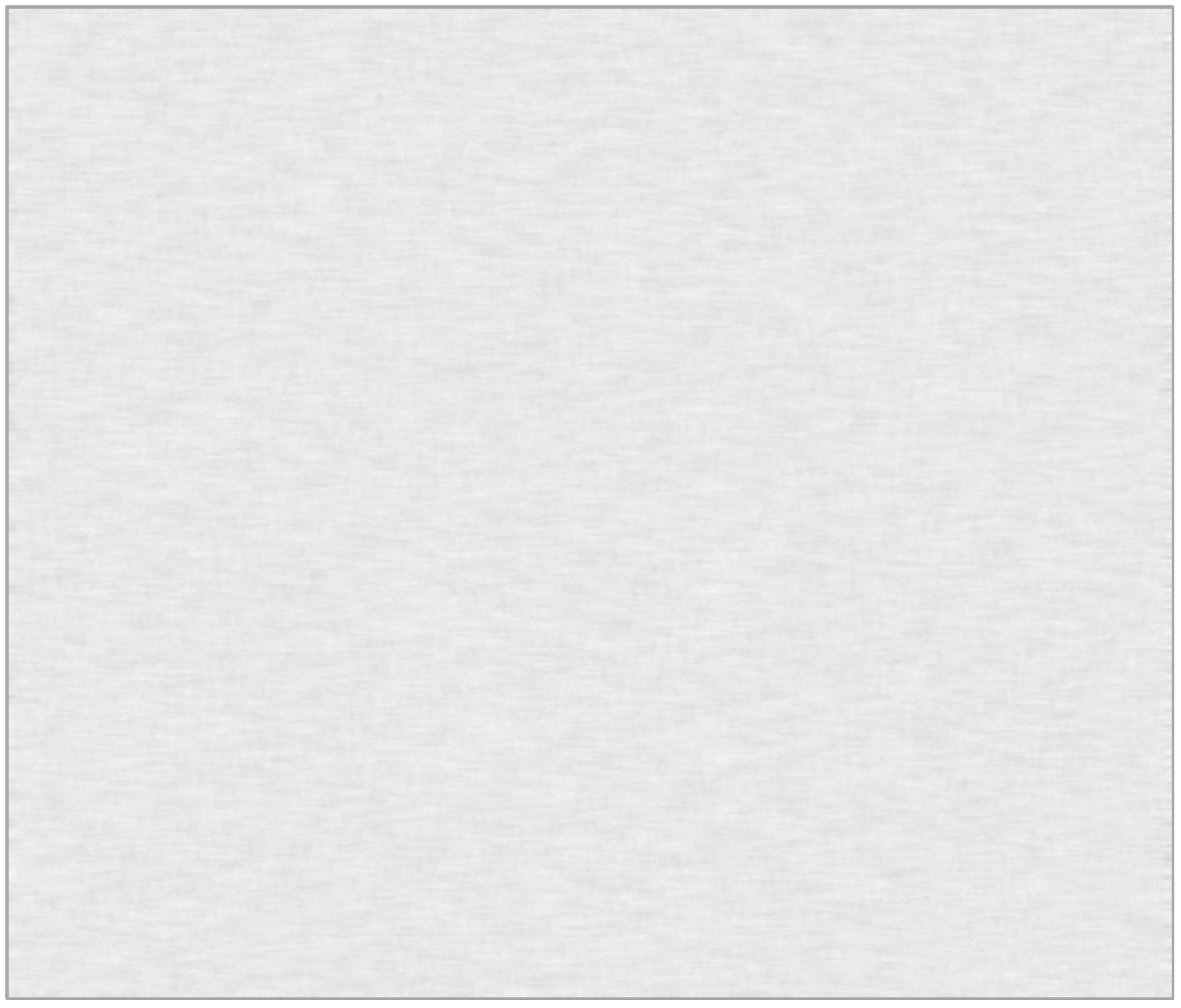} \hfill
\includegraphics[width=\dotwidC,angle=0]{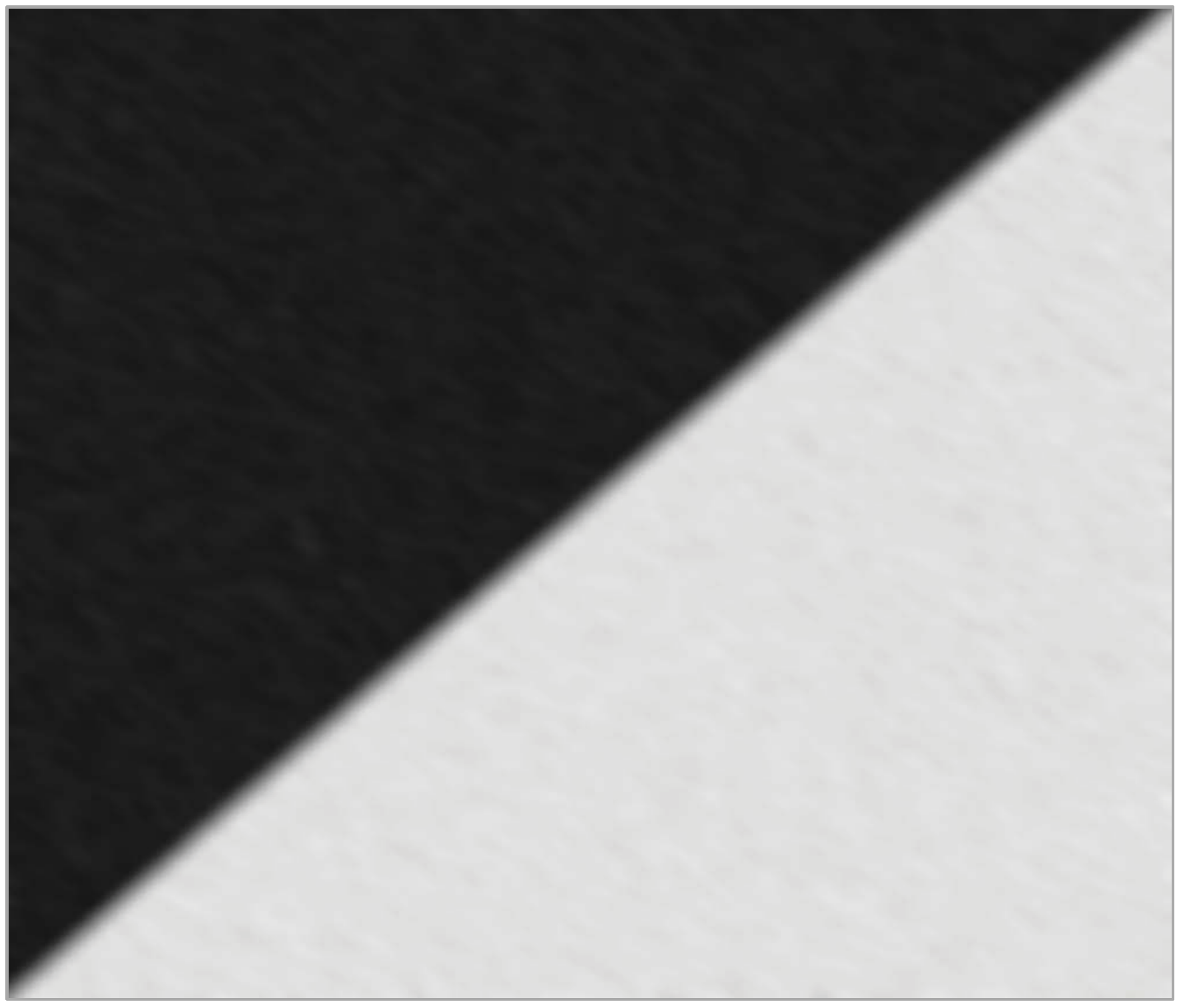} \hfill
\includegraphics[width=\dotwidC,angle=0]{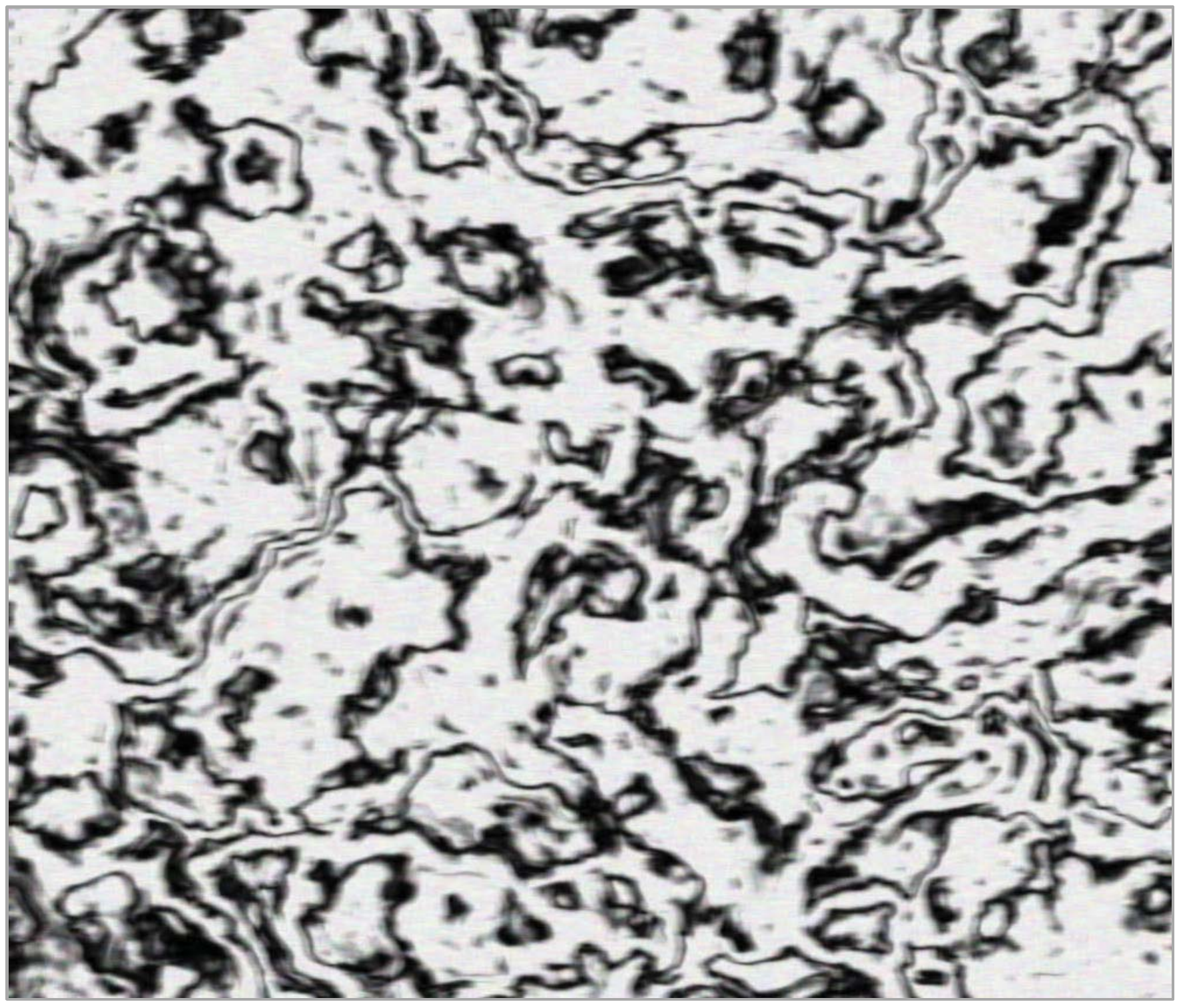} \hfill}
\centerline{\hfill 
\includegraphics[width=\dotwidD,angle=0]{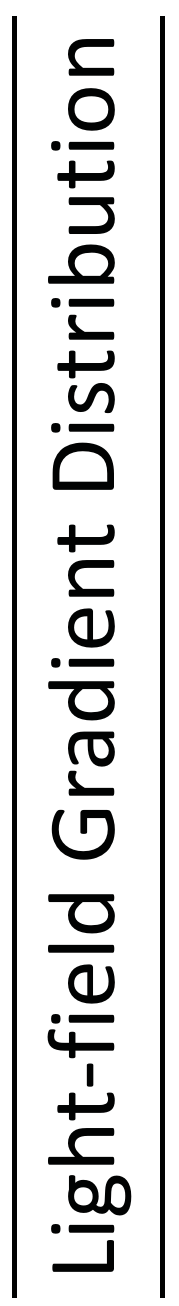} \hfill
\includegraphics[width=\dotwidC,angle=0]{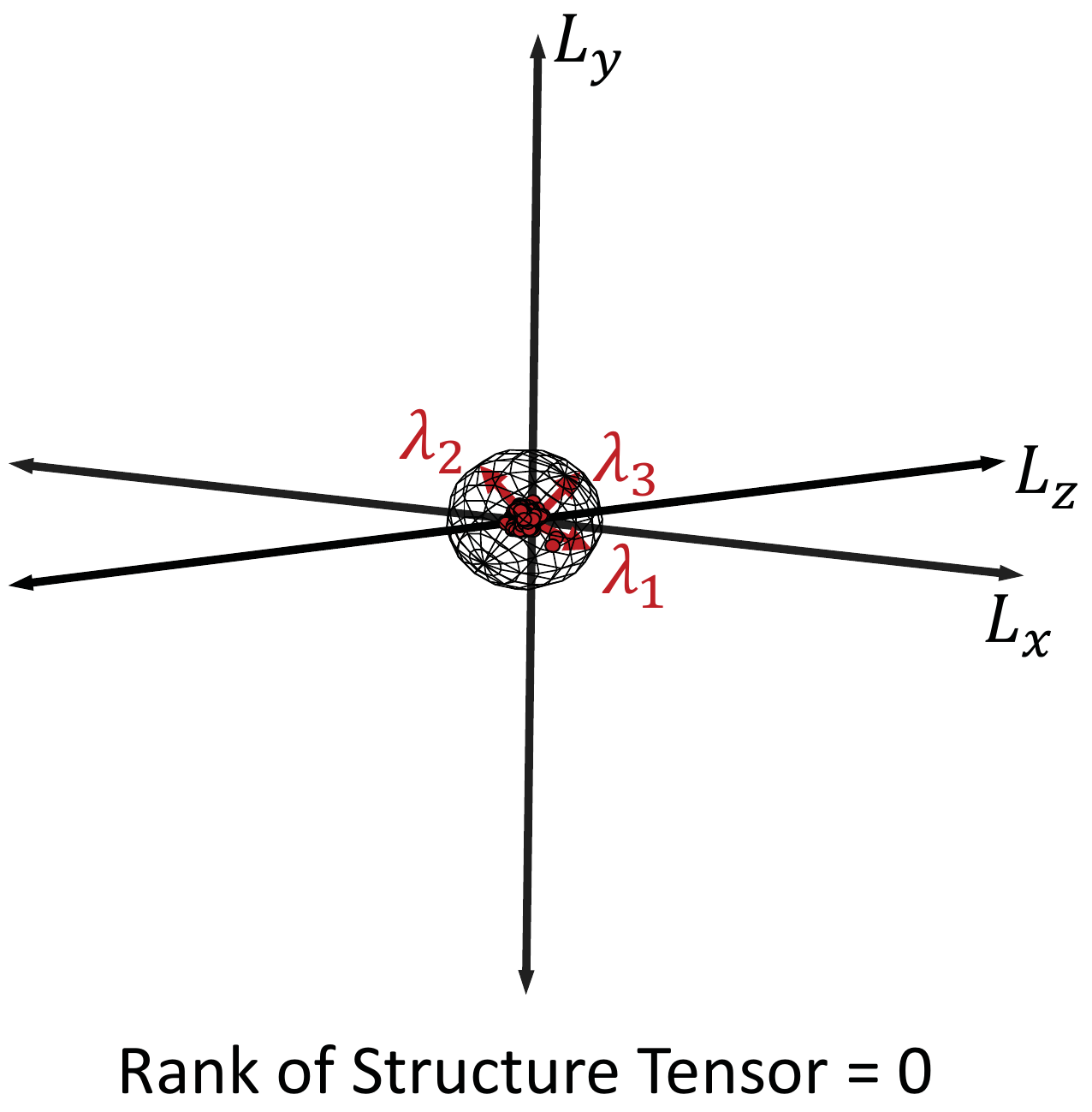} \hfill
\includegraphics[width=\dotwidC,angle=0]{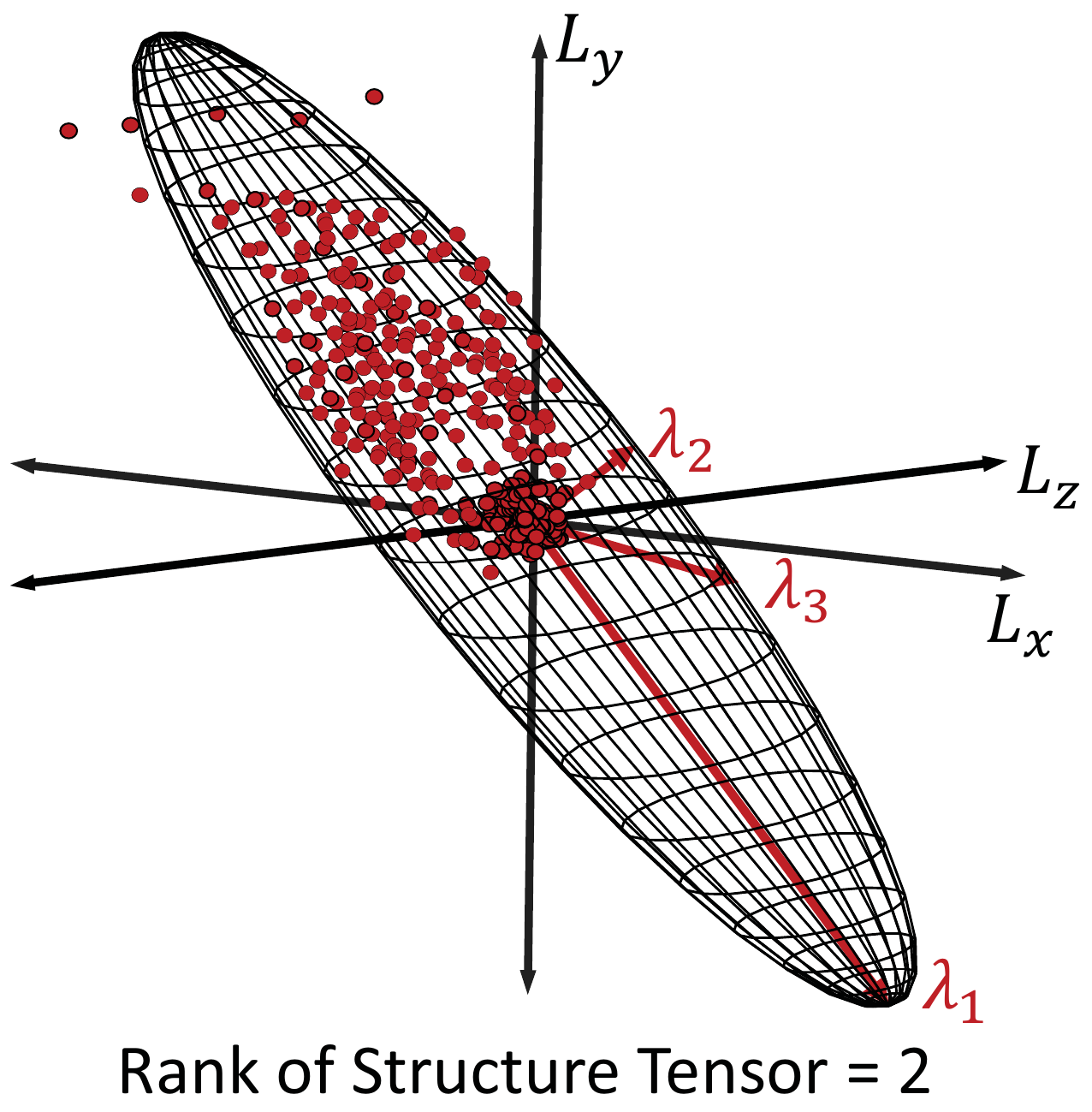} \hfill
\includegraphics[width=\dotwidC,angle=0]{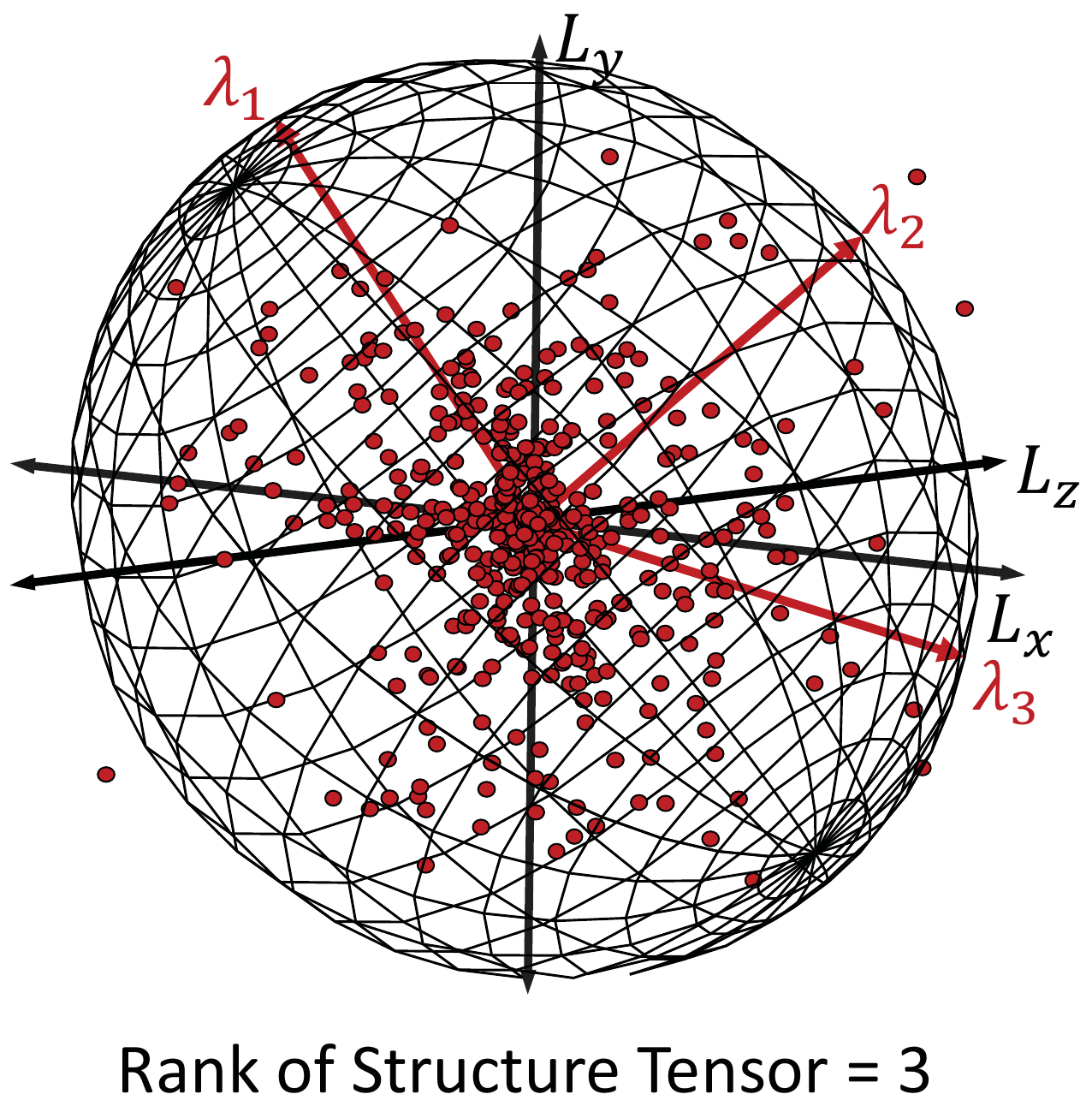} \hfill}
\centerline{\hfill 
\includegraphics[width=\dotwidD,angle=0]{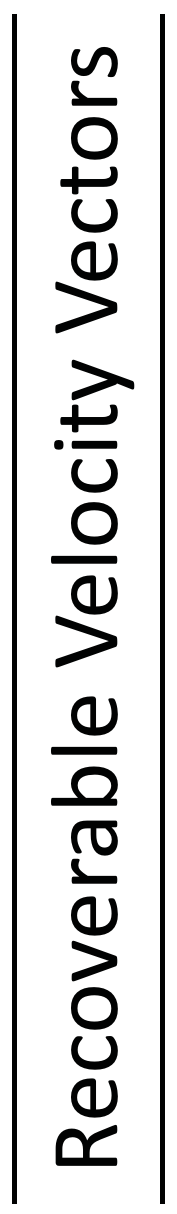} \hfill
\includegraphics[width=\dotwidC,angle=0]{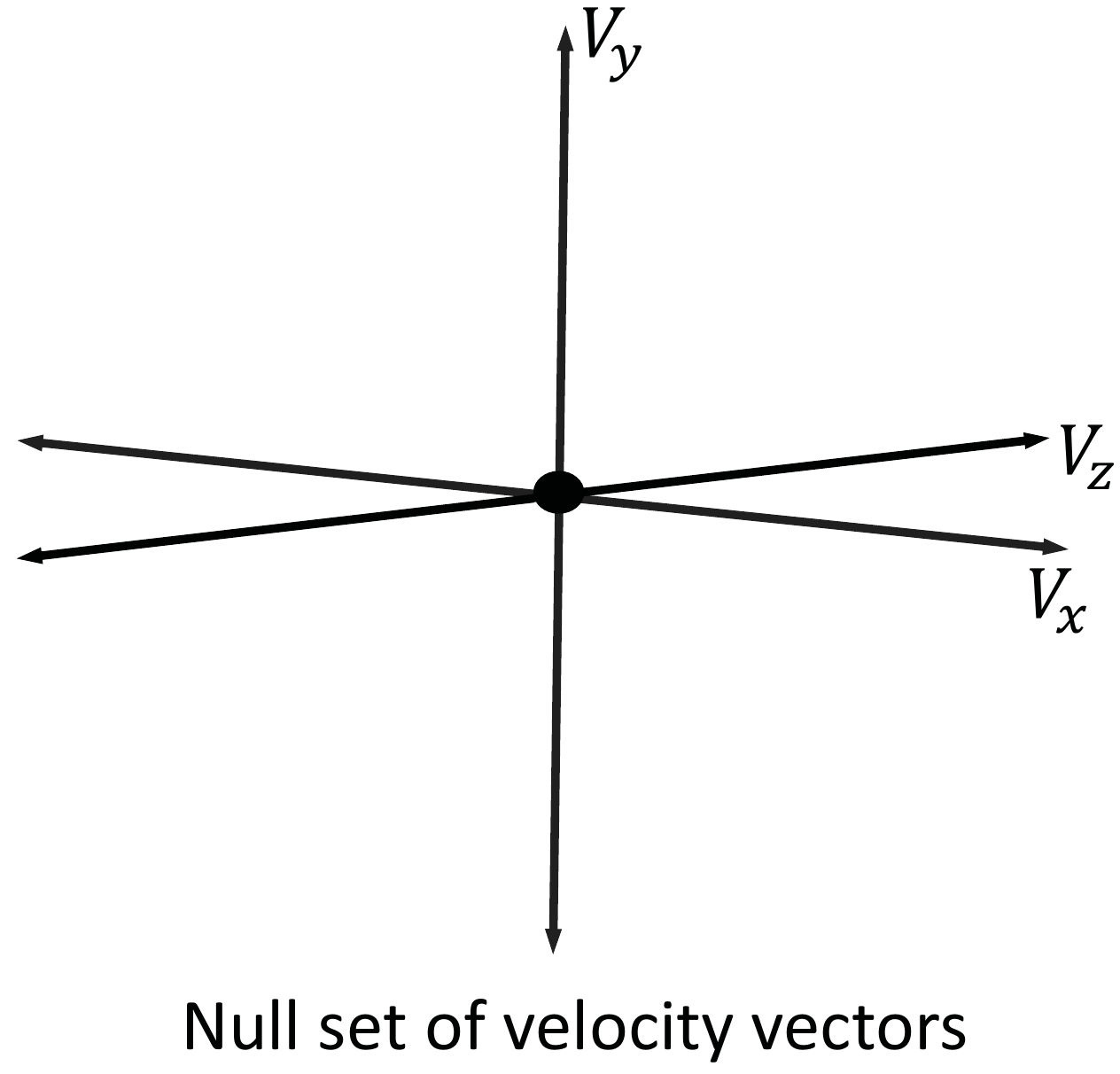} \hfill
\includegraphics[width=\dotwidC,angle=0]{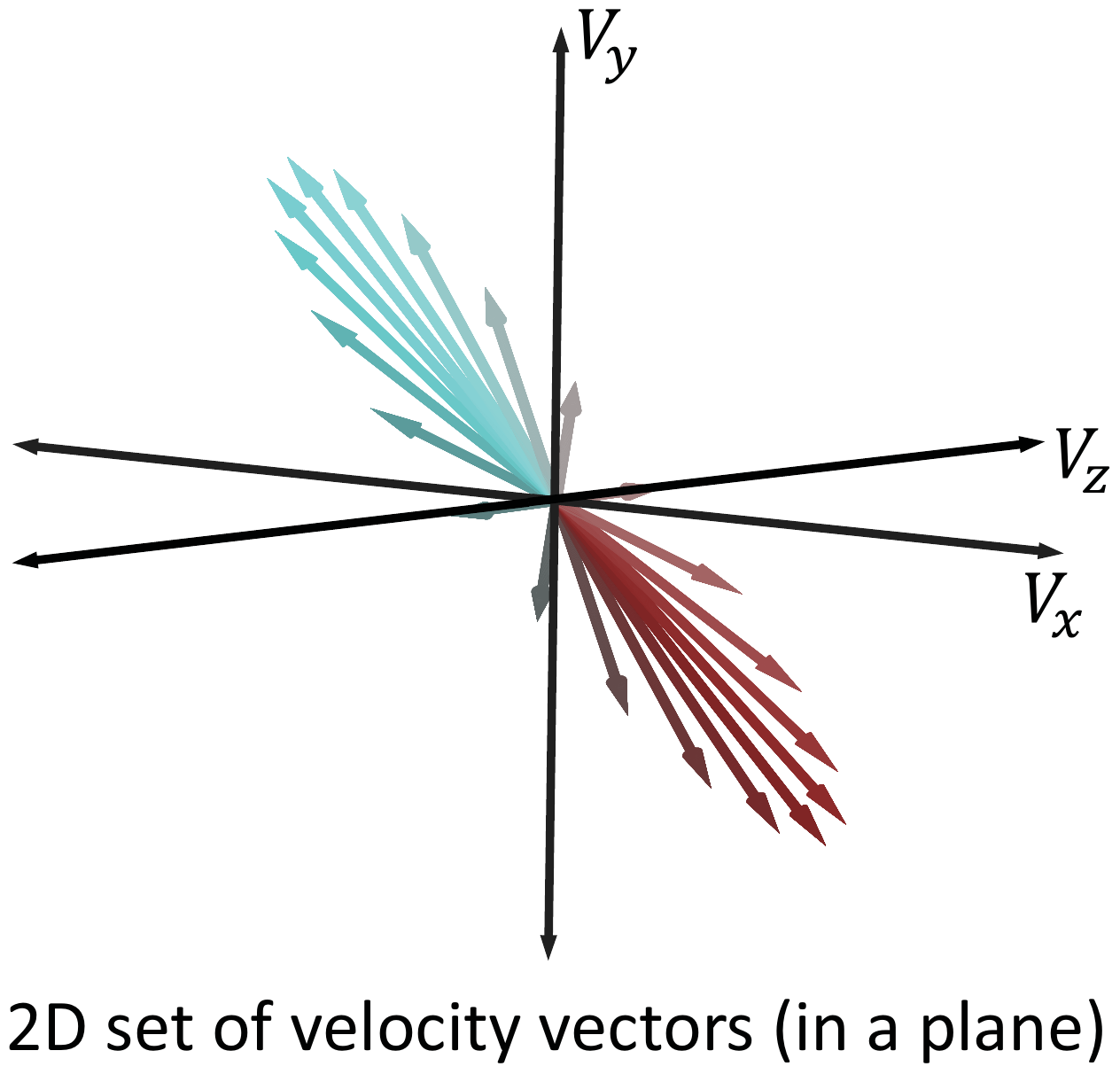} \hfill
\includegraphics[width=\dotwidC,angle=0]{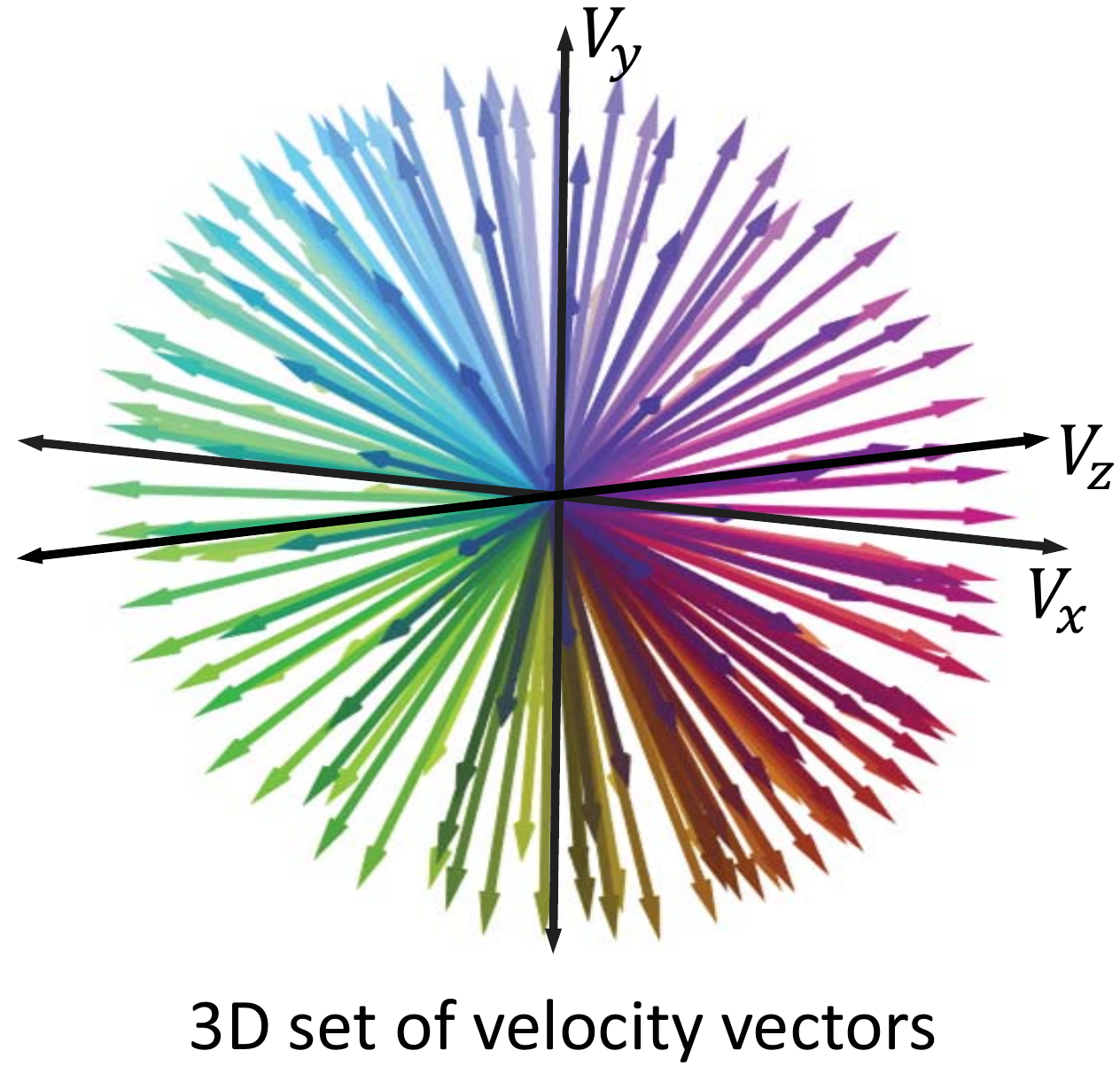} \hfill}
\caption{{\bf Relationship between scene texture, rank of the light field structure tensor, and the space of recoverable motions.} {\bf (Top)} Scene patches. {\bf (Middle)} Distribution of light field gradients; each dot represents the gradient  $\left( L_X, L_Y, L_Z \right)$ computed at one location in a light field window. The covariance of the gradients is represented by ellipsoids whose principal axes are proportional to the three eigenvalues $\lambda_1, \lambda_2, \lambda_3$ of the structure tensor. {\bf (Bottom)} Set of recoverable motion vectors. {\bf (Left)} For a light field window corresponding to a smooth patch, the gradients $\left( L_X, L_Y, L_Z \right)$ are approximately zero, and concentrated around the origin in the gradient space. The rank of the structure tensor is $0$, implying that no motion vector can be recovered reliably. {\bf (Center)} For a patch with a single edge, non-zero gradients are distributed approximately along a plane in the gradient space, resulting in a rank $2$ structure tensor ($1$-D null space). As a result, a 2D family of motions (orthogonal to the edge) can be recovered. {\bf (Right)} For a patch with 2D texture, non-zero gradients are distributed nearly isotropically in the gradient space. Therefore, structure tensor has rank $=3$. Thus, the entire space of 3D motions are recoverable. } \label{fig:structure_tensor} 
\end{figure*}

\subsection{What is the Space of Recoverable Motions?}
\label{sec:AnalysisLocal} 
In the previous section, we discussed that it is impossible to recover the complete 3D motion vector from a single ray flow equation. A natural question to ask is: what is the space of recoverable motions with the additional local constancy constraint? Intuitively it depends on the local structure of the light field. 
For example, if the local window corresponds to a textureless scene, then no motion is recoverable. 
One way to address this question is by understanding the properties of the $3 \times 3$ symmetric matrix $\mathbf{S} = \mathbf{A}^T\mathbf{A}$.
\begin{equation}
\mathbf{S} = \begin{bmatrix}
\sum\displaylimits_{\mathbf{i}=1}^\mathbf{n} L_{Xi}^2 & \sum\displaylimits_{\mathbf{i}=1}^\mathbf{n} L_{Xi} L_{Yi} & \sum\displaylimits_{\mathbf{i}=1}^\mathbf{n} L_{Xi} L_{Zi} \\
\sum\displaylimits_{\mathbf{i}=1}^\mathbf{n} L_{Yi} L_{Xi} & \sum\displaylimits_{\mathbf{i}=1}^\mathbf{n} L_{Yi}^2 & \sum\displaylimits_{\mathbf{i}=1}^\mathbf{n} L_{Yi} L_{Zi} \\
\sum\displaylimits_{\mathbf{i}=1}^\mathbf{n} L_{Zi} L_{Xi} & \sum\displaylimits_{\mathbf{i}=1}^\mathbf{n} L_{Zi} L_{Yi} & \sum\displaylimits_{\mathbf{i}=1}^\mathbf{n} L_{Zi}^2 \\
\end{bmatrix},
\label{eq:structure_tensor}\end{equation}
where $L_{*i}$ is short for $L_*(\vec{x}_i)$.
We define $\mathbf{S}$ as the \emph{light field structure tensor}; it encodes the local structure of the light field.\footnote{Structure tensors have been researched and defined differently in the light field community (\eg \citet{neumann2004hierarchy}). Here it is defined by the gradients w.r.t. the 3D motion and  is thus a $3\times 3$ matrix.}
To estimate motion using Eq.~\ref{eq:normal_equation}, $\mathbf{S}$ must be invertible.  Thus, the performance of the local method can be understood in terms of $rank(\mat{S})$.

\smallskip\noindent {\bf Result 2 (Rank of structure tensor)}
Structure tensor $\mathbf{S}$ has three possible ranks: $0$, $2$, and $3$ for a local 4D light field window. These correspond to scene patches with no texture (smooth regions), an edge, and 2D texture, respectively.

\smallskip
\noindent {\bf Intuition:} In the following, we provide an intuition for the above result by considering three cases. A detailed proof is given in the appendix.

\smallskip
\noindent \emph{Case 1: Smooth region.} In this case, $L_X = L_Y = L_Z = 0$ for all the locations in the light field window. Therefore, all the entries of the structure tensor (given in Eq.~\ref{eq:structure_tensor}) are zero, resulting in it being a rank $0$ matrix. All three eigenvalues $\lambda_1, \lambda_2, \lambda_3 = 0$, as shown in the left column of Figure~\ref{fig:structure_tensor}. As a result, it has a $3$-D null space, and no motion vector can be recovered reliably for this window. 

\smallskip
\noindent \emph{Case 2: Single step edge.} Without loss of generality, suppose the light field window corresponds to a fronto-parallel scene patch with a vertical edge, \ie $L_Y = 0$. The middle row of the structure tensor is all zeros, resulting in a rank $2$ matrix, with a $1$-D null space (only one eigenvalue $\lambda_3 = 0$). As a result, a 2D family of motions (motion orthogonal to the edge) can be recovered, as illustrated in the second column of Figure~\ref{fig:structure_tensor}.
 
\smallskip
\noindent \emph{Case 3: 2D texture.} All three derivatives are non-zero and independent. The structure tensor is full rank (rank $=3$) and the entire space of 3D motions are recoverable.

\medskip \noindent {\bf Comparisons with structure tensor for optical flow:} The structure tensor for 2D optical flow is a $2\times 2$ matrix and can have all possible ranks from $0$ to $2$ \citep{shi1994good}. \emph{For light fields, the structure tensor cannot have rank $1$.} This is because even a 4D window with a single step edge results in a rank $2$ structure tensor.\footnote{Although the structure tensor theoretically has rank 2, the ratio $\frac{\lambda_1}{\lambda_2}$ of the largest and second largest eigenvalues can be large. This is because the eigenvalue corresponding to $Z$ motion depends on the range of $(u,v)$ coordinates, which is limited by the size of the light field window. Therefore, a sufficiently large window size is required for motion recovery.} For more conceptual comparisons between optical flow and ray flow, please refer to Table~\ref{tab:comparison}. 

\medskip\noindent \textbf{Dependence on camera parameters.} Besides scene texture and light field structure, the imaging parameters of the light field camera also influences the performance of ray flow methods. Using the ray flow equation requires computing angular light field gradients ($L_X$ and $L_Y$), whose accuracy depends on the angular resolution of the light field camera. Most off-the-shelf light field cameras have a relatively low angular resolution (\eg $15 \times 15$ for Lytro Illum), resulting in aliasing \citep{neumann2003polydioptric}. To mitigate aliasing, we apply  Gaussian pre-filtering before computing the gradients. Another important parameter is the aperture size, which limits the range of recoverable motion. This is because ray flow changes the $(x,y)$ coordinates of the ray. When the motion is too large, most of the rays will escape the aperture and the motion cannot be recovered (see Fig.~\ref{fig:RayFlow}). See Section~\ref{sec:results_camparam} for a detailed discussion on the effects of various camera parameters.

\subsection{Enhanced Local Methods}
Our analysis so far assumes small (differential) scene motion. If the inter-frame scene motion is large, then the simple linear ray flow equation is not valid. Another way to relate the scene motion and the resulting change in the captured light field is to define a warp function on the light field, which describes the change in coordinates $\mathbf{x} = (x, y, u, v)$ of a light ray due to scene motion $\mathbf{V}$ (Eq.~\ref{eq:Representations}): 
\begin{equation}\label{eq:warp}
\vec{w}(\mathbf{x},\mathbf{V}) = (x+V_X-\frac{u}{\Gamma}V_Z,y+V_Y-\frac{v}{\Gamma}V_Z,u,v) \,.
\end{equation} 
Then, the local method can be formulated as a local light field registration problem:
\begin{equation}\label{eq:local_registration}
\min_{\mathbf{V}} \sum_{\vec{x_i}\in\mathscr{N}(\vec{x_c})}(L_0(\mathbf{x_i})-L_1(\vec{w}(\mathbf{x_i},\mathbf{V})))^2 \,.
\end{equation}
The method described by Eq.~\ref{eq:normal_equation} is \emph{the same} as locally linearizing Eq.~\ref{eq:local_registration}. Using this formulation, we develop an enhanced local method where the motion vector $\vec{V}$ is solved over a light field pyramid for dealing with large (non-differential) scene motions.

\section{Global `Horn-Schunck' Ray Flow} \label{sec:global}
The local constancy assumption made by the local ray-flow methods is too restrictive when dealing with non-rigid motion. In this section, we propose a family of global ray flow methods that are inspired by global `Horn-Schunck' optical flow \citep{horn1981determining}. The basic, less limiting assumption is that the 3D flow field varies smoothly over the scene. Therefore, we regularize the flow computation by introducing a smoothness term that penalizes large variations of $\vec{V}$ and minimize a global functional:
\begin{equation}
E(\mathbf{V})= E_D (\mathbf{V}) + E_S (\mathbf{V}) ,\qquad \textrm{where}
\label{eq:GlobalRayFlow}\end{equation}
\begin{equation*}
E_D (\mathbf{V})  =  \int_{\Omega}\left(L_X V_X + L_Y V_Y + L_Z V_Z + L_t \right)^2 dx\,dy\,du\,dv \;,
\end{equation*}
\begin{equation*}
E_S (\mathbf{V})  =  \int_{\Omega} \left( \lambda|\nabla V_X|^2+\lambda|\nabla V_Y|^2+\lambda_Z|\nabla V_Z|^2 \right) dx\,dy\,du\,dv \;.
\end{equation*} 

\noindent Note that $\Omega$ is the 4D light field domain, and $\nabla p$ is the 4D gradient of a scalar field $p$: $\nabla p = (\parder{p}{x},\parder{p}{y},\parder{p}{u},\parder{p}{v})$. Since the computation of X/Y flow and Z flow are asymmetric, we use different weights for the X/Y and Z smoothness terms. In practice we use $\lambda=8$ and $\lambda_Z=1$. $E (\mathbf{V})$ is a convex functional, and its minimum can be found by the Euler-Lagrange equations. See the appendix for details.

\medskip\noindent \textbf{Enhanced global methods.}
The quadratic penalty functions used in the basic global ray flow method (Eq.~\ref{eq:GlobalRayFlow}) penalizes flow discontinuities, leading to over-smoothing around motion boundaries. In the optical flow community \citep{odobez1995robust,black1996robust,brox2004high}, it has been shown that robust penalty functions perform significantly better around motion discontinuities. Based on this, we developed an enhanced global method that uses the generalized Charbonnier function $\rho(x)=(x^2+\epsilon^2)^a$ with $a=0.45$ as suggested in \citet{sun2010secrets}.

\begin{figure*}[t!]
  \begin{center}
  \includegraphics[width=\linewidth]{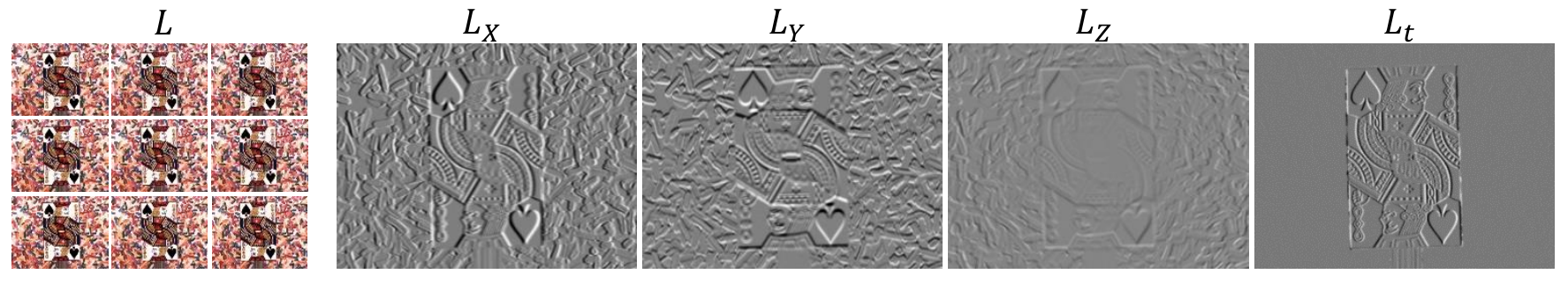}
  \end{center}
\caption{\textbf{Measured light field gradients.} Light field for an example scene (a card moving in the X-Z plane in front of a static background) is shown as a $3 \times 3$ subset of sub-aperture images (left). Light field gradients are only shown for the central sub-aperture. {\bf Zoom in for details.}}
\label{fig:gradients}
\end{figure*}

\section{Structure-Aware Global Ray Flow}
\label{sec:combined}
The ray flow methods considered so far treat the motion of each light ray separately. However, a light field camera captures multiple rays from the same scene point, all of which share the same motion. Can we exploit this constraint to further improve the performance of ray flow based motion recovery methods? Consider a ray with coordinates $(x,y,u,v)$, coming from a scene point $S = (X,Y,Z)$. The coordinates of all the rays coming from $S$ form a 2D plane $\mathscr{P}(u,v)$ \citep{srinivasan2015oriented,dansereau2011plenoptic,johannsen2015linear} in the 4D light-field:
\begin{equation}
\begin{split}
\mathscr{P}(u,v) = \{(x_i,y_i,u_i,v_i) \mid & u_i = u-\alpha(x_i-x),\\
& v_i = v-\alpha(y_i-y)\}, \label{eq:LFPlane}
\end{split}
\end{equation}
\noindent where the parameter $\alpha=\frac{\Gamma}{Z}$ is the disparity between sub-aperture images, and is a function of the depth $Z$ of $S$. All these rays share the same flow vector $\vec{V}=(V_X,V_Y,V_Z)$. Therefore, we can estimate $\vec{V}$ by minimizing the following function: 
\begin{equation}\label{eq:LocalPlaneEnergy}
\min_{\vec{V}}\sum_{\vec{x}_i\in \mathscr{P}(u,v)}(L_{Xi}V_X+L_{Yi}V_Y+L_{Zi}V_Z+L_{ti})^2.
\end{equation}
Given the parameter $\alpha$ (which can be determined using light-field based depth estimation \citep{wanner2014variational}), this function can be minimized similarly as the local method (Section~\ref{sec:local}), which assumes constancy of ray motion in a local 4D ray neighborhood $\mathscr{N}(u,v)$. While the local constancy assumption is only approximate, the constancy of motion over the 2D plane described in Eq.~\ref{eq:LFPlane} is an \emph{exact constraint}, resulting in better performance. Moreover, in order to further regularize the problem, we can leverage the global smoothness of motion assumption used in global methods in Section~\ref{sec:global}. Based on these observations, we propose a \emph{structure-aware global} (SAG) ray flow method, whose formulation is inspired by the combined local-global optical flow \citep{bruhn2005lucas}. The data term is given by minimizing the local term (Eq.~\ref{eq:LocalPlaneEnergy}) for each ray \emph{in the central view} $\Omega_c$:
\begin{equation}
E_D(\vec{V}) = \int_{\Omega_c}\sum_{\vec{x}_i\in \mathscr{P}(u,v)}(L_{Xi}V_X+L_{Yi}V_Y+L_{Zi}V_Z+L_{ti})^2du\,dv\,.
\end{equation}
This local data term is combined with a global smoothness term defined on $\Omega_c$.
\begin{equation}
E_S (\mathbf{V}) = \int_{\Omega_c} \left( \lambda|\nabla V_X|^2+\lambda|\nabla V_Y|^2+\lambda_Z|\nabla V_Z|^2 \right) du\,dv\,.
\end{equation}
This formulation estimates motion only for the 2D central view $\Omega_c$ while utilizing the information from the whole light field, thereby simultaneously achieving computational efficiency and high accuracy. Furthermore, by adopting the enhancements of local and global methods, the SAG method outperforms individual local and global methods (see Section~\ref{sec:results_quant}) (please refer to the Appendix for implementation details). Also notice that the SAG ray flow method uses the estimated depths only \emph{implicitly} as an additional constraint for regularization. Therefore, unlike previous methods \citep{heber2014scene,navarro2016variational,srinivasan2015oriented}, estimating depths accurately is not critical for recovering motion (see Section~\ref{sec:results_depth}). 


\section{Performance Analysis and Experimental Results} \label{sec:results}

For our experiments, we use a Lytro Illum camera, calibrated using a geometric calibration toolbox \citep{bok2017geometric}. We extract the central $9\times 9$ subaperture images, each of which has a spatial resolution of $552\times 383$. Figure \ref{fig:gradients} shows an example light field and the computed gradients. We compare our structure-aware global method (SAG) with the RGB-D scene flow method (PD-Flow) of \citet{jaimez2015primal} and light field scene flow method (called OLFW in this paper) of \citet{srinivasan2015oriented}. For a fair comparison, we use the same modality (light fields) for depth estimation in PD-Flow (depth estimated from light field is the depth channel input), using the same algorithm as in OLFW \citep{tao2013depth}. 
\textbf{Please refer to the supplementary video for a better visualization of the scene motion.}

\subsection{Quantitative Results for Controlled Experiments} \label{sec:results_quant}
\begin{figure*}[t!]
  \begin{center}
  \includegraphics[width=\linewidth]{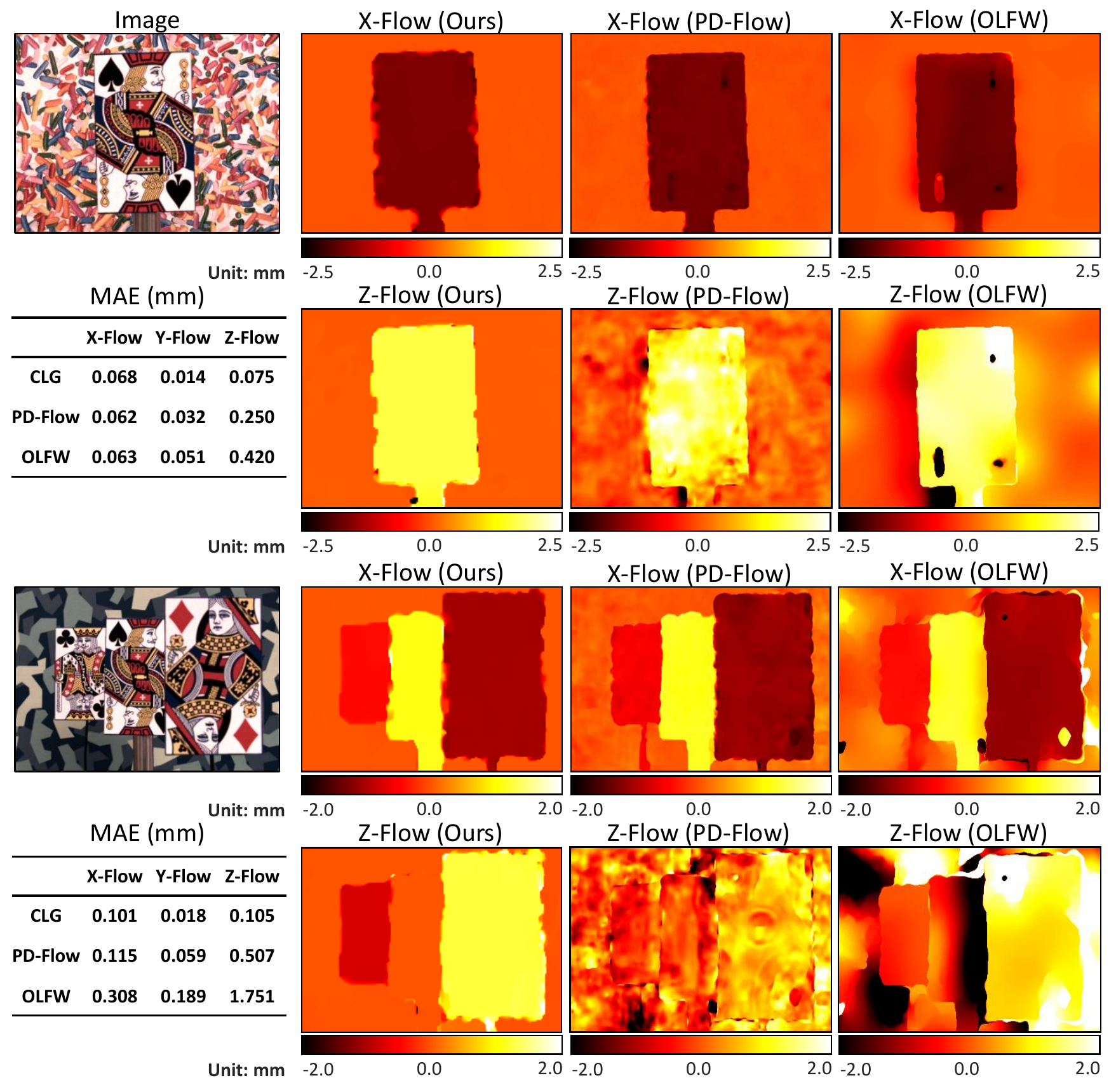}
  \end{center}
\caption{\textbf{Controlled experiments on a translation stage.} \textbf{(Top)} A single card moving diagonally. \textbf{(Bottom)} Three cards moving diagonally forward, laterally, and diagonally backward, respectively. Mean absolute error (MAE) for the three motion components are shown in the tables. While all methods recover the lateral motion relatively accurately, the proposed SAG ray-flow approach estimates the Z-motion more accurately than previous approaches. This is because previous approaches rely on, and are thus prone to errors in, depth estimation. In contrast, our approach estimates the motion directly from light-field gradients, thereby achieving high accuracy.}
\label{fig:exp_controlled}
\end{figure*}

Figure \ref{fig:exp_controlled} shows scene flow recovery \noindent results for a scene that is intentionally chosen to have simple geometry and sufficient texture to compare the baseline performance of the methods. The moving objects (playing cards) are mounted on controllable translation stages such that they can move in the X-Z plane with measured ground truth motion. Mean absolute error (MAE) for the three dimensions (ground truth Y-motion is zero) are computed and shown in the table. All three methods perform well for recovering the X-motion. However, PD-Flow and OLFW cannot recover the Z-motion reliably because errors in depth estimation are large compared to the millimeter-scale Z-motion. The proposed ray flow method estimates the Z-motion directly, thereby achieving higher accuracy.

\begin{figure}[t!]
  \begin{center}
  \includegraphics[width=\linewidth]{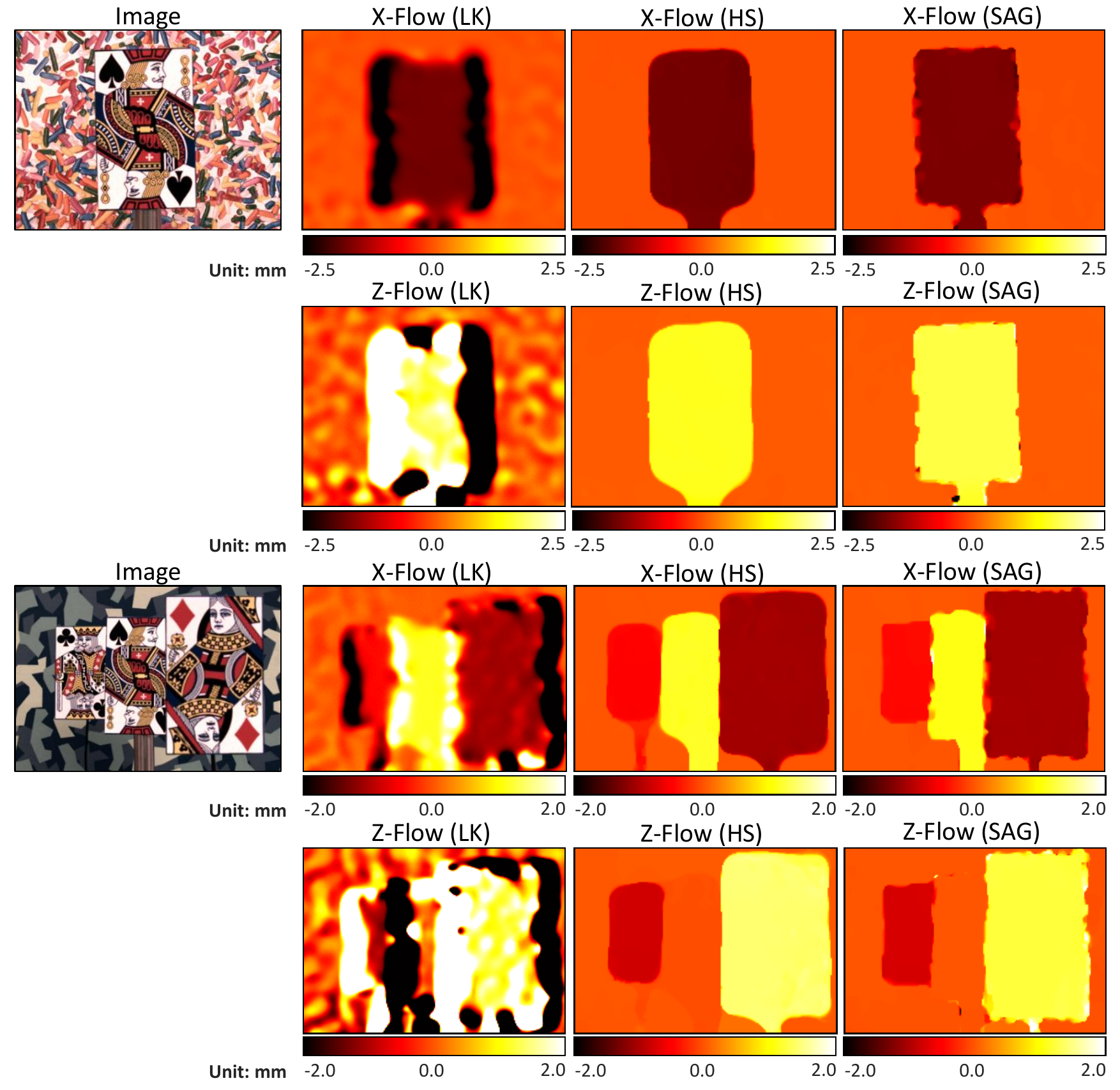}
  \end{center}
\caption{\textbf{Comparison of local (LK), global (HS) and structure-aware global (SAG) methods.} The local method can recover the motion in the center of the objects but fails to preserve the boundaries. The global method does much better at the boundaries but oversmoothed the motion field around the corners. The structure-aware global method estimates the motion most accurately.}
\label{fig:exp_comp_lg}
\end{figure}

\smallskip\noindent \textbf{Comparison between the local, global and structure-aware methods.}
We compare the performance of the local Lucas-Kanade (LK) method, global Horn-Schunck (HS) method and the structure-aware global method (SAG) using controlled experiments so that ground truth motion is available. Figure~\ref{fig:exp_comp_lg} shows the recovered motion using the three methods. The local method works relatively 
well at the center of the objects, but performs poorly at the boundaries. 
The global method produces better results around the boundaries, but the recovered motion boundary is oversmoothed at the corners. The structure-aware global method makes use of the estimated light field structure and recovers a sharper boundary. Overall, in all our experiments, the structure-aware global method achieves the best performance. 
In the rest of the paper, we will only show results and comparisons using only the structure-aware method.

\begin{figure*}[t!]
  \begin{center}
  \includegraphics[width=\linewidth]{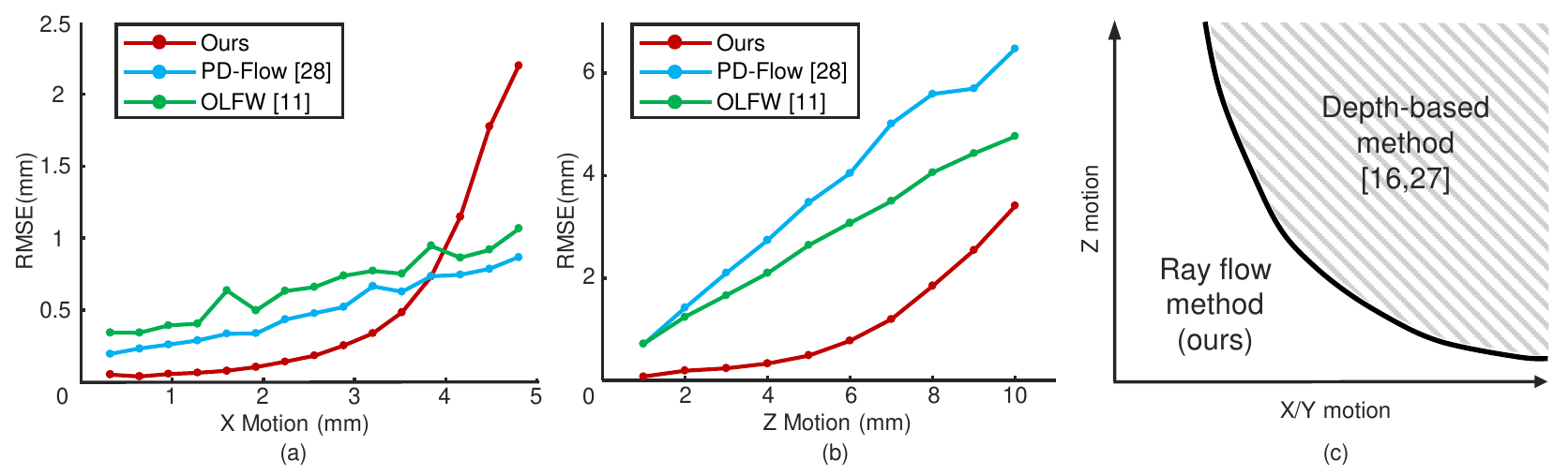}
  \end{center}
\caption{\textbf{Effect of the amount and kind of motion.} We use a single textured plane as the scene to exclude the effect of other factors (motion boundaries, occlusions). \textbf{(a)} For X-motion, the error of our method increases rapidly when the motion is larger than 3.5mm, while PD-Flow and OLFW degrade gracefully. \textbf{(b)} For Z-motion, our method outperforms previous methods since it does not rely on accurate depth estimates. \textbf{(c)} This plot qualitatively shows the method best suited for estimating different amount and kind of motion. While previous approaches can reliably measure large motions, the proposed method is better suited for small, especially axial, motions.}
\label{fig:exp_performance}
\end{figure*}

\smallskip\noindent \textbf{Dependency of the performance on the amount and kind of motion.}
We mount a textured plastic sheet on the translation stage and move it either laterally (X-motion) or axially (Z-motion). Figures \ref{fig:exp_performance}(a)(b) plot the RMSE of the estimated motion, against the amount of motion. The proposed method achieves higher precision for small motion.
However, its accuracy decreases as the amount of motion increases. This is because of the limit imposed by the aperture size, as discussed in Section \ref{sec:AnalysisLocal}. On the other hand, previous depth-based methods \citep{srinivasan2015oriented} can recover motion over a large range, albeit with lower precision. This complementary set of capabilities of our method and previous methods are shown qualitatively in Figure \ref{fig:exp_performance}(c). Although for the rest of the paper we focus on showing our methods' capability in recovering small motion (\eg for applications in finger gesture and facial expression recognition), previous approaches \citep{srinivasan2015oriented} may perform better for measuring large scale motion, such as gait recognition. 

\subsection{Performance Analysis using Simulated Images}\label{sec:results_camparam}
In this section, we analyze the dependence of the performance of the proposed methods on various scene and camera parameters. We only show the results for the structure-aware global method, but we expect similar results for the other methods. We use simulated images so that we can evaluate the performance over a wide range of parameter settings. We implement a light field rendering engine based on POV-Ray, which not only models the geometric distribution of light rays captured by a light field camera, but also generates physically realistic noise on the images. We use the affine noise model which includes both photon noise and read noise \citep{hasinoff2010noise}.

\begin{figure}[t]
  \begin{center}
  \includegraphics[width=\linewidth]{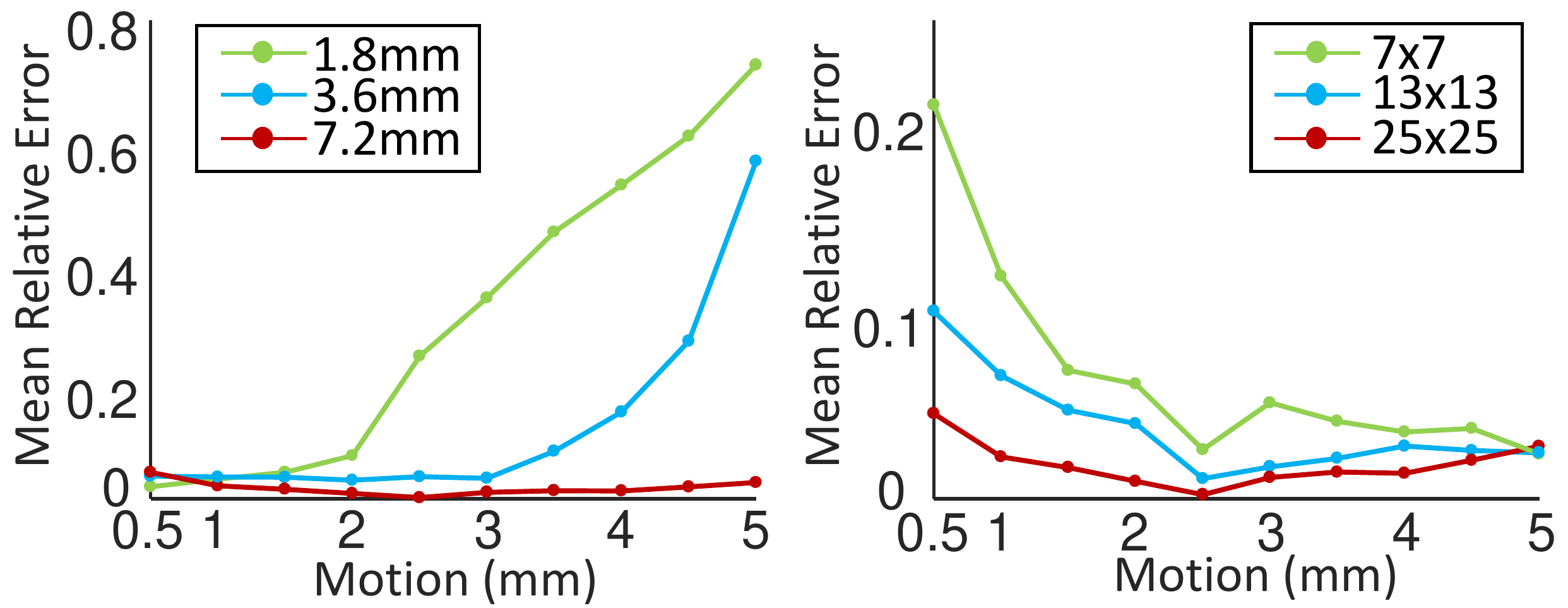}
  \end{center}
\caption{\textbf{Dependence on camera parameters.} We plot the mean relative error for different amount of motion. \textbf{(Left) Effect of camera aperture.} Results for three different effective aperture sizes: 1.8mm, 3.6mm and 7.2mm. The maximum recoverable motion increases as the aperture size increases. \textbf{(Right) Effect of angular resolution.} Results for three different angular resolutions: 7x7, 13x13 and 25x25. When the motion is small, error due to low angular resolution is dominant. When motion is large, the difference between different resolution is not significant.}
\label{fig:exp_aperture}
\end{figure}

\smallskip\noindent \textbf{Aperture Size.} The aperture size of a light field camera defines the range of $x$, $y$ coordinates in the captured light fields.
We reproduce the single plane experiment in the previous section in our simulation engine: We synthesize a sequence of images with different amount of motion along the X-axis. Figure \ref{fig:exp_aperture} (Left) plots the mean relative error against the amount of motion, for three different effective aperture sizes: 1.8mm, 3.6mm and 7.2mm. The error increases as the amount of motion increases. This is because scene motion causes a ray to translate in the $x$, $y$ dimensions (see Eq.~\ref{eq:DerivRelat} and Fig.~\ref{fig:RayFlow}). As a result, the maximum motion that can be detected is limited by the aperture size. The maximum recoverable motion increases as the aperture size increases.

\smallskip\noindent \textbf{Angular Resolution.}
The accuracy of motion estimation is also determined by the angular resolution of the light field camera. Figure \ref{fig:exp_aperture} (Right) shows the results for the single-plane sequence for three different angular resolutions: 7x7, 13x13 and 25x25. When the motion is small, higher angular resolution results in much smaller error. When motion is large, the difference between different resolution is not significant. This is because, since scene motion is related to the light field gradients along $x$ and $y$ coordinates (angular dimensions), the resolution of motion recovery depends on the angular resolution. For small motion, the error due to low resolution dominates the relative error. For large motion this error is negligible, and other factors (noise, convergence to local minimum, etc.) come into effect.

These experiments demonstrate that the aperture size has a stronger influence on the performance for larger motion, whereas the angular resolution plays a more important part for small motions. A detailed theoretical analysis of the influence of various imaging parameters on the motion estimation performance is an important direction for future work. 

\begin{figure}[t]
  \begin{center}
  \includegraphics[width=0.49\linewidth]{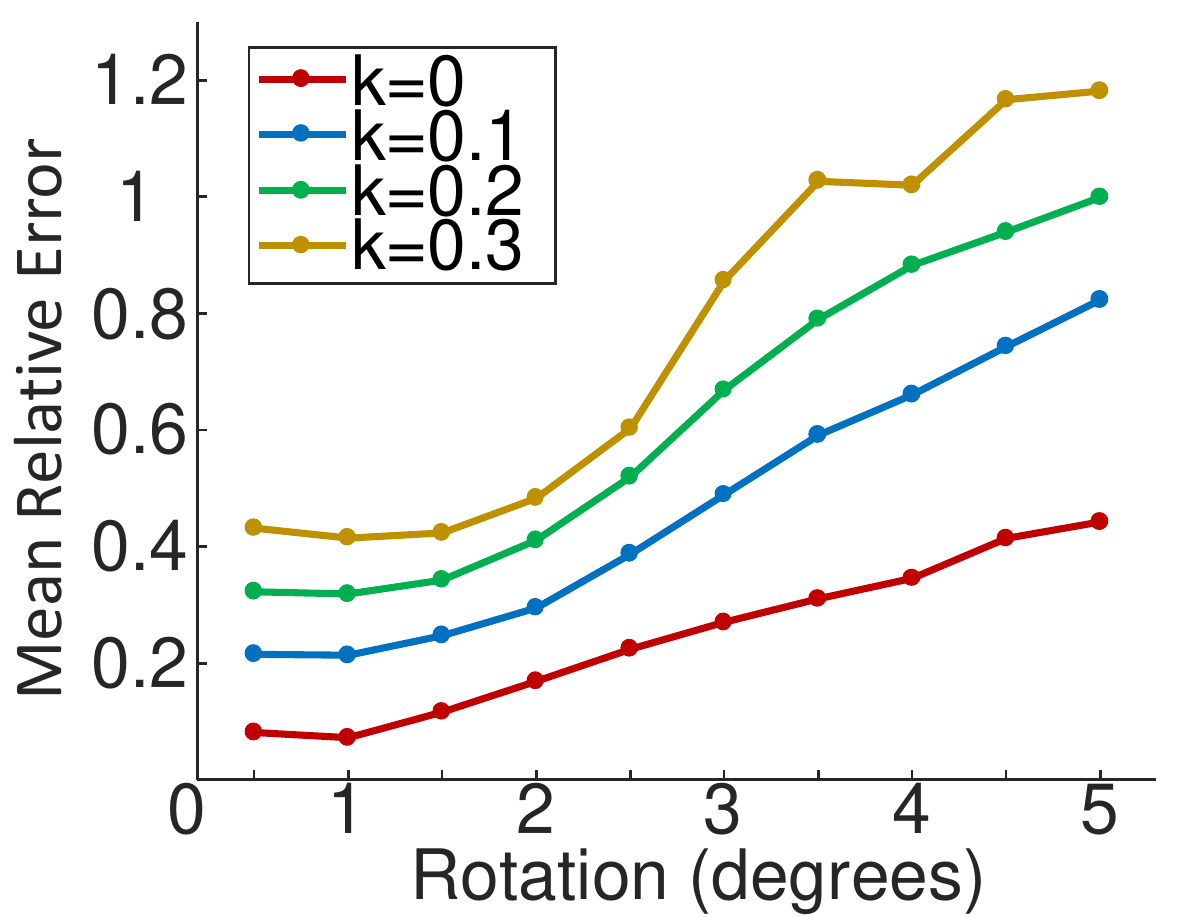}
  \includegraphics[width=0.49\linewidth]{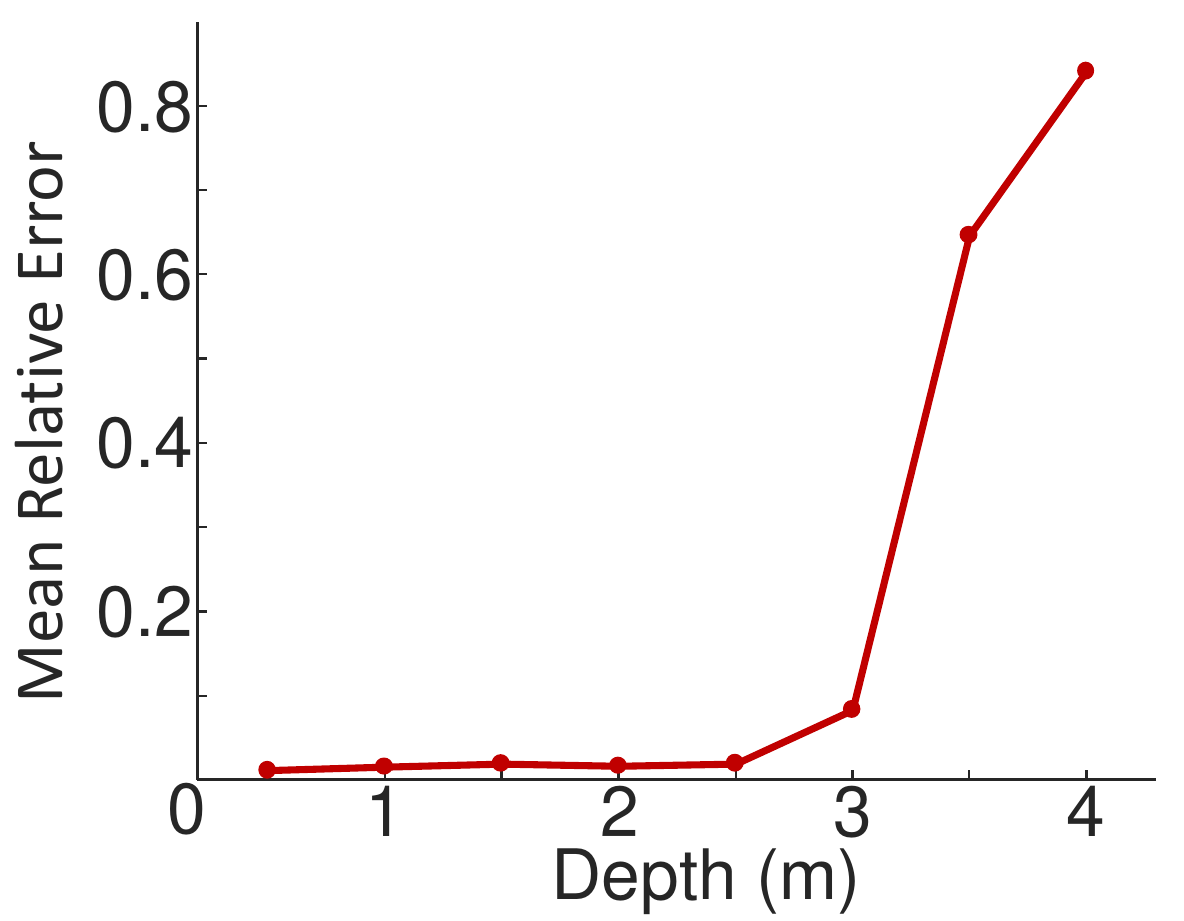}
  \end{center}
\caption{\textbf{Dependence on scene and motion parameters.} \textbf{(Left) Dependence on amount of rotation and BRDF.} The proposed method works for smaller rotation. The error is high for large rotation because the motion of the scene points on the far end of the plane exceeds the maximum recoverable range. As the specularity ($k$) of the surface increases, the error also increases. \textbf{(Right) Dependence on scene depth.} The error of motion estimation increases abruptly after a certain depth threshold. This because the observed light field is convolved with a 4D point spread function, which makes the gradient computation unreliable for larger depths.}
\label{fig:exp_rotation_depth}
\end{figure}

\smallskip\noindent\textbf{Dependence on the amount of rotation and BRDF.}
\label{sec:results_rotation}
Figure~\ref{fig:exp_rotation_depth} (Left) plots the accuracy of our algorithm for different amount of rotation and different BRDFs. The scene consists of a fronto-parallel plane rotating around the y-axis, which is illuminated by a single point light source. We use the Phong reflection model \citep{phong_illumination_1975} to render the scene. The ambient and diffuse coefficients are kept the same for all experiments. The specular coefficient $k$ varies from 0 to 0.3. 

For the perfectly Lambertian case ($k=0$), the method works well for small rotations, despite the theoretical model accounting only for translation. As the amount of rotation increases, the error goes up since the motion of the scene points that are farther away from the rotation axis exceeds the maximum recoverable motion range. For nonzero $k$, the performance of the method is affected by the specular highlights because the brightness constancy assumption no longer holds. The error is higher for stronger specular highlights (larger $k$).

\smallskip\noindent\textbf{Dependence on scene depth.}
Figure~\ref{fig:exp_rotation_depth} (Right) plots the accuracy of the proposed method at different depths. The motion of the plane is kept the same for all the experiments. The errors remain small for small depths, and then rise abruptly after a certain depth value. As mentioned in Section~\ref{sec:invar_depth}, this is because the captured light field is convolved with a 4D low-pass point spread function (PSF). As the depth gets larger, the high frequency information is lost, and the gradients cannot be computed reliably, which leads to erroneous motion estimation. This result shows that, similar to depth estimation from light fields, motion estimation from light fields is only possible within a certain depth range. This working depth range depends on the actual PSF of the light field camera.

\begin{figure}[t]
  \begin{center}
  \includegraphics[width=\linewidth]{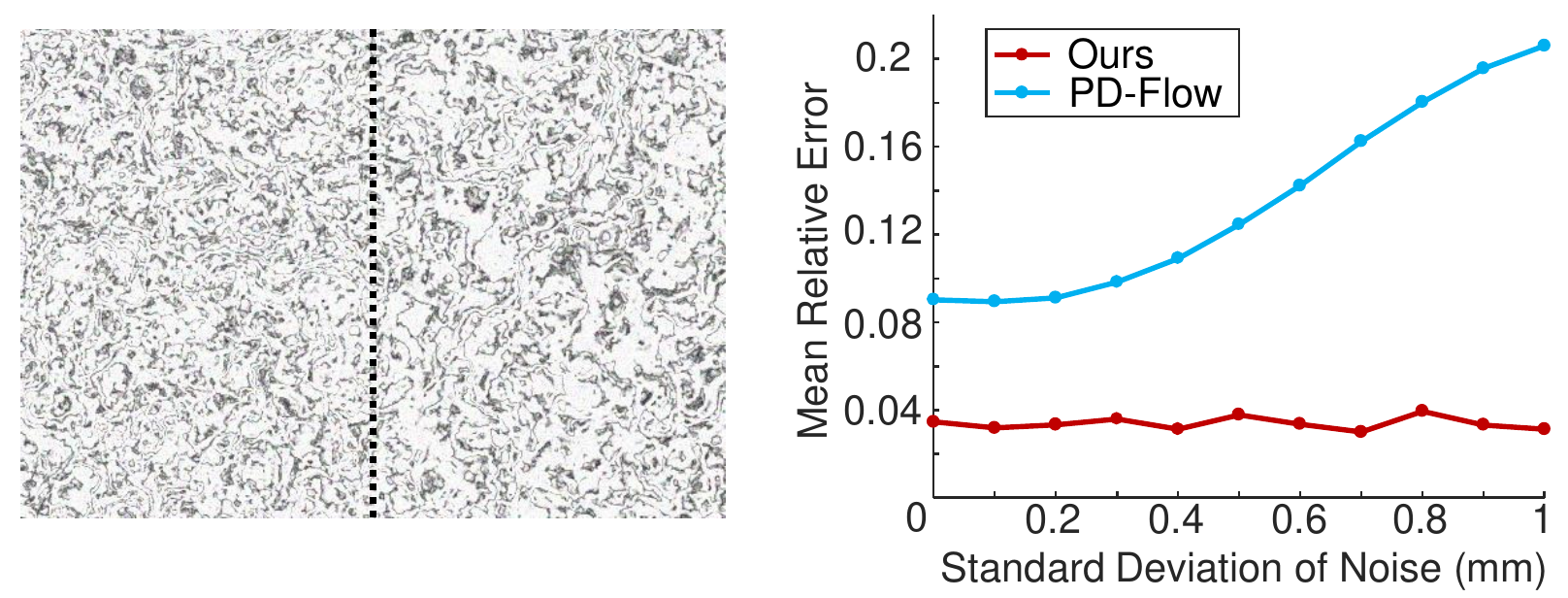}
  \end{center}
\caption{\textbf{Dependence on accuracy of depth estimation.}  \textbf{(Left)} The scene used for the simulation consists of two planes at different depths, moving in opposite directions. The boundary of the two planes is visualized as the dotted line. \textbf{(Right)} Mean relative error of the two methods against different amount of noise. The performance of PD-Flow degrades as the amount of noise increases, while the performance of SAG doesn't change much. This is because PD-Flow relies on explicitly computing change in depth, while SAG estimates Z-motion directly.}
\label{fig:exp_depth_accuracy}
\end{figure}

\begin{figure*}[t!]
  \begin{center}
  \includegraphics[width=\linewidth]{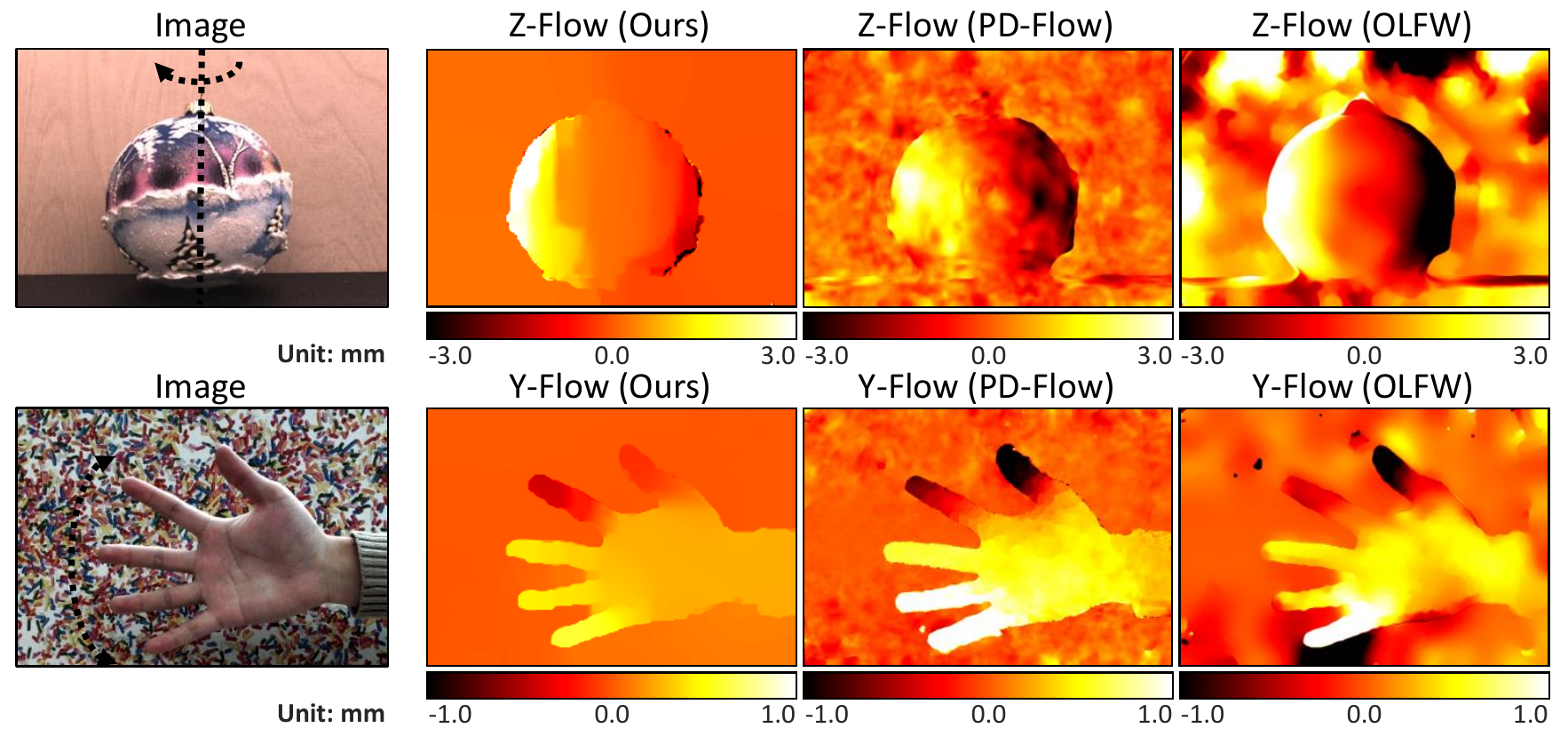}
  \end{center}
\caption{\textbf{Recovering non-planar and non-rigid motion.} \textbf{(Top)} A rotating spherical ornament. All methods can estimate the gradually changing Z-motion, but only our method recovers the background correctly. \textbf{(Bottom)} An expanding hand. The expansion is demonstrated by the different Y-motion of the fingers. }
\label{fig:exp_nonrigid}
\end{figure*}

\begin{figure*}[t!]
  \begin{center}
  \includegraphics[width=\linewidth]{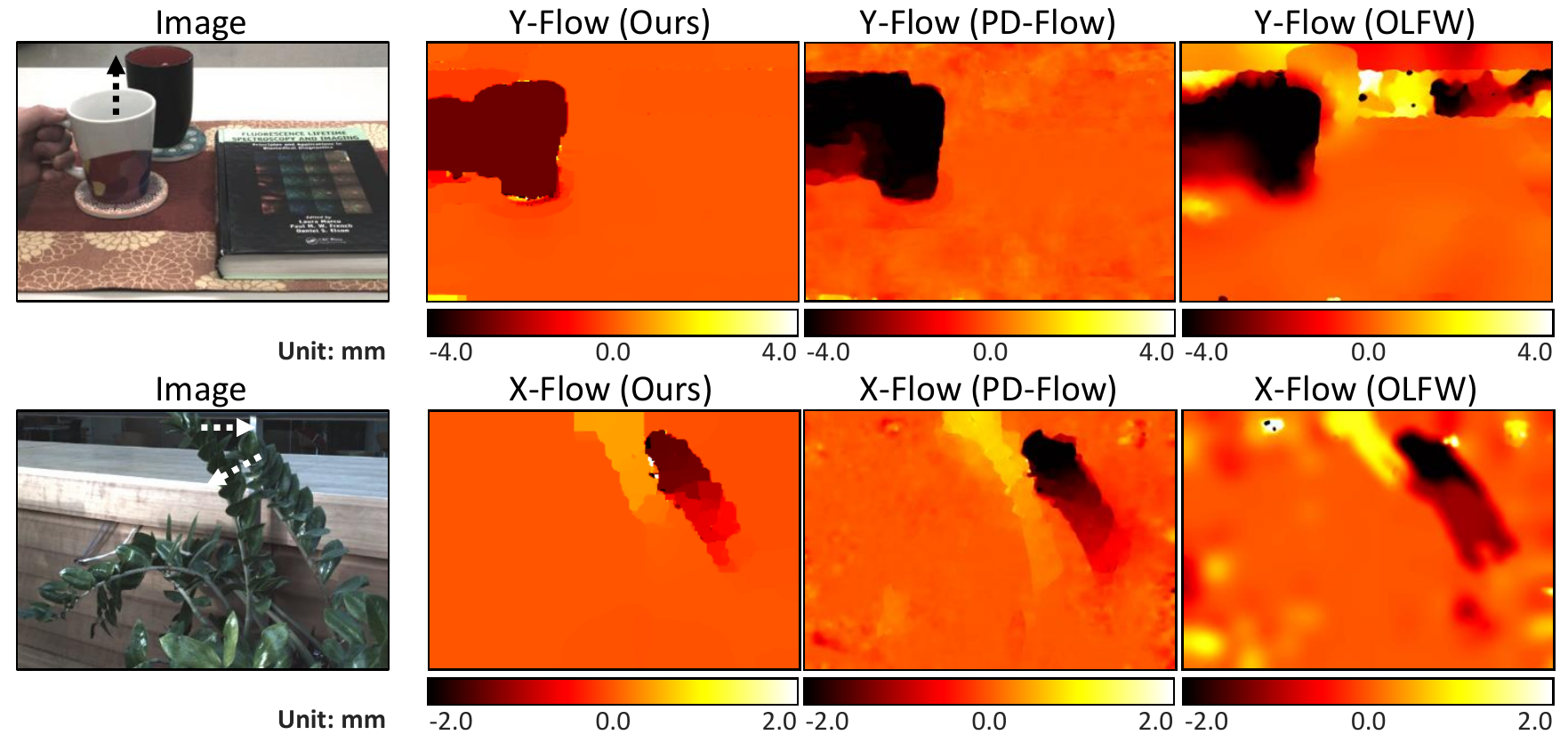}
  \end{center}
\caption{\textbf{Recovering motion in natural environments with occlusions.}  \textbf{(Top)} The mug on the left is picked up by a hand. Our method estimates the motion boundaries accurately. \textbf{(Bottom)} The top two vertical branches of the plant quiver in the wind. Our method can correctly compute the motion of the two complex-shaped branches.}
\label{fig:exp_clutter}
\end{figure*}

\smallskip\noindent \textbf{Dependence on Accuracy of Depth Estimation.}\label{sec:results_depth}
We compare the dependence of performance of different methods on the accuracy of depth estimation in a simulated scene. The scene consists of two textured, fronto-parallel planes at different depths (0.3m and 0.4m), moving for 2.24mm diagonally in opposite directions (Figure~\ref{fig:exp_depth_accuracy} (Left)). We use the ground truth depth with different amounts of Gaussian noise added as the depth input. 

Figure~\ref{fig:exp_depth_accuracy} (Right) plots the errors of a depth-based method (PD-Flow \citep{jaimez2015primal}) and the structure-aware ray flow method as a function of the amount of noise. The error of PD-Flow raises as the amount of noise increases, while the error of SAG remains almost constant. This is because the structure-aware ray flow method only uses \emph{estimated depth} to compute the 2D plane to establish the correspondences between rays. Since the noise in the depth estimate is \emph{small} compared to depth, the performance of SAG is not significantly affected. However, PD-Flow uses the \emph{change in estimated depth} to compute Z-motion. Since the noise is \emph{large} compared to the change in depth, the performance of PD-Flow depends heavily on the accuracy of depth estimation.

\begin{figure*}[t!]
  \begin{center}
  \includegraphics[width=\linewidth]{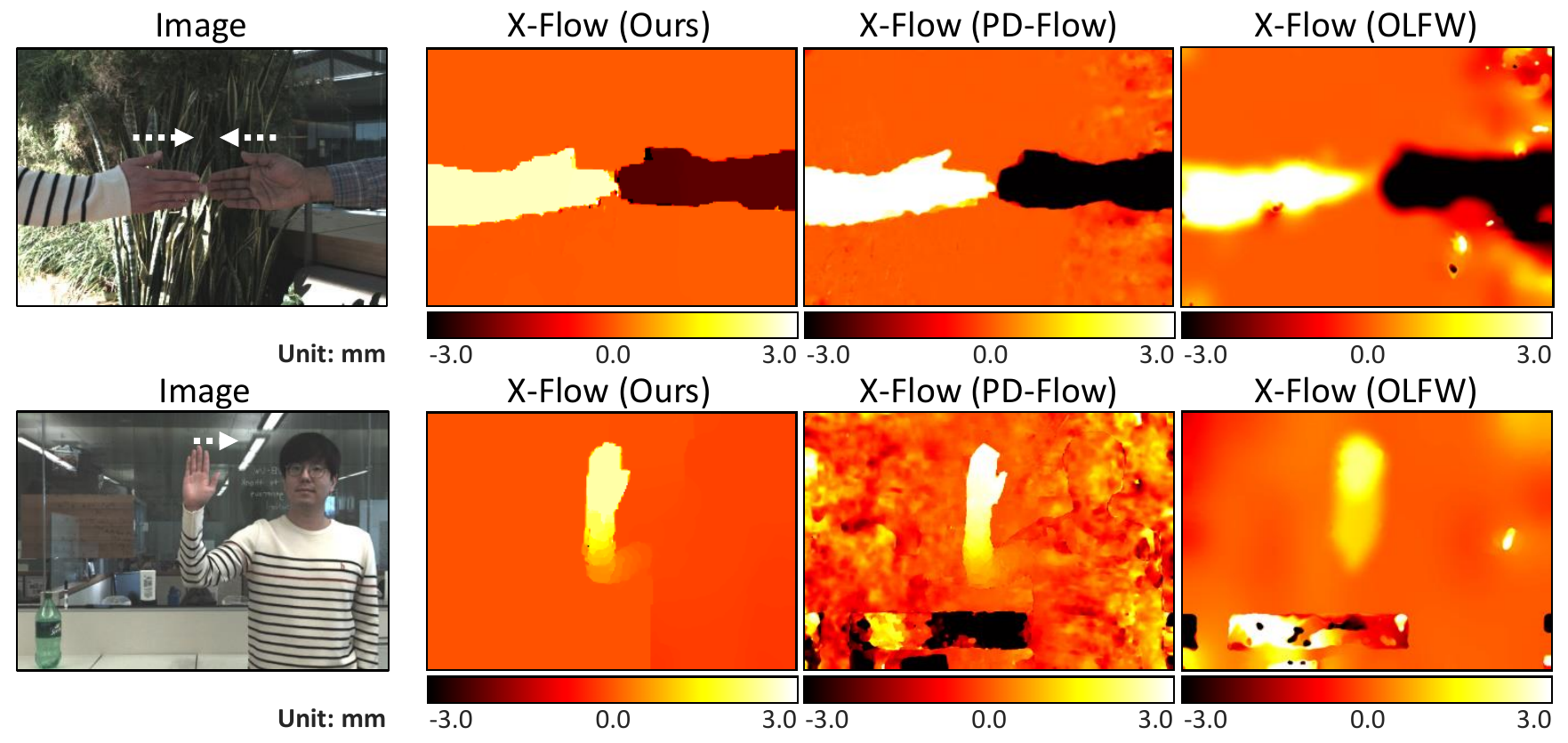}
  \end{center}
\caption{\textbf{Recovering human actions.} \textbf{(Top)} Handshaking. All the three methods compute the joining movements of the hands correctly, while our method preserves the hand boundary best. \textbf{(Bottom)} Waving hand. Our method correctly estimates the motion in spite of the reflections and textureless regions in the background, which is challenging for depth estimation algorithms.}
\label{fig:exp_human}
\end{figure*}

\begin{figure*}[t!]
  \begin{center}
  \includegraphics[width=\linewidth]{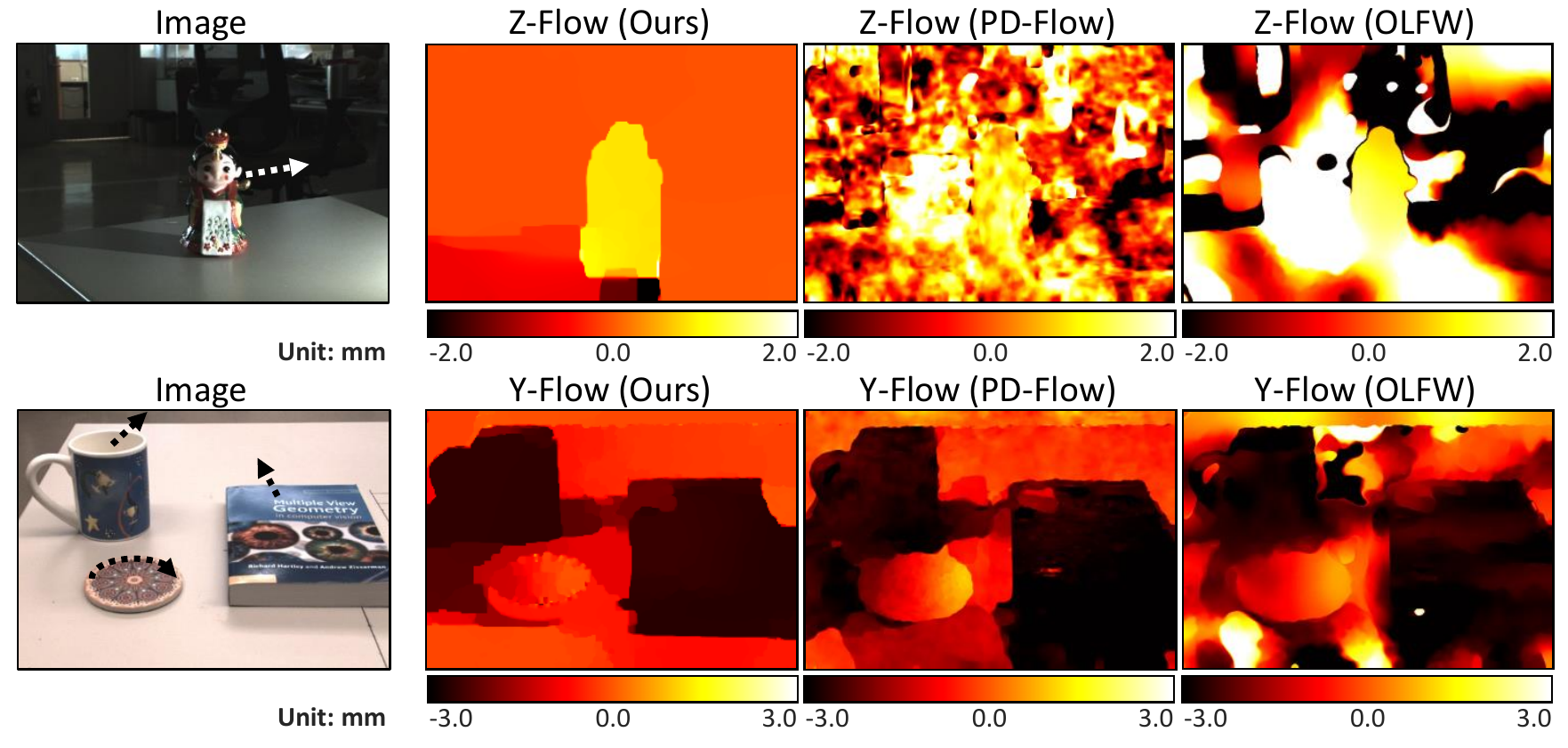}
  \end{center}
\caption{\textbf{Recovering motion under challenging lighting conditions.} \textbf{(Top)} A figurine moves under weak, directional lighting. Our method still preserves the overall shape of the object, although its reflection on the table is also regarded as moving. \textbf{(Bottom) Failure case:} a few objects move independently. Due to shadows and lack of texture in the background, boundaries of the objects are not distinguishable in the recovered motion field of all the three methods. }
\label{fig:exp_lighting}
\end{figure*}

\subsection{Qualitative Comparisons}
Figure \ref{fig:exp_nonrigid}--\ref{fig:exp_lighting} shows qualitative comparisons of the three methods for complex, non-rigid motion and in challenging natural environments. 
For each experiment we only show one component of the recovered 3D flow. \textbf{Please see the supplementary report for the full 3D flow visualization}. In all the examples, our method is able to estimate the complex, gradually changing motion fields and preserve the motion boundaries better than the other methods, especially for experiments involving small Z-motion, and where depth estimation is unreliable (\eg scenes with occlusions or reflections in the background).
In Fig.~\ref{fig:exp_lighting} (bottom) all three methods have difficulty in preserving the object boundaries due to shadows, which is a inherent drawback of the brightness constancy assumption.


\section{Limitations}\label{sec:Discussion}
\noindent {\bf Recoverable range of motion:} As discussed in Sections~\ref{sec:AnalysisLocal} and~\ref{sec:results}, the maximum recoverable amount of motion for ray flow methods is limited by the aperture size. A future research direction is to develop hybrid methods that combine the ray flow method and depth-based methods \citep{srinivasan2015oriented,jaimez2015primal} according to the amount and nature of scene motion. \smallskip

\noindent {\bf Running time:} Currently our methods are implemented via unoptimized MATLAB code, which takes approximately 10 minutes to compute scene flow between two frames. Further work includes reducing the computational complexity of the algorithm and implementing the algorithm efficiently (\eg on a GPU), for real-time applications.



\appendix
\section{Proof of Result 2}
{\bf Result 2 (Rank of structure tensor)}
Structure tensor $\mathbf{S}$ has three possible ranks: $0$, $2$, and $3$ for a local 4D light field window. These correspond to scene patches with no texture (smooth regions), an edge, and 2D texture, respectively.

\noindent \textbf{Proof.} 
We first show the three cases of rank $0$, $2$ and $3$, and then we prove that the structure tensor cannot be rank $1$. 

Since $rank(\mat{S})=rank(\mat{A}^T\mat{A})=rank(\mat{A})$, we only need to look at the rank of the $n\times 3$ matrix $\mat{A}$:
\begin{equation}\mat{A}=\begin{bmatrix}\mat{A}_1 & \mat{A}_2 & \mat{A}_3\end{bmatrix} =\begin{bmatrix}
L_{X1} & L_{Y1} & L_{Z1} \\
L_{X2} & L_{Y2} & L_{Z2} \\
\vdots & \vdots & \vdots \\
L_{Xn} & L_{Yn} & L_{Zn} \\
\end{bmatrix}\,.
\end{equation}

\noindent \emph{Case 1: Smooth region.} In this case, $L_X = L_Y = L_Z = 0$ for all the locations in the light field window. Therefore, all the entries of $\mat{A}$ are zero, resulting in a rank $0$ structure tensor. All three eigenvalues $\lambda_1, \lambda_2, \lambda_3 = 0$. As a result, it has a $3$-D null space, and no motion vector can be recovered reliably. \smallskip

\noindent \emph{Case 2: Single step edge.} Without loss of generality, suppose the light field window corresponds to a fronto-parallel scene patch with a vertical edge, \ie $L_Y = 0$ everywhere and thus $\mat{A}_2=\vec{0}$. 

Consider a point $P$ on the edge. Consider two rays from $P$ that are captured on two horizontally-separated sub-aperture images indexed by $(x_a,y)$ and $(x_b,y)$. Let the coordinates of the two rays be $(x_a,y,u_a,v_a)$ and $(x_b,y,u_b,v_b)$, and let the light field gradients at these two rays be $(L_{Xa},L_{Ya},L_{Za})$ and $(L_{Xb},L_{Yb},L_{Zb})$. Note that these two gradients are two rows 
in $\mat{A}$. 
Recall that
\begin{equation}
L_{Z}=-\frac{u}{\Gamma}L_{X}-\frac{v}{\Gamma}L_{Y} \label{eq:lz}\,.
\end{equation}
Since $L_{Ya}=0$ and $L_{Yb}=0$, we have,
\begin{equation}
L_{Za}=-\frac{u_a}{\Gamma}L_{Xa},\; L_{Zb}=-\frac{u_b}{\Gamma}L_{Xb}\,.
\end{equation}
Next, suppose there exists $k\neq 0$ such that $\mat{A}_3=k\mat{A}_1$. This implies \begin{equation}\label{eq:klx}
kL_{Xa}=-\frac{u_a}{\Gamma}L_{Xa},\;kL_{Xb}=-\frac{u_b}{\Gamma}L_{Xb}\,.
\end{equation}
By eliminating $k$ from Eq.~\ref{eq:klx}, we get $u_a=u_b$. However, since the scene point has a finite depth, the disparity $u_a - u_b \neq 0$. Therefore, $u_a \neq u_b$. This contradiction means such $k$ doesn't exist and $\mat{A}_1$ and $\mat{A}_3$ are linearly independent, which means the rank of $\mat{A}$ (and $\mat{S}$) is 2. As a result $\mat{S}$ has a $1$-D null space (only one eigenvalue $\lambda_3 = 0$) and a 2D family of motions (motion orthogonal to the edge) can be recovered. \smallskip

\noindent \emph{Case 3: 2D texture.} In general $\mat{A}_1$, $\mat{A}_2$ and $\mat{A}_3$ are nonzero and independent. The structure tensor is full rank (rank $=3$) and the entire space of 3D motions are recoverable. \smallskip

Now we show that the rank cannot be 1.

\noindent (\emph{Proof by contradiction}) Assume there exists a 4D patch such that its corresponding matrix $\mat{A}$ is rank 1. 

First $\mat{A}_1$ and $\mat{A}_2$ cannot both be zero. If we assume they are both zero then according to Eq.~\ref{eq:lz}
all entries in $\mat{A}_3$ will also be zero, which results in a rank 0 matrix. Therefore $\mat{A}_1\neq \vec{0}$ or $\mat{A}_2\neq \vec{0}$.

Without loss of generality, assume $\mat{A}_1 \neq 0$. Since $\mat{A}$ is rank 1, there exists $k,l\in\mathbb{R}$ such that
\begin{equation} \mat{A}_2 = k\mat{A}_1\,, \label{eq:multipleA}\end{equation}
\begin{equation} \mat{A}_3 = l\mat{A}_1\,. \label{eq:multipleB}\end{equation}

Let us pick a ray $\vec{x_a}=(x_a,y_a,u_a,v_a)$ with light field gradient $(L_{Xa},L_{Ya},L_{Za})$ such that $L_{Xa}\neq 0$. Such $\vec{x_a}$ exists because $\mat{A}_1\neq 0$. Note that this ray is captured by the sub-aperture image indexed by $(x_a,y_a)$. Assume the scene point corresponding to $\vec{x_a}$ is observed in another sub-aperture image $(x_b,y_b)$ with $y_a=y_b$, in other words a sub-aperture image that is at the same horizontal line as $(x_a,y_a)$.  Denote the corresponding ray as $\vec{x_b}=(x_b,y_a,u_b,v_b)$ with light field gradient $(L_{Xb},L_{Yb},L_{Zb})$. $L_{Xb}$ is also nonzero.

From Eq.~\ref{eq:multipleA} we know that $L_{Ya}=kL_{Xa}$ and $L_{Yb}=kL_{Xb}$. According to Eq.~\ref{eq:lz} we have 
\begin{equation}
L_{Za}=-\frac{u_a}{\Gamma}L_{Xa}-\frac{v_a}{\Gamma}\cdot L_{Ya}=-\frac{u_a+kv_a}{\Gamma}L_{Xa}\,,
\label{eq:lza}\end{equation}
\begin{equation}
L_{Zb}=-\frac{u_b}{\Gamma}L_{Xb}-\frac{v_b}{\Gamma}\cdot L_{Yb}=-\frac{u_b+kv_b}{\Gamma}L_{Xb}\,.
\label{eq:lzb}\end{equation}

From Eq.\ref{eq:multipleB} we know that $L_{Za}=l L_{Xa}$, $L_{Zb}=l L_{Xb}$, and from 
Eq.\ref{eq:lza}--\ref{eq:lzb} we have
\begin{equation}
l=-\frac{u_a+kv_a}{\Gamma}=-\frac{u_b+kv_b}{\Gamma}\,. \label{eq:klrelation}
\end{equation}

However since $x_a\neq x_b$, $y_a=y_b$, we have $u_a\neq u_b$ and $v_a=v_b$ due to simple epipolar geometry. Therefore Eq.~\ref{eq:klrelation} cannot hold, which means our assumption is false, and 
$rank(\mat{A})$ cannot be 1. \qed
\section{Implementation Details}
\subsection{Global Method}
In Section~\ref{sec:global}, we introduced the global `Horn-Schunck' ray flow method, which solves for the 3D scene motion by minimizing a functional:
\begin{equation}
E(\mathbf{V})= \underbrace{E_D (\mathbf{V})}_{\textbf{Error term}} + \underbrace{E_S (\mathbf{V})}_{\textbf{Smoothness term}} , \ \ \textrm{where}
\end{equation}
\begin{eqnarray*}
E_D (\mathbf{V}) & = & \int_{\Omega}\left(L_X V_X + L_Y V_Y + L_Z V_Z + L_t \right)^2 dx\,dy\,du\,dv \,, \\
E_S (\mathbf{V}) & = & \int_{\Omega} \left( \lambda|\nabla V_X|^2+\lambda|\nabla V_Y|^2+\lambda_Z|\nabla V_Z|^2 \right) dx\,dy\,du\,dv \,.
\end{eqnarray*}
This is a convex functional and its minimum can be found by the Euler-Lagrange equations,
\begin{align}
L_X(L_XV_X+L_YV_Y+L_ZV_Z)-\lambda\Delta V_X&=-L_XL_t\,, \nonumber \\
L_Y(L_XV_X+L_YV_Y+L_ZV_Z)-\lambda\Delta V_Y&=-L_YL_t \,, \\
L_Z(L_XV_X+L_YV_Y+L_ZV_Z)-\lambda_Z\Delta V_Z&=-L_ZL_t\,. \nonumber
\end{align}
These equations can be discretized as a sparse linear system, and is solved using SOR (Successive Over-Relaxation).

\subsection{Structure-Aware Global Method}
In this section we discuss an enhanced version of the structure-aware global method, which adopts the enhancing techniques for the local and global method, as discussed in Section~\ref{sec:combined} of the main paper. 

\subsubsection{Data Term} 
The data term is defined as:
\begin{equation}
E_D (\vec{V})=\int_{\Omega_c}\sum_{\vec{x}_i\in \mathscr{P}(u,v)}h_i\rho_D((L_0(\vec{x_i})-L_1(\vec{w}(\vec{x_i},\mathbf{V})))^2)du\,dv\,,
\end{equation}
where $\mathscr{P}(u,v)$ is the 2D plane defined in Equation (11) in the main paper. 

\smallskip\noindent \textbf{Weighted 2D window.} Notice that each ray in the 2D plane is given a different weight $h_i$, which is given by
\begin{equation}
h_i=h(\vec{x_i},\vec{x_c})=h_g(\vec{x_i},\vec{x_c})\cdot h_o(\vec{x_i},\vec{x_c})\,,
\end{equation}
\begin{equation}
h_g(\vec{x_i},\vec{x_c}) = e^{-\frac{(x_i-x_c)^2+(y_i-y_c)^2+(u_i-u_c+\alpha(x_i-x_c))^2+(v_i-v_c+\alpha(y_i-y_c))^2}{\sigma_g^2}}\,,
\end{equation}
\begin{equation}
h_o(\vec{x_i},\vec{x_c}) = e^{-\frac{(d_{\alpha i}-d_{\alpha c})^2}{\sigma_o^2}}\,,
\end{equation}
where $\vec{x_c}$ denotes the center ray of the window. 
$d_\alpha=1/\alpha$ and is proportional to the actual depth of the scene point.

$h_g$ defines a Gaussian weight function that is based on the distance between $\vec{x_i}$ and $\vec{x_c}$ in the 2D plane. 
$h_o$ defines an occlusion weight by penalizing the difference in the estimated disparity $\alpha$ at $\vec{x_i}$ and $\vec{x_c}$. Notice that not all rays on $\mathscr{P}(u,v)$ corresponds to the same scene point as $\vec{x}_c$ because of occlusion. If the scene point corresponding to $\vec{x}_i$ occludes or is occluded by the scene point corresponding to $\vec{x}_c$, they will have a different $\alpha$ and thus a small value of $h_o$.

\subsubsection{Smoothness Term}
The smoothness term is defined as
\begin{equation}
\begin{split}
E_S(\vec{V})=\int_{\Omega}g(\vec{x})(\lambda\sum_{i=1}^2\rho_S(V_{X(i)}^2)
+\lambda\sum_{i=1}^2\rho_S(V_{Y(i)}^2)\\
+\lambda_Z\sum_{i=1}^2\rho_S(V_{Z(i)}^2))du\, dv\,,
\end{split}
\end{equation}
where $V_{X(i)}$ is short for $\parder{V_X}{u^{(i)}}$.
(For simplicity we denote $u,v$ as $u^{(1)},u^{(2)}$ respectively.) $g(\vec{x})$ is a weight function that varies across the light field. The error term $E_C(\vec{V})$ uses the warp function (Eq. 8) in the main paper.

\smallskip\noindent\textbf{Best practices from optical flow.} We choose the penalty function $\rho_D$, $\rho_S$ to be the generalized Charbonnier penalty function $\rho(x^2)=(x^2+\epsilon^2)^a$ with $a=0.45$ as suggested in \citet{sun2010secrets}.

\smallskip\noindent\textbf{Weight function for the regularization term.}
The weight function $g(\vec{x})$ consists of two parts, which is combined using a harmonic mean:
\begin{equation}
g(\vec{x}) = \frac{g_c(\vec{x})g_d(\vec{x})}{g_c(\vec{x})+g_d(\vec{x})}\,.
\end{equation}

\noindent\emph{Consistency between XY-motion and Z-motion.} In practice we notice that motion discontinuity is preserved better in XY-motion than in Z-motion. To improve the accuracy of the Z-motion, we solve the 3D motion $\vec{V}$ in a two-step process. We compute an initial estimate of the XY-motion, denoted as $\vec{U}=(U_X,U_Y)$, in the first pass.
We then use $\vec{U}$ to compute a weight map for the regularization term:
\begin{equation}
g_c(\vec{x}) = \frac{1}{1+(|\nabla U_X|^2+|\nabla U_Y|^2)/\sigma_c^2}\,,
\end{equation}
where $\mathscr{N}(\vec{x})$ denotes a local neighborhood of the point $\vec{x}$. Then the full 3D motion $\vec{V}$ is computed in a second pass. Notice that $g(\vec{x})$ is small where gradient of $\vec{U}$ is large, in other words the regularization term will contribute less to the whole energy where there is a discontinuity in $\vec{U}$. 

\noindent\emph{Consistency between motion boundaries and depth boundaries.} We also assume the motion boundaries are likely to align with depth boundaries. In other words, we give a lower weight for points where the depth gradient is large:
\begin{equation}
g_d(\vec{x}) = \frac{1}{1+|\nabla d_\alpha|^2/\sigma_d^2}\,.
\end{equation}

\subsubsection{Optimization}
The error term $E_D(V)$ can be linearized as ,
\begin{equation}
E_D' (\vec{V})=\int_{\Omega_c}\sum_{\vec{x}_i\in \mathscr{P}(u,v)}h_i\rho_D((L_{Xi}V_X+L_{Yi}V_Y+L_{Zi}V_Z+L_{ti})^2)du\,dv\,.
\end{equation}
Then the entire energy $E'=E_D'+E_S$ can be minimized using Euler-Lagrange equations:
\begin{equation}
\begin{split}
\sum_{\vec{x}_i\in \mathscr{P}(u,v)}h_i\rho_D'L_X\delta_L-\lambda\sum_{i=1}^2\frac{\partial}{\partial u^{(i)}}(g\rho_S'(V_{X(i)})V_{X(i)})= \\
-\sum_{\vec{x}_i\in \mathscr{P}(u,v)}h_i\rho_D'L_XL_t\,,  \\
\sum_{\vec{x}_i\in \mathscr{P}(u,v)}h_i\rho_D'L_Y\delta_L-\lambda\sum_{i=1}^2\frac{\partial}{\partial u^{(i)}}(g\rho_S'(V_{Y(i)})V_{Y(i)})=\\
-\sum_{\vec{x}_i\in \mathscr{P}(u,v)}h_i\rho_D'L_YL_t\,, \\
\sum_{\vec{x}_i\in \mathscr{P}(u,v)}h_i\rho_D'L_Z\delta_L-\lambda_Z\sum_{i=1}^2\frac{\partial}{\partial u^{(i)}}(g\rho_S'(V_{Z(i)})V_{Z(i)})=\\
-\sum_{\vec{x}_i\in \mathscr{P}(u,v)}h_i\rho_D'L_ZL_t \,,
\end{split}
\end{equation}
where $\rho_D'$ is short for $\rho_D'((L_XV_X+L_YV_Y+L_ZV_Z+L_t)^2)$, $\delta_L=L_XV_X+L_YV_Y+L_ZV_Z$. Again, these equations are discretized and solved using SOR. The linearization step can then be repeated in an iterative, multi-resolution framework. 

\bibliographystyle{spbasic}      
\bibliography{ms}   

\begin{thebibliography}{42}
\providecommand{\natexlab}[1]{#1}
\providecommand{\url}[1]{{#1}}
\providecommand{\urlprefix}{URL }
\expandafter\ifx\csname urlstyle\endcsname\relax
  \providecommand{\doi}[1]{DOI~\discretionary{}{}{}#1}\else
  \providecommand{\doi}{DOI~\discretionary{}{}{}\begingroup
  \urlstyle{rm}\Url}\fi
\providecommand{\eprint}[2][]{\url{#2}}

\bibitem[{Adelson and Wang(1992)}]{adelson_single_1992}
Adelson EH, Wang JYA (1992) Single {Lens} {Stereo} with a {Plenoptic} {Camera}.
  IEEE Transactions on Pattern Analysis and Machine Intelligence (TPAMI)
  14(2):99--106

\bibitem[{Alexander et~al.(2016)Alexander, Guo, Koppal, Gortler, and
  Zickler}]{alexander2016focal}
Alexander E, Guo Q, Koppal S, Gortler S, Zickler T (2016) Focal flow: Measuring
  distance and velocity with defocus and differential motion. In: European
  Conference on Computer Vision (ECCV), Springer, Heidelberg, pp 667--682

\bibitem[{Black and Anandan(1996)}]{black1996robust}
Black MJ, Anandan P (1996) The robust estimation of multiple motions:
  Parametric and piecewise-smooth flow fields. Computer Vision and Image
  Understanding 63(1):75--104

\bibitem[{Bok et~al.(2017)Bok, Jeon, and Kweon}]{bok2017geometric}
Bok Y, Jeon HG, Kweon IS (2017) Geometric calibration of micro-lens-based light
  field cameras using line features. IEEE Transactions on Pattern Analysis and
  Machine Intelligence (TPAMI) 39(2):287--300

\bibitem[{Brox et~al.(2004)Brox, Bruhn, Papenberg, and Weickert}]{brox2004high}
Brox T, Bruhn A, Papenberg N, Weickert J (2004) High accuracy optical flow
  estimation based on a theory for warping. European Conference on Computer
  Vision (ECCV) pp 25--36

\bibitem[{Bruhn et~al.(2005)Bruhn, Weickert, and Schn{\"o}rr}]{bruhn2005lucas}
Bruhn A, Weickert J, Schn{\"o}rr C (2005) Lucas/kanade meets horn/schunck:
  Combining local and global optic flow methods. International journal of
  computer vision (IJCV) 61(3):211--231

\bibitem[{Chandraker(2014{\natexlab{a}})}]{chandraker2014shape}
Chandraker M (2014{\natexlab{a}}) On shape and material recovery from motion.
  In: European Conference on Computer Vision (ECCV), Springer, Heidelberg, pp
  202--217

\bibitem[{Chandraker(2014{\natexlab{b}})}]{chandraker2014camera}
Chandraker M (2014{\natexlab{b}}) What camera motion reveals about shape with
  unknown brdf. In: IEEE Conference on Computer Vision and Pattern Recognition
  (CVPR), IEEE, Washington, pp 2171--2178

\bibitem[{Chandraker(2016)}]{chandraker2016information}
Chandraker M (2016) The information available to a moving observer on shape
  with unknown, isotropic brdfs. IEEE Transactions on Pattern Analysis and
  Machine Intelligence (TPAMI) 38(7):1283--1297

\bibitem[{Dansereau et~al.(2011)Dansereau, Mahon, Pizarro, and
  Williams}]{dansereau2011plenoptic}
Dansereau DG, Mahon I, Pizarro O, Williams SB (2011) Plenoptic flow:
  Closed-form visual odometry for light field cameras. In: IEEE/RSJ
  International Conference on Intelligent Robots and Systems (IROS), IEEE,
  Washington, pp 4455--4462

\bibitem[{Dansereau et~al.(2017)Dansereau, Schuster, Ford, and
  Wetzstein}]{dansereau2017wide}
Dansereau DG, Schuster G, Ford J, Wetzstein G (2017) A wide-field-of-view
  monocentric light field camera. In: IEEE Conference on Computer Vision and
  Pattern Recognition (CVPR), IEEE, Washington

\bibitem[{Gottfried et~al.(2011)Gottfried, Fehr, and
  Garbe}]{gottfried2011computing}
Gottfried JM, Fehr J, Garbe CS (2011) Computing range flow from multi-modal
  kinect data. In: International Symposium on Visual Computing, Springer,
  Heidelberg, pp 758--767

\bibitem[{Hasinoff et~al.(2010)Hasinoff, Durand, and
  Freeman}]{hasinoff2010noise}
Hasinoff SW, Durand F, Freeman WT (2010) Noise-optimal capture for high dynamic
  range photography. In: IEEE Conference on Computer Vision and Pattern
  Recognition (CVPR), IEEE, pp 553--560

\bibitem[{Heber and Pock(2014)}]{heber2014scene}
Heber S, Pock T (2014) Scene flow estimation from light fields via the
  preconditioned primal-dual algorithm. In: Jiang X, Hornegger J, Koch R (eds)
  Pattern Recognition, Springer International, Cham, pp 3--14

\bibitem[{Horn and Schunck(1981)}]{horn1981determining}
Horn BK, Schunck BG (1981) Determining optical flow. Artificial intelligence
  17(1-3):185--203

\bibitem[{Hung et~al.(2013)Hung, Xu, and Jia}]{hung2013consistent}
Hung CH, Xu L, Jia J (2013) Consistent binocular depth and scene flow with
  chained temporal profiles. International journal of computer vision (IJCV)
  102(1-3):271--292

\bibitem[{Jaimez et~al.(2015)Jaimez, Souiai, Gonzalez-Jimenez, and
  Cremers}]{jaimez2015primal}
Jaimez M, Souiai M, Gonzalez-Jimenez J, Cremers D (2015) A primal-dual
  framework for real-time dense rgb-d scene flow. In: IEEE International
  Conference on Robotics and Automation (ICRA), IEEE, Washington, pp 98--104

\bibitem[{Jo et~al.(2015)Jo, Gupta, and Nayar}]{jo_spedo_2015}
Jo K, Gupta M, Nayar SK (2015) {SpeDo} : 6 {DOF} {Ego}-{Motion} {Sensor}
  {Using} {Speckle} {Defocus} {Imaging}. In: {IEEE} {International}
  {Conference} on {Computer} {Vision} ({ICCV}), IEEE, Washington, pp 4319--4327

\bibitem[{Johannsen et~al.(2015)Johannsen, Sulc, and
  Goldluecke}]{johannsen2015linear}
Johannsen O, Sulc A, Goldluecke B (2015) On linear structure from motion for
  light field cameras. In: IEEE International Conference on Computer Vision
  (ICCV), IEEE, Washington, pp 720--728

\bibitem[{Letouzey et~al.(2011)Letouzey, Petit, and Boyer}]{letouzey2011scene}
Letouzey A, Petit B, Boyer E (2011) Scene flow from depth and color images. In:
  British Machine Vision Conference (BMVC), BMVA Press, pp 46--56

\bibitem[{Levoy and Hanrahan(1996)}]{levoy1996light}
Levoy M, Hanrahan P (1996) Light field rendering. In: SIGGRAPH Conference on
  Computer Graphics and Interactive Techniques, ACM, New York, pp 31--42

\bibitem[{Li et~al.(2017)Li, Xu, Ramamoorthi, and Chandraker}]{li2017robust}
Li Z, Xu Z, Ramamoorthi R, Chandraker M (2017) Robust energy minimization for
  brdf-invariant shape from light fields. In: IEEE Conference on Computer
  Vision and Pattern Recognition (CVPR), IEEE, Washington, vol~1

\bibitem[{Lucas et~al.(1981)Lucas, Kanade et~al.}]{lucas1981iterative}
Lucas BD, Kanade T, et~al. (1981) An iterative image registration technique
  with an application to stereo vision. In: International Joint Conference on
  Artificial Intelligence, Morgan Kaufmann, San Francisco, pp 674--679

\bibitem[{Ma et~al.(2018)Ma, Smith, and Gupta}]{ma_3d_2018}
Ma S, Smith BM, Gupta M (2018) 3{D} scene flow from 4{D} light field gradients.
  In: European {Conference} on {Computer} {Vision} (ECCV), Springer
  International Publishing, vol~8, pp 681--698

\bibitem[{Navarro and Garamendi(2016)}]{navarro2016variational}
Navarro J, Garamendi J (2016) Variational scene flow and occlusion detection
  from a light field sequence. In: International Conference on Systems, Signals
  and Image Processing (IWSSIP), IEEE, Washington, pp 1--4

\bibitem[{Neumann et~al.(2003)Neumann, Fermuller, and
  Aloimonos}]{neumann2003polydioptric}
Neumann J, Fermuller C, Aloimonos Y (2003) Polydioptric camera design and 3d
  motion estimation. In: IEEE Conference on Computer Vision and Pattern
  Recognition (CVPR), IEEE, Washington, vol~2, pp II--294

\bibitem[{Neumann et~al.(2004)Neumann, Ferm{\"u}ller, and
  Aloimonos}]{neumann2004hierarchy}
Neumann J, Ferm{\"u}ller C, Aloimonos Y (2004) A hierarchy of cameras for 3d
  photography. Computer Vision and Image Understanding 96(3):274--293

\bibitem[{Ng et~al.(2005)Ng, Levoy, Br{\'e}dif, Duval, Horowitz, and
  Hanrahan}]{ng2005light}
Ng R, Levoy M, Br{\'e}dif M, Duval G, Horowitz M, Hanrahan P (2005) Light field
  photography with a hand-held plenoptic camera. Computer Science Technical
  Report CSTR 2(11):1--11

\bibitem[{Odobez and Bouthemy(1995)}]{odobez1995robust}
Odobez JM, Bouthemy P (1995) Robust multiresolution estimation of parametric
  motion models. Journal of Visual Communication and Image Representation
  6(4):348--365

\bibitem[{Phong(1975)}]{phong_illumination_1975}
Phong BT (1975) Illumination for computer generated pictures. Communications of
  the ACM 18(6):311--317, \doi{10.1145/360825.360839}

\bibitem[{Shi and Tomasi(1994)}]{shi1994good}
Shi J, Tomasi C (1994) Good features to track. In: IEEE Conference on Computer
  Vision and Pattern Recognition (CVPR), IEEE, Washington, pp 593--600

\bibitem[{Smith et~al.(2018)Smith, O'Toole, and Gupta}]{smith_cvpr2018}
Smith B, O'Toole M, Gupta M (2018) Tracking multiple objects outside the line
  of sight using speckle imaging. In: IEEE Conference on Computer Vision and
  Pattern Recognition (CVPR), IEEE

\bibitem[{Smith et~al.(2017)Smith, Desai, Agarwal, and
  Gupta}]{smith_colux:_2017}
Smith BM, Desai P, Agarwal V, Gupta M (2017) {CoLux}: multi-object 3d
  micro-motion analysis using speckle imaging. ACM Transactions on Graphics
  36(4):1--12

\bibitem[{Srinivasan et~al.(2015)Srinivasan, Tao, Ng, and
  Ramamoorthi}]{srinivasan2015oriented}
Srinivasan PP, Tao MW, Ng R, Ramamoorthi R (2015) Oriented light-field windows
  for scene flow. In: IEEE International Conference on Computer Vision (ICCV),
  IEEE, Washington, pp 3496--3504

\bibitem[{Sun et~al.(2010)Sun, Roth, and Black}]{sun2010secrets}
Sun D, Roth S, Black MJ (2010) Secrets of optical flow estimation and their
  principles. In: IEEE Conference on Computer Vision and Pattern Recognition
  (CVPR), IEEE, Washington, pp 2432--2439

\bibitem[{Sun et~al.(2015)Sun, Sudderth, and Pfister}]{sun2015layered}
Sun D, Sudderth EB, Pfister H (2015) Layered rgbd scene flow estimation. In:
  IEEE Conference on Computer Vision and Pattern Recognition (CVPR), IEEE,
  Washington, pp 548--556

\bibitem[{Tao et~al.(2013)Tao, Hadap, Malik, and Ramamoorthi}]{tao2013depth}
Tao MW, Hadap S, Malik J, Ramamoorthi R (2013) Depth from combining defocus and
  correspondence using light-field cameras. In: IEEE International Conference
  on Computer Vision (ICCV), IEEE, Washington, pp 673--680

\bibitem[{Vedula et~al.(1999)Vedula, Baker, Rander, Collins, and
  Kanade}]{vedula1999three}
Vedula S, Baker S, Rander P, Collins R, Kanade T (1999) Three-dimensional scene
  flow. In: IEEE International Conference on Computer Vision (ICCV), IEEE,
  Washington, vol~2, pp 722--729

\bibitem[{Wang et~al.(2016)Wang, Chandraker, Efros, and
  Ramamoorthi}]{wang2016svbrdf}
Wang TC, Chandraker M, Efros AA, Ramamoorthi R (2016) Svbrdf-invariant shape
  and reflectance estimation from light-field cameras. In: IEEE Conference on
  Computer Vision and Pattern Recognition (CVPR), IEEE, Washington, pp
  5451--5459

\bibitem[{Wanner and Goldluecke(2014)}]{wanner2014variational}
Wanner S, Goldluecke B (2014) Variational light field analysis for disparity
  estimation and super-resolution. IEEE Transactions on Pattern Analysis and
  Machine Intelligence (TPAMI) 36(3):606--619

\bibitem[{Wedel et~al.(2008)Wedel, Rabe, Vaudrey, Brox, Franke, and
  Cremers}]{wedel2008efficient}
Wedel A, Rabe C, Vaudrey T, Brox T, Franke U, Cremers D (2008) Efficient dense
  scene flow from sparse or dense stereo data. In: European Conference on
  Computer Vision (ECCV), Springer, Heidelberg, pp 739--751

\bibitem[{Zhang et~al.(2017)Zhang, Li, Yang, Yu, Lin, and Yu}]{zhang2017light}
Zhang Y, Li Z, Yang W, Yu P, Lin H, Yu J (2017) The light field 3d scanner. In:
  IEEE International Conference on Computational Photography (ICCP), IEEE,
  Washington, pp 1--9

\end{thebibliography}


\end{document}


\maketitle

In this report we show the full 3D visualization of the recovered motion for the experiments described in the main paper. 

\begin{figure}[ht]
  \begin{center}
  \includegraphics[width=\linewidth]{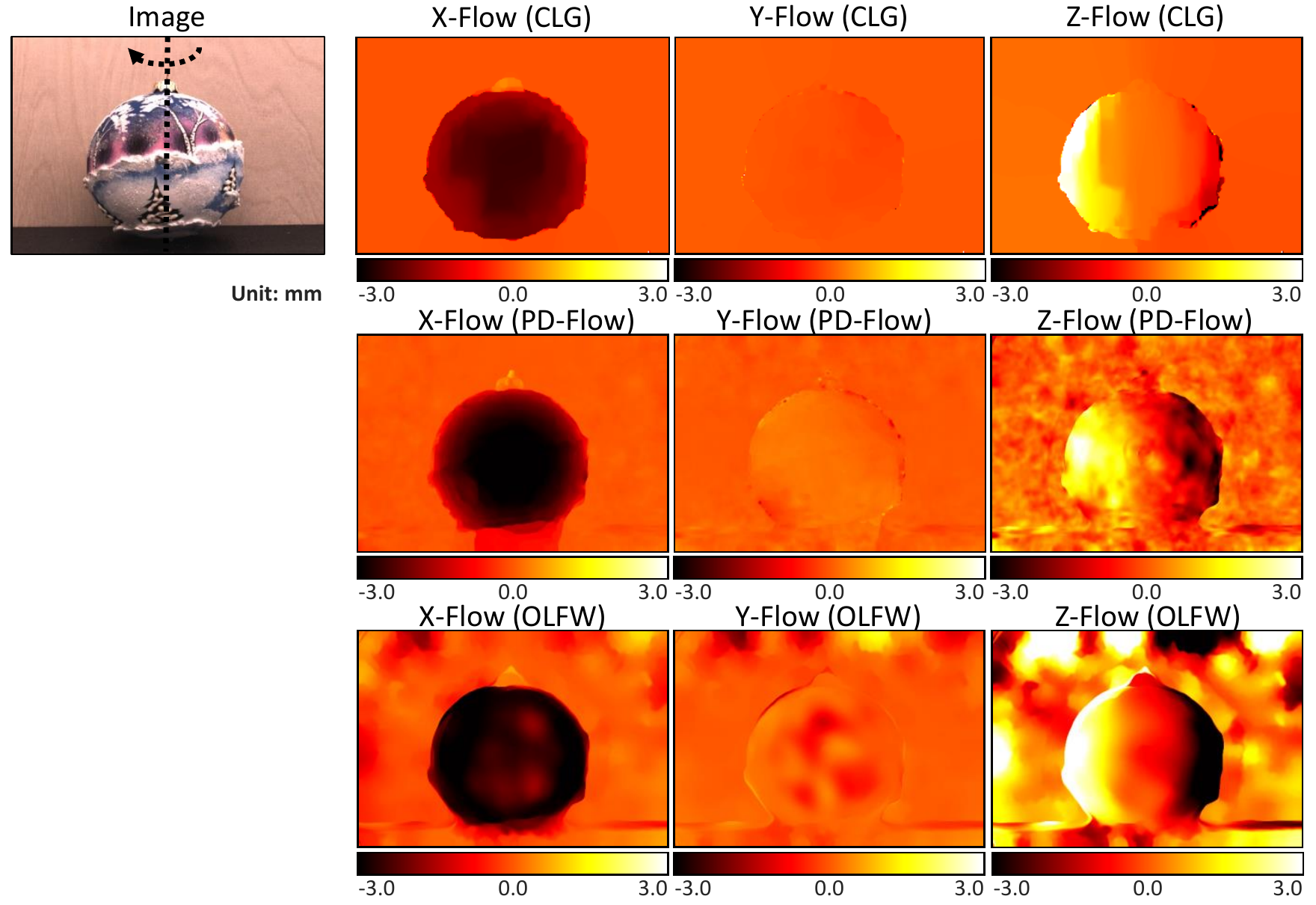}
  \end{center}
\caption{\textbf{Recovering non-planar motion.} A rotating spherical ornament. All methods can estimate the gradually changing Z-motion, but only our method recovers the background correctly.}
\label{fig:exp_sphere}
\end{figure}

\begin{figure}[t]
  \begin{center}
  \includegraphics[width=\linewidth]{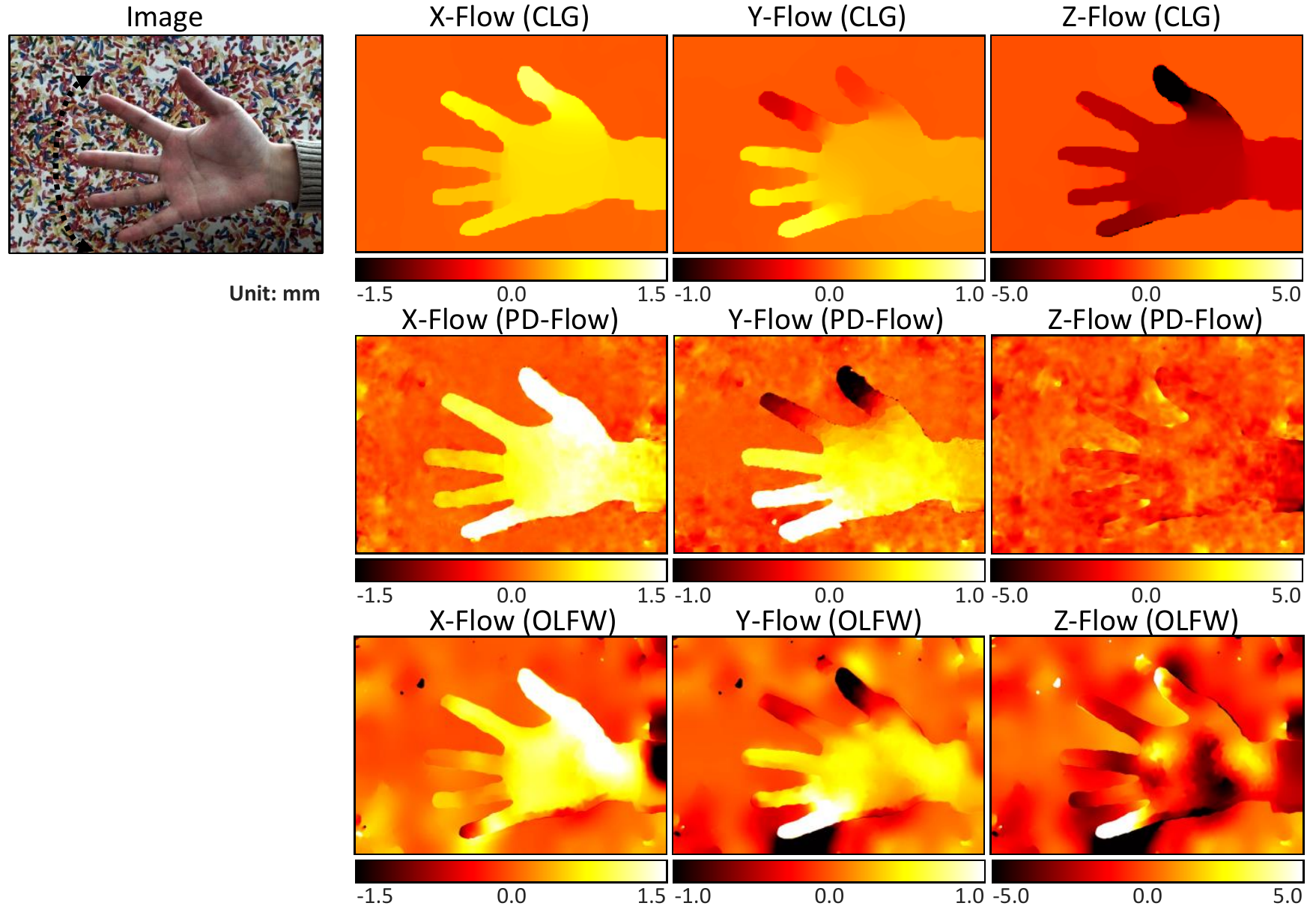}
  \end{center}
\caption{\textbf{Recovering non-rigid motion.} An expanding hand. The expansion is demonstrated by the different Y-motion of the fingers.}
\end{figure}

\begin{figure}[t]
  \begin{center}
  \includegraphics[width=\linewidth]{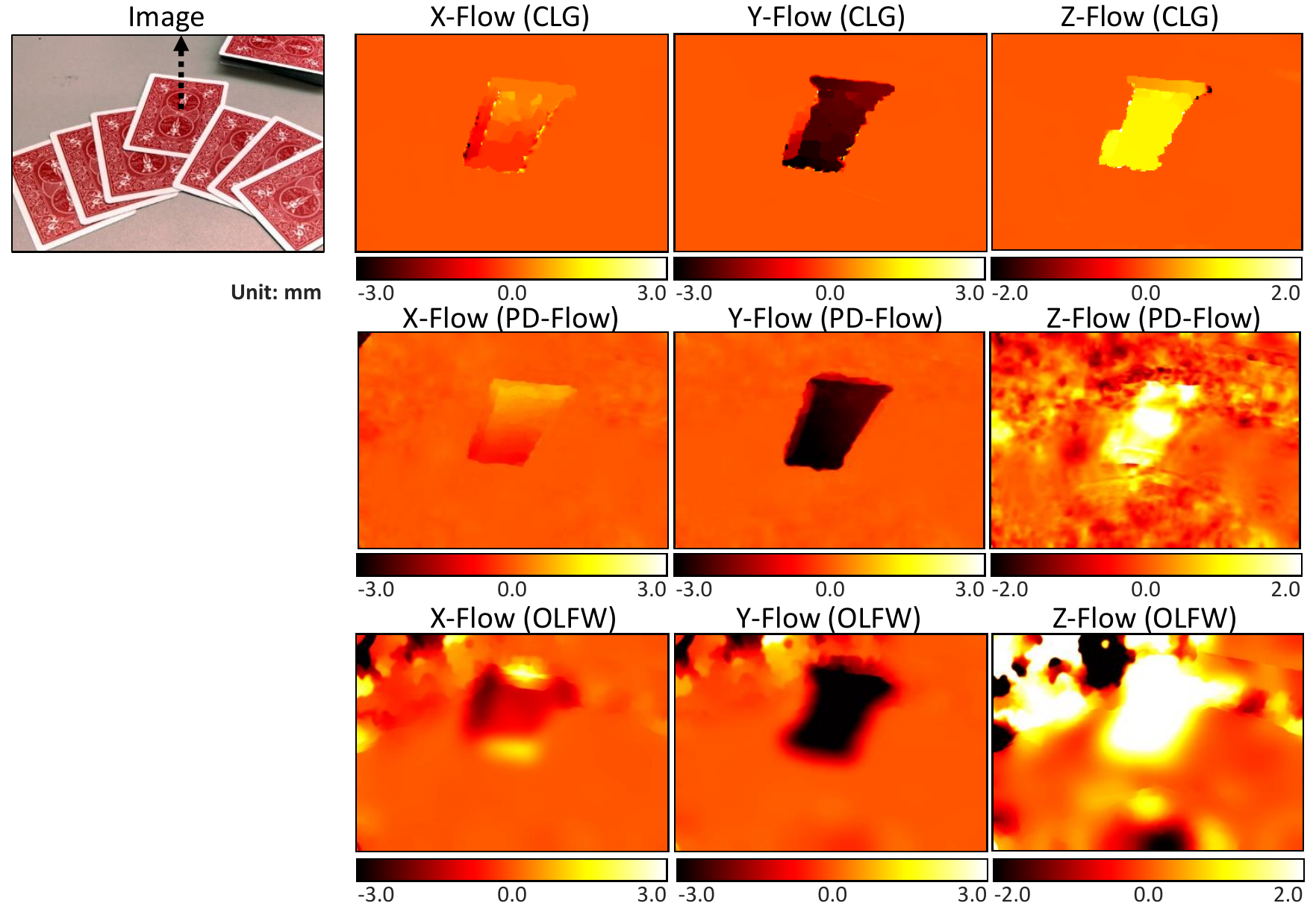}
  \end{center}
\caption{\textbf{Recovering motion in natural environments with occlusions.} The card in the center moves upward. Our method can recover the motion of the card, despite occlusions and lack of texture around the white boundaries.}
\end{figure}

\begin{figure}[t]
  \begin{center}
  \includegraphics[width=\linewidth]{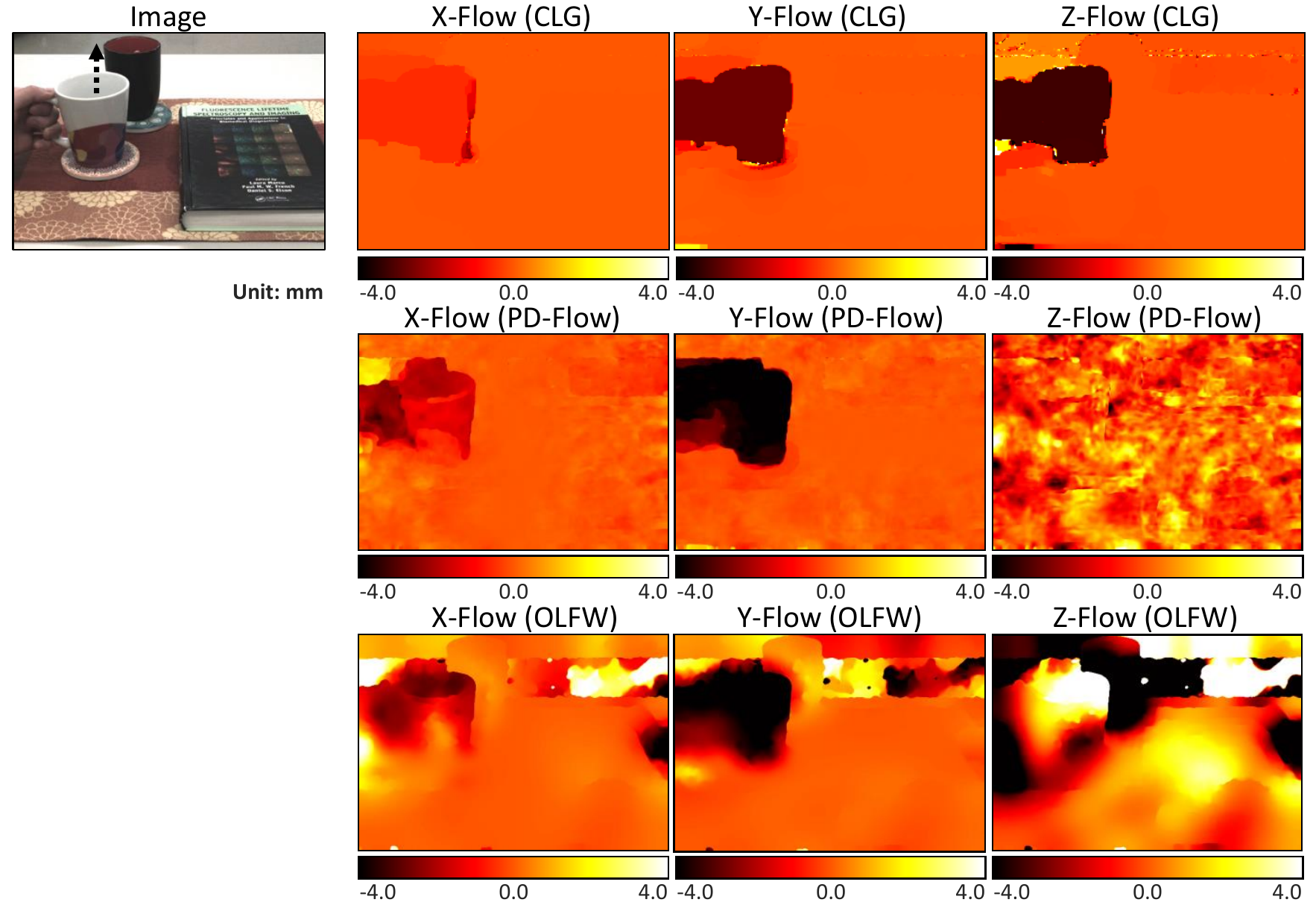}
  \end{center}
\caption{\textbf{Recovering motion in natural environments with occlusions.} The mug on the left is picked up by a hand. Our method estimates the motion boundaries accurately.}
\end{figure}

\begin{figure}[t]
  \begin{center}
  \includegraphics[width=\linewidth]{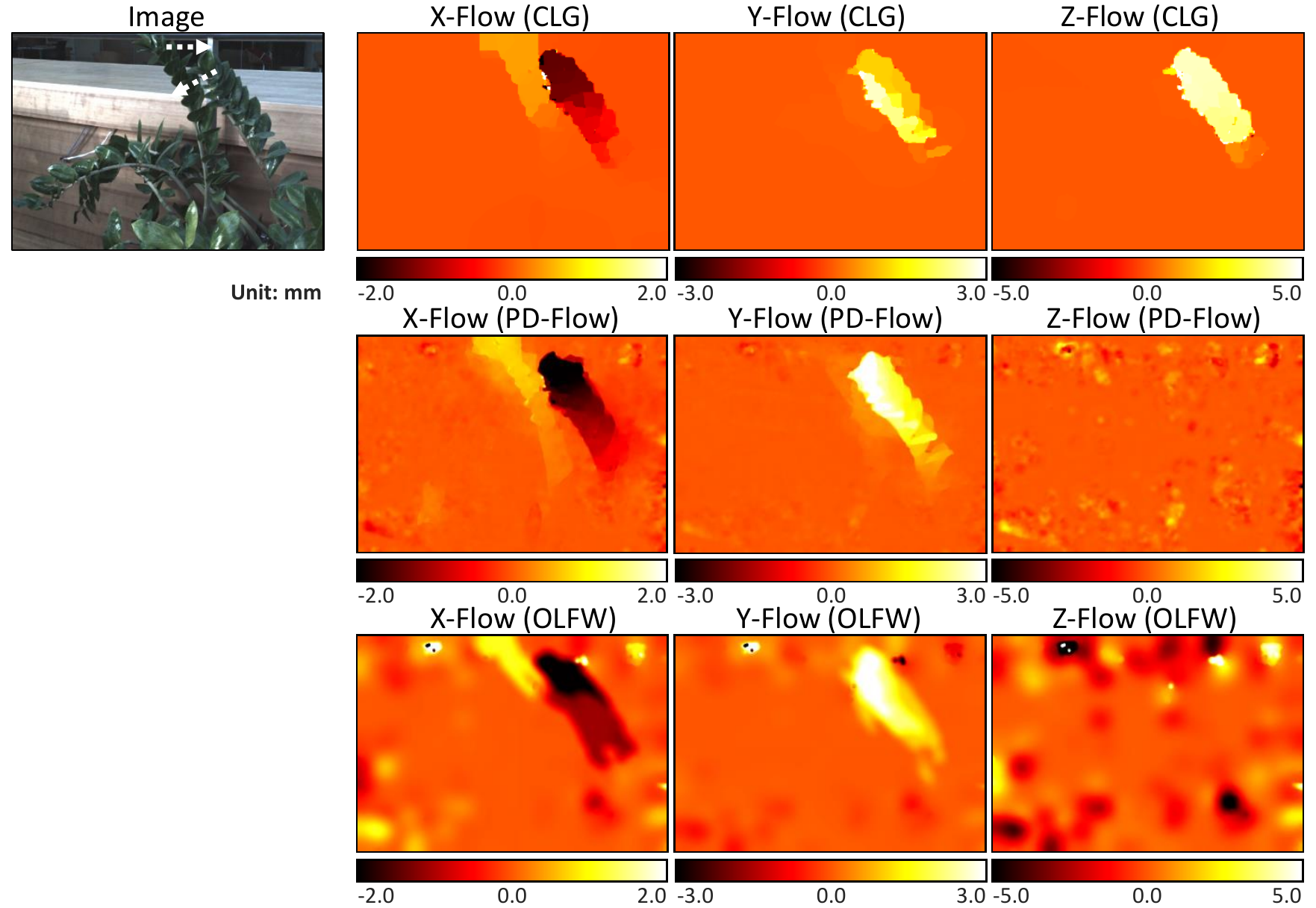}
  \end{center}
\caption{\textbf{Recovering motion in natural environments with occlusions.} The top two vertical branches of the plant quiver in the wind. Our method can correctly compute the motion of the two complex-shaped branches.}
\end{figure}

\begin{figure}[t]
  \begin{center}
  \includegraphics[width=\linewidth]{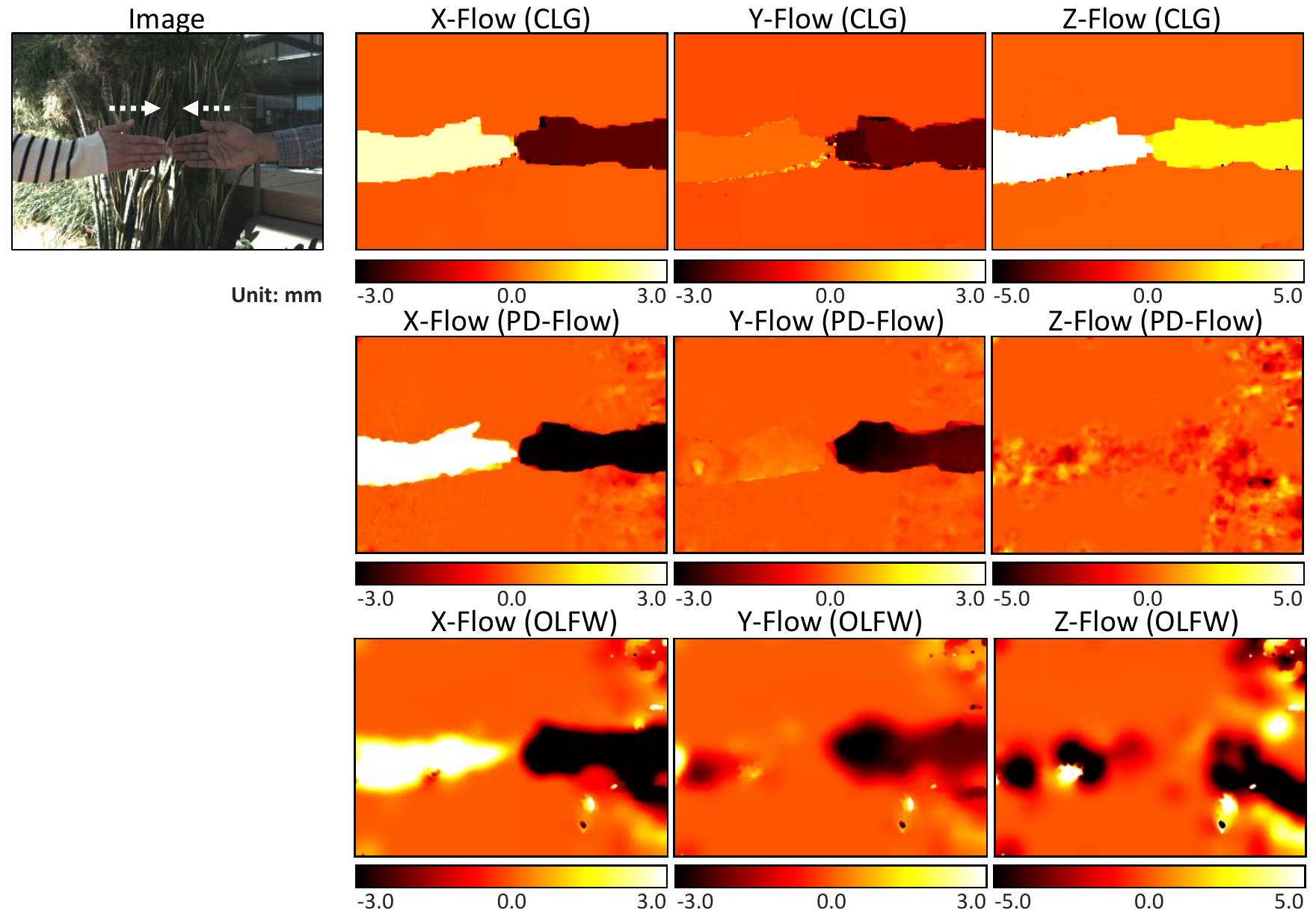}
  \end{center}
\caption{\textbf{Recovering human actions.} Handshaking. All the three methods compute the joining movements of the hands correctly, while our method preserves the hand boundary best.}
\end{figure}

\begin{figure}[t]
  \begin{center}
  \includegraphics[width=\linewidth]{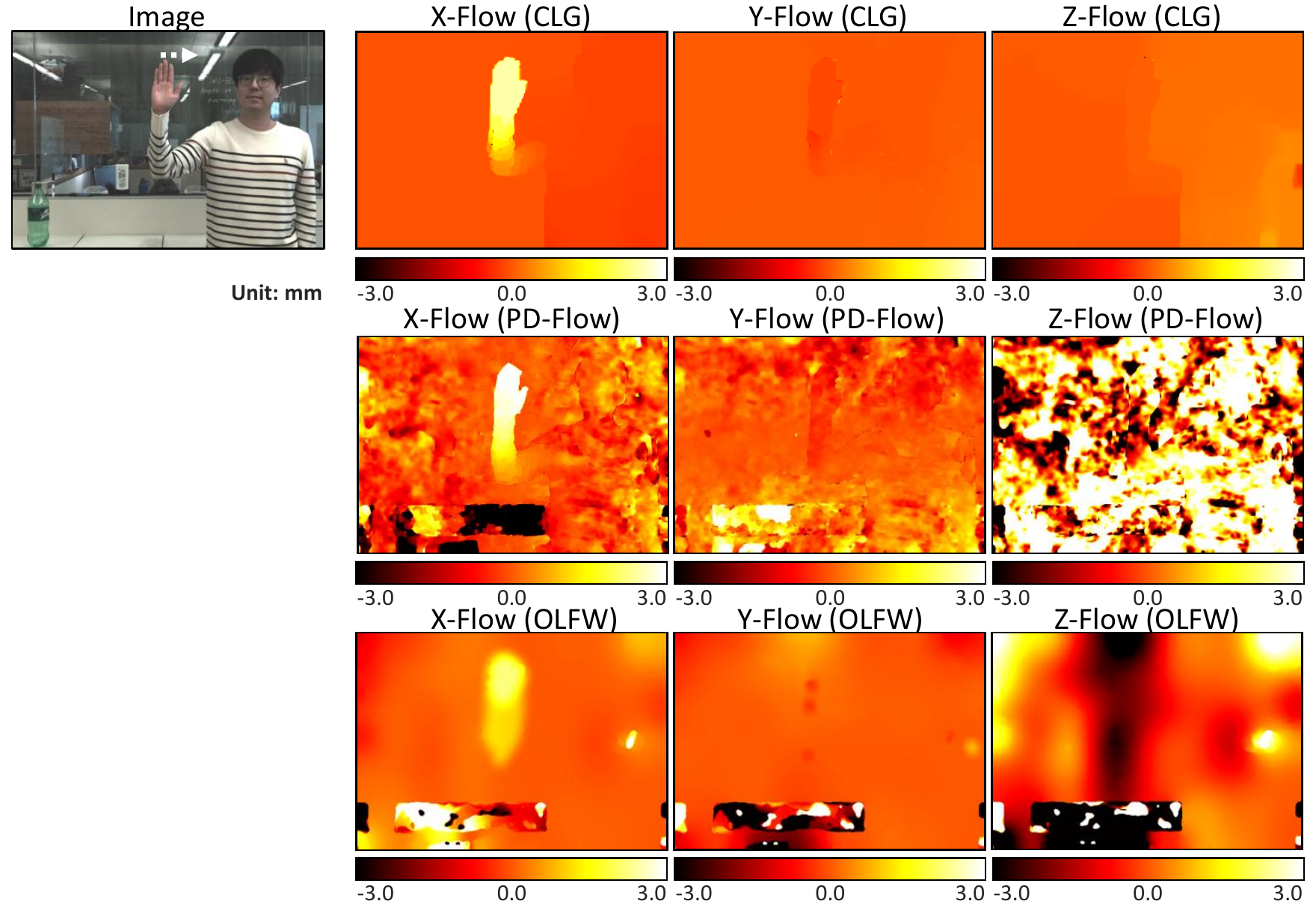}
  \end{center}
\caption{\textbf{Recovering human actions.} Waving hand. Our method correctly estimates the motion in spite of the reflections and textureless regions in the background, which is challenging for depth estimation algorithms.}
\end{figure}

\begin{figure}[t]
  \begin{center}
  \includegraphics[width=\linewidth]{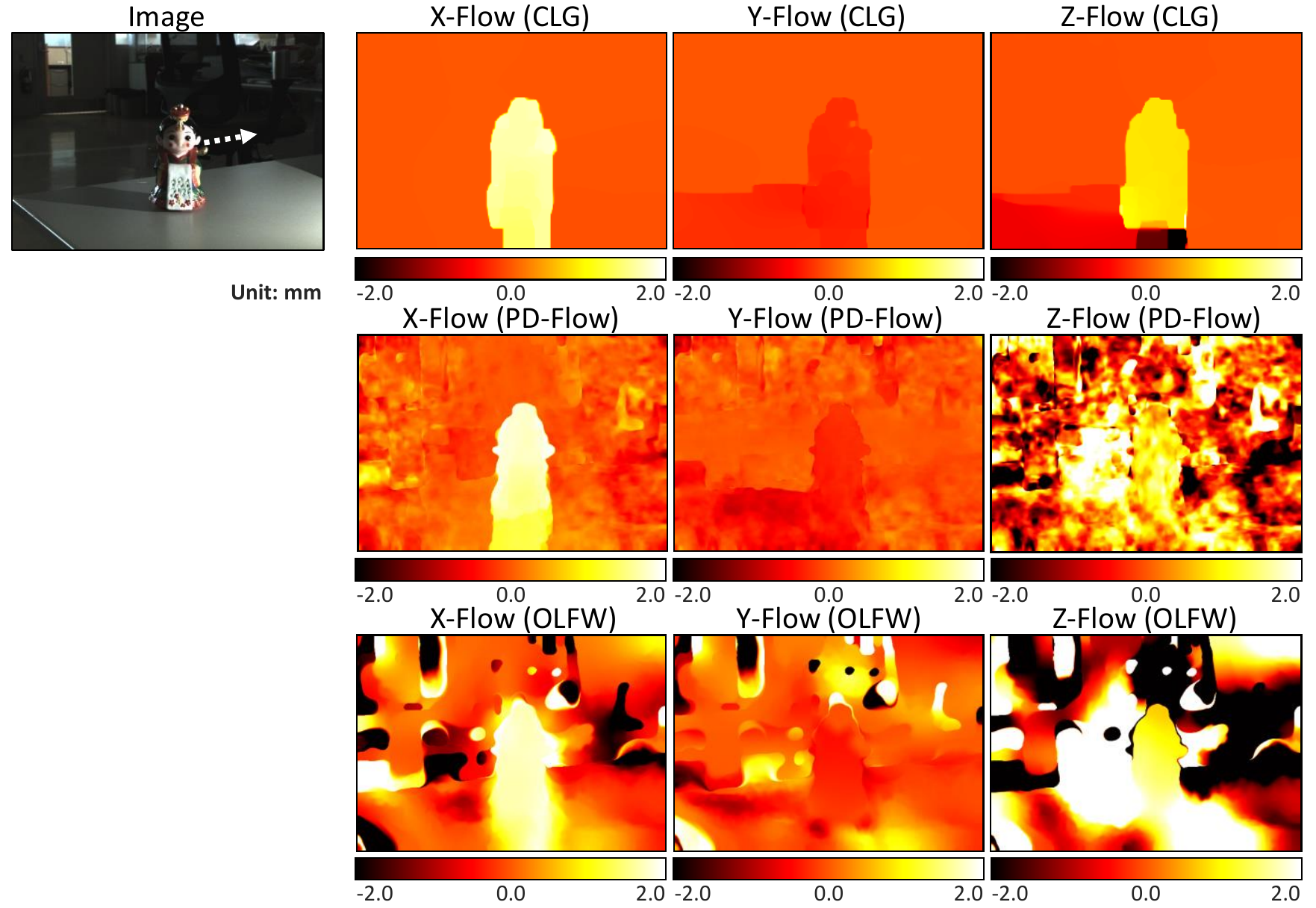}
  \end{center}
\caption{\textbf{Recovering motion under challenging lighting conditions.} A figurine moves under weak, directional lighting. Our method still preserves the overall shape of the object, although its reflection on the table is also regarded as moving.}
\end{figure}

\begin{figure}[t]
  \begin{center}
  \includegraphics[width=\linewidth]{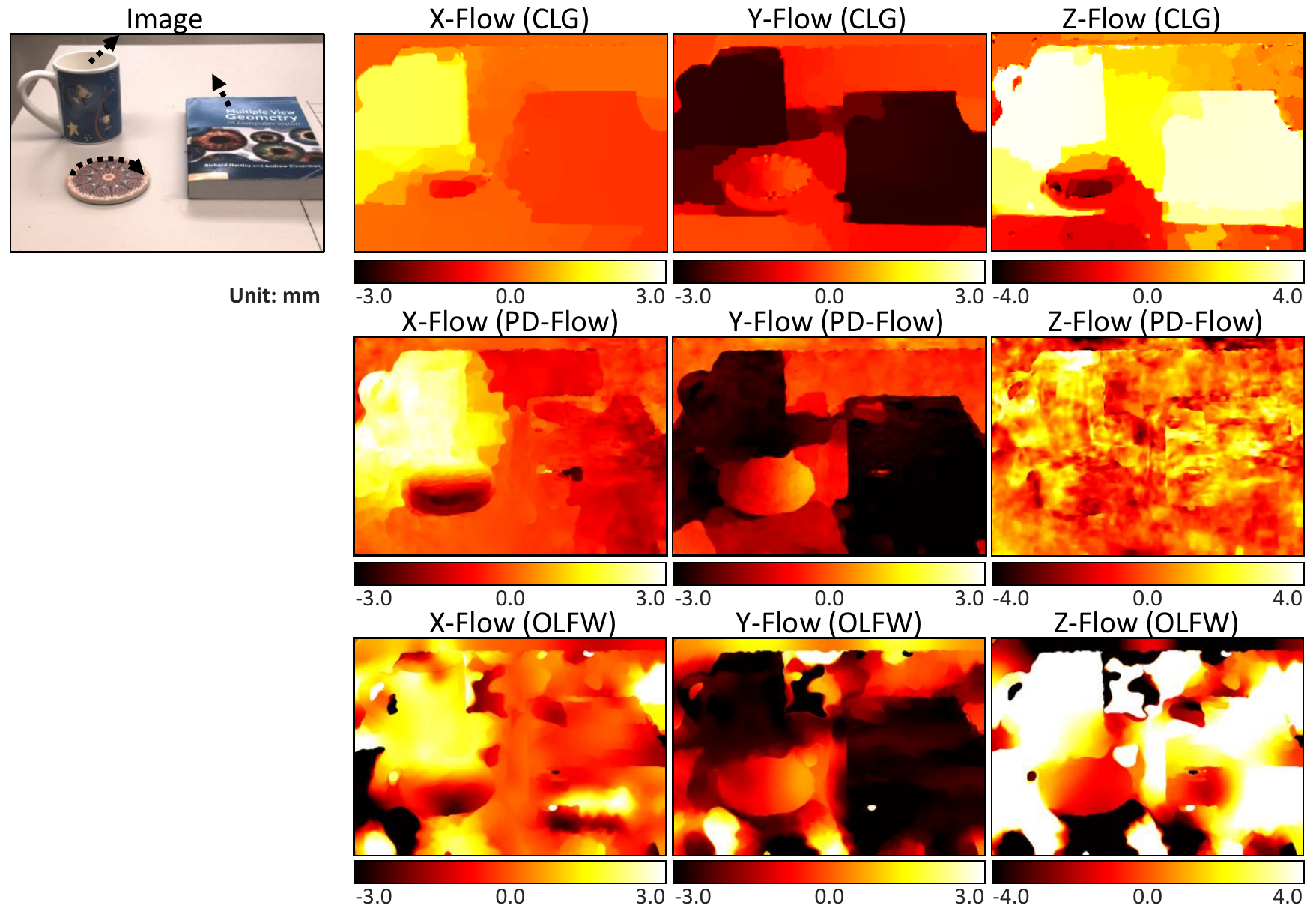}
  \end{center}
\caption{\textbf{Recovering motion under challenging lighting conditions.} A few objects move independently. Due to shadows and lack of texture in the background, boundaries of the objects are not distinguishable in the recovered motion field of all the three methods.}
\label{fig:exp_desktop}
\end{figure}
